%% file: arXiv.tex
\pdfoutput=1

\documentclass[10pt,twocolumn,letterpaper]{article}

\usepackage[pagenumbers]{cvpr} 

\input{preamble}
\input{math_commands.tex}

\definecolor{cvprblue}{rgb}{0.21,0.49,0.74}
\usepackage[pagebackref,breaklinks,colorlinks,citecolor=cvprblue]{hyperref}

\def\paperID{*****} 
\def\confName{CVPR}
\def\confYear{2024}

\title{On Exact Inversion of DPM-Solvers}

\author{Seongmin Hong$^{1}$, \quad Kyeonghyun Lee$^{1}$, \quad Suh Yoon Jeon$^{1}$, \quad Hyewon Bae$^{1}$, \quad Se Young Chun$^{1,2,}$\thanks{Corresponding author} \\
$^1$Dept. of ECE, \ $^2$INMC \& IPAI, \ Seoul National University, \ Republic of Korea\\
{\tt\small \{smhongok, litiphysics, euniejeon, hyewon0309, sychun\}@snu.ac.kr}
}

\begin{document}
\maketitle 

\begin{abstract}
Diffusion probabilistic models (DPMs) are a key component in modern generative models. DPM-solvers have achieved reduced latency and enhanced quality significantly, but have posed challenges to find the exact inverse (\textit{i.e.}, finding the initial noise from the given image). Here we investigate the exact inversions for DPM-solvers and propose algorithms to perform them when samples are generated by the first-order as well as higher-order DPM-solvers. For each explicit denoising step in DPM-solvers, we formulated the inversions using implicit methods such as gradient descent or forward step method to ensure the robustness to large classifier-free guidance unlike the prior approach using fixed-point iteration. Experimental results demonstrated that our proposed exact inversion methods significantly reduced the error of both image and noise reconstructions, greatly enhanced the ability to distinguish invisible watermarks and well prevented unintended background changes consistently during image editing. Project page: \url{https://smhongok.github.io/inv-dpm.html}. 
\end{abstract}

\section{Introduction}

Diffusion probabilistic models (DPMs) are rapidly advancing as a key component in modern generative models for various applications such as unconditional image generation~\cite{ho2020denoising,song2020denoising,song2020improved,rombach2022high}, conditional image synthesis~\cite{ho2021classifierfree,rombach2022high} including text-guided image generation~\cite{ramesh2022hierarchical,rombach2022high,ruiz2023dreambooth} and solving inverse problems in imaging~\cite{lugmayr2022repaint, li2022srdiff}.
DPMs create (or sample) diverse and high-quality images by gradually denoising random initial noises either in the image domain~\cite{xiao2021tackling} or in the latent space~\cite{rombach2022high} (called latent diffusion model or LDM).
However, this iterative denoising in DPMs usually takes a long sampling time~\cite{xiao2021tackling}.

There have been a considerable amount of studies to speed up the sampling time or the generative process in DPMs~\cite{song2020denoising,lu2022dpm, lu2022dpm++, lyu2022accelerating, zheng2023fast, salimans2022progressive, meng2023distillation}. For example, denoising diffusion implicit model (DDIM)~\cite{song2020denoising} has attempted to reduce the iterations (or steps) by formulating the denoising process of DPM as an ordinary differential equation (ODE), namely the diffusion ODE, and then by using the forward Euler method to sample a high-quality image 
with much fewer denoising steps (\textit{e.g.}, 50) than the diffusion steps (\textit{e.g.}, 1000) that were used in training. High-order DPM-solvers~\cite{lu2022dpm, lu2022dpm++} leverage fast ODE solvers such as exponential integrators to further reduce the number of denoising steps (\textit{e.g.}, 10), leading to significantly decreased sampling time compared to DDIM (first-order DPM-solver). However, fast DPM-solvers make it challenging to trace back the generative process and find the initial noise for a given image.

There have been great interests in tracing the generative process back, or \emph{inversion}, which is a key component in a number of applications such as image editing~\citep{hertz2022prompt, Kim_2022_CVPR, patashnik2023localizing, wallace2023edict}, style transfer~\cite{zhang2023inversion}, image-to-image translation~\cite{su2022dual}, model attacks~\cite{chen2023trojdiff}, watermark detection~\cite{wen2023tree} and image restoration~\cite{kawar2022denoising}.
For example, image editing using DPM involves finding the latent vector for a given image through \emph{inversion} and then using a different prompt in the generation process from that latent noise~\citep{hertz2022prompt,su2022dual,wallace2023edict}.
Unfortunately, the \emph{exact inversion} of DPM-solvers is challenging. The na\"ive DDIM inversion does not subtract the estimated Gaussian noise, but adds it to the clean image to find the corresponding initial noise. As DDIM solves the diffusion ODE using the forward Euler method, the na\"ive DDIM inversion uses the same method in reverse order along the time axis. This inversion is valid under the assumption that the estimated noises are almost the same in both $t$ and $t+dt$, where $dt$ is the time step. While this assumption approximately holds for the methods with many (small) diffusion steps, DPM-solvers with fewer denoising steps will break this assumption so that the na\"ive DDIM inversion will not properly work anymore, leading to distortions~\citep{wallace2023edict}.

Recently, several exact inversion methods have been proposed to achieve smaller reconstruction errors compared to the na\"ive DDIM inversion. One approach is to replace the standard DDIM with new invertible generation methods for image editing so that the initial noise for the generated image by those methods can be estimated~\cite{wallace2023edict, zhang2023exact}. However, they can not be used for the images generated by the standard DDIM. \citet{Pan_2023_ICCV} proposed an exact inversion method that can be applicable for DDIM-generated images, but it suffers a significant performance drop as the classifier-free guidance increases ($>1$) for enhancing image quality~\cite{ho2021classifierfree}. Note that all these prior works~\cite{wallace2023edict, zhang2023exact, Pan_2023_ICCV} can not be applicable for high-order DPM-solvers. The invertibility of DPM-solvers is an important theoretical property that could unlock a broader range of applications with DPMs just like the invertibility works for other generative models such as generative adversarial networks (GANs)~\cite{zhu2020domain,xia2022gan,Wang_2022_CVPR} and normalizing flows~\cite{rezende2015variational,papamakarios2021normalizing,pmlr-v139-whang21b}.

\begin{table}[!b]
\centering
\setlength{\tabcolsep}{8pt}
\begin{tabular}{@{~}c@{~~~}c@{~~~}c@{~}}
\hline
Order & Sampling ($T \rightarrow 0$) & Inversion ($0 \rightarrow T$) \\
\midrule
$1$ & 
\begin{tabular}{@{}c@{}}backward Euler\\ (-)\end{tabular} &  
\begin{tabular}{@{}c@{}}forward Euler\\ (na\"ive DDIM inversion)\end{tabular} \\
& & \\
$1$ & 
\begin{tabular}{@{}c@{}}forward Euler\\ (DDIM~\cite{song2020denoising})\end{tabular} & 
 
\begin{tabular}{@{}c@{}}backward Euler\\ (\cite{Pan_2023_ICCV}, Alg.~\ref{alg:inv_ddim})\end{tabular} \\
& & \\
$\geq 2$ & 
\begin{tabular}{@{}c@{}}linear multistep\\ (DPM-Solver++~\cite{lu2022dpm++})\end{tabular}
& 
\begin{tabular}{@{}c@{}}backward Euler with\\ high-order term \\   approximation (Alg. \ref{alg:2})
\end{tabular} \\
\hline
\end{tabular}
\caption{Summary of sampling and its corresponding inversion. Na\"ive DDIM inversion is not the corresponding inversion of DDIM, thus resulting in errors. For DDIM~\cite{song2020denoising}, our \cref{alg:inv_ddim} and the concurrent work~\cite{Pan_2023_ICCV} will be the corresponding inversion, but only ours can use classifier-free guidance $>1$ for stably enhancing quality. For DPM-Solver++(2M) with a linear multistep method~\cite{lu2022dpm++}, our \cref{alg:2} using the backward Euler with high-order term approximation will be the corresponding inversion.}
\label{tab:1}
\end{table}

In this work, we investigate the exact inversions for DDIM (first-order DPM-solver) as well as the faster high-order DPM-solvers.
For the standard DDIM with the forward Euler method, we propose the backward Euler method for its exact inversion, which is
an implicit technique to solve an optimization problem at each step (see \cref{alg:inv_ddim}).
For high-order DPM-solvers with linear multistep methods, exact inversion is more challenging since linear multistep methods rely on past states so their exact inversions require knowledge of unknown future states. To address this issue, we propose the backward Euler with approximate high-order terms as illustrated in Figure~\ref{fig:algabs} (see \cref{alg:2}).
Lastly, note that the na\"ive DDIM inversion is, in fact, the forward Euler method applied to the inversion.
\Cref{tab:1} summarizes the existing sampling and inversion methods as well as our contributions for them.
Then, we evaluate our proposed algorithms in various scenarios and applications such as reconstruction of images and noise in pixel-space DPM as well as LDM (\cref{sec:exp:recon}),
watermark detection and \emph{classification} (\cref{sec:exp:WM}) and the background-preserving image editing (\cref{sec:exp:edit}). While these experiments were the tasks from~\cite{wallace2023edict, wen2023tree, patashnik2023localizing}, the proposed methods significantly reduce reconstruction errors, thus enabling a new task like watermark \emph{classification} and allowing the background-preserving image editing without using any original latent vectors.
The contributions of this paper are:
\begin{itemize}
    \item proposing the exact inversion methods to find the initial noise of the images generated by various existing diffusion probabilistic models including high-order DPM-solvers by our proposed high-order term approximation,    
    \item implementing the backward Euler with either the gradient descent or the forward step method that enables exact inversion with large classifier-free guidance ($>1$) for enhancing image quality, and
    \item demonstrating that our exact inversion methods significantly reduce reconstruction errors for existing ODE-driven generation methods (DDIM, DPM-Solver++) in both image and latent spaces, better detect noise-space watermarks and even enable to classify which watermarks were used, and substantially improve background-preserving image editing.
\end{itemize}

\section{Related Work}

\paragraph{Diffusion probabilistic models:} DPMs 
are a class of generative models that iteratively denoise, ultimately generating original clean data. DPMs show notable advantages in generating diverse and high-quality image~\cite{dhariwal2021diffusion,ho2020denoising} (pixel-space DPM). In particular, latent diffusion models~\cite{rombach2022high} (LDMs) enable high-resolution image generation through latent space processes. Now, DPMs are widely applicable across various domains and applications such as 
image generation~\cite{ho2020denoising,song2020denoising,song2020improved,rombach2022high}, conditional image synthesis~\cite{ho2021classifierfree,rombach2022high,ramesh2022hierarchical,rombach2022high,ruiz2023dreambooth} and solving inverse problems~\cite{lugmayr2022repaint, li2022srdiff}.

\paragraph{Fast ODE solvers for DPMs:} The iterative denoising in DPMs usually takes a long sampling time~\cite{xiao2021tackling} and overcoming this drawback of DPM has been an active research area.
Early-stopping~\cite{lyu2022accelerating}, neural operator~\cite{zheng2023fast}, and progressive distillation~\cite{salimans2022progressive, meng2023distillation} can reduce sampling time, but require additional training. DDIM~\cite{song2020denoising}, DPM-Solver~\cite{lu2022dpm}, and DPM-Solver++~\cite{lu2022dpm++} formulate the denoising process of DPM as an ODE and then solve it using the forward Euler method or fast ODE solvers like exponential integrators to reduce the number of sampling steps from 1000 to 50 or 10 steps, respectively. Since these methods are training-free, they can be practically used with open-source DPMs~\cite{song2020denoising, rombach2022high}. 

\paragraph{Exact inversion methods:} Inversion has been important for various applications such as image editing~\citep{hertz2022prompt, Kim_2022_CVPR, patashnik2023localizing, wallace2023edict,su2022dual,zhang2023inversion}, model attacks~\cite{chen2023trojdiff}, watermark detection~\cite{wen2023tree} and image restoration~\cite{kawar2022denoising}. Exact inversions have been proposed beyond the na\"ive DDIM inversion. \citet{wallace2023edict} proposed a new sampling method, which performs exact diffusion inversion through invertible affine coupling transformations that alternately track and modify two separate quantities.
\citet{zhang2023exact} proposed bi-directional approximation integration to ensure symmetry between sampling and inversion algorithms. However, these prior exact inversion methods~\cite{wallace2023edict,zhang2023exact} proposed new sampling methods, thus exact inversions can be performed only for the images generated by these special methods, not for the images generated by the standard sampling methods such as DDIM.
Recently, \citet{Pan_2023_ICCV} proposed an exact inversion method with fixed point iterations (FPIs) for the standard DDIM-generated images. However, FPI sometimes does not converge, thus resulting in poor performance with the increased classifier-free guidance ($>1$) while strong classifier-free guidance was supposed to enhance image fidelity. 
For real image editing, performing the exact inversion of high-order DPM-solvers was not necessary since there is no true noise vector. However, there are other applications where accurate inversion is important.
\Cref{tab:related_work} summarizes the differences between those exact inversion methods.

\begin{table}
    \small
    \centering
    \setlength{\tabcolsep}{4pt}
    \begin{tabular}{cccc}
         \hline
         & \begin{tabular}{@{}c@{}}Standard \\ sampling \\ methods\end{tabular} & \begin{tabular}{@{}c@{}} Inversion of \\ high-order \\ DPM-solvers\end{tabular} & \begin{tabular}{@{}c@{}} Inversion with \\ classifier-free \\ guidance $> 1$\end{tabular} \\
         \midrule
         \begin{tabular}{@{}c@{}} \citet{wallace2023edict}\\ \citet{zhang2023exact} \end{tabular}& \xmark   & \xmark   & \cmark  \\
         \cmidrule{1-1}
         \citet{Pan_2023_ICCV} &  \cmark & \xmark & \xmark  \\
         \cmidrule{1-1}
         Ours & \cmark & \cmark & \cmark \\
         \hline
    \end{tabular}
    \caption{Property comparisons of exact inversion methods. }
    \label{tab:related_work}
\end{table}

\section{Background}

\subsection{Fast Sampling in DPM}
DDIM~\citep{song2020denoising}, DPM-solver~\citep{lu2022dpm}, and DPM-solver++~\citep{lu2022dpm++} are designed to recover $\vx_0 \in \R^D$ (image) from $\vx_T \in \R^D$ (noise), which is considered to have undergone the following diffusion process (gradually adding Gaussian noise) defined in $t \in [0,T]$:
\begin{equation}\label{eqn:conditional}
    q_{t0}(\vx_t|\vx_0) = \mathcal{N} ( \vx_t ; \alpha_t\vx_0, \sigma_t^2\mI),
\end{equation}
where $\alpha_t^2 / \sigma_t^2$, referred to the signal-to-noise ratio (SNR), is a strictly decreasing function of $t$~\citep{kingma2021variational}. Sampling $\vx_0$ can be done by solving the diffusion ODE, expressed as
\begin{equation}\label{eq:diffusion_ode_x0}
    \frac{d \vx_t}{d t} = \left(f(t) + \frac{g^2(t)}{2\sigma_t^2} \right)\vx_t - \frac{\alpha_t g^2(t)}{2\sigma^2_t}\vx_\vtheta(\vx_t,t),
\end{equation}
where $\vx_T\sim \mathcal{N} (\mathbf{0}, \tilde\sigma^2\mI)$, $f(t) \coloneqq \frac{d \log \alpha_t}{d t}$, $g^2(t) \coloneqq \frac{d \sigma_t^2}{dt} - 2\frac{d \log\alpha_t}{dt}\sigma_t^2$~\citep{kingma2021variational}. $\vx_\vtheta(\vx_t,t)$ is the data prediction model parameterized by learnable $\vtheta$, aiming to estimate $\vx_0$ from $\vx_t$.
Note that we employ the diffusion ODE defined with data prediction ($\vx_\vtheta$) rather than noise prediction ($\bm{\epsilon}_\vtheta$), as it is known to better perform in guided sampling at higher order~\cite{lu2022dpm++} (For the first order DDIM, they are equivalent).

\citet{lu2022dpm, zhang2022fast} have demonstrated that ODE solvers utilizing exponential integrators~\citep{hochbruck2010exponential} exhibit significantly faster convergence compared to conventional solvers when addressing \cref{eq:diffusion_ode_x0}. When provided with an initial value $\vx_s$ at time $s>0$, \citet{lu2022dpm++} derived the solution $\vx_t$ for the diffusion ODE (\cref{eq:diffusion_ode_x0}) at time $t$ using an exponential integrator as follows:
\begin{equation}
\label{eq:exact_solution_x0}
    \vx_t = \frac{\sigma_t}{\sigma_s}\vx_s + \sigma_t \int_{\lambda_s}^{\lambda_t} e^{\lambda} \vx_\vtheta(\vx_\lambda,\lambda)d\lambda,
\end{equation}
where $\vx_\lambda \coloneqq \vx_{t_\lambda(\lambda)}$ is the change-of-variable forms for the log-SNR ($\lambda$). $\lambda_t \coloneqq \log (\alpha_t / \sigma_t)$ is the inverse of $t_\lambda(\cdot)$.

Using the Taylor expansion at $\lambda_{t_{i-1}}$, DPM-Solver++ approximates the exact solution at time $t_i$, given $\vx_{t_{i-1}}$ at time $t_{i-1}$:
\begin{equation}
\label{eq:dpm_taylor_x0}
\begin{aligned}
    \vx_{t_i} &= \frac{\sigma_{t_i}}{\sigma_{t_{i-1}}}\vx_{t_{i-1}} + \sigma_{t_i}\sum_{n=0}^{k-1}
        \underbrace{%
            \vphantom{\int_{\lambda_{t_{i-1}}}^{\lambda_{t_i}} e^{\lambda}\frac{(\lambda - \lambda_{t_{i-1}})^n}{n!}d\lambda}
            \vx^{(n)}_\theta(\vx_{\lambda_{t_{i-1}}}, \lambda_{t_{i-1}})}_{\mathclap{\text{estimated}}
        }\\
        &\underbrace{\int_{\lambda_{t_{i-1}}}^{\lambda_{t_i}} e^{\lambda}\frac{(\lambda - \lambda_{t_{i-1}})^n}{n!}d\lambda}_{\mathclap{\text{analytically computed}}}
        + \underbrace{%
            \vphantom{\int_{\lambda_{t_{i-1}}}^{\lambda_{t_i}} e^{\lambda}\frac{(\lambda - \lambda_{t_{i-1}})^n}{n!}d\lambda}
            \gO(h_i^{k+1})}_{\mathclap{\text{omitted}}},
\end{aligned}
\end{equation}
where $h_i \coloneqq \lambda_{t_i} - \lambda_{t_{i-1}}$.
Since the integral part (w.r.t. $\lambda$) can be computed analytically and $\gO(h_i^{k+1})$ can be omitted, the only thing we need to find is $\vx^{(n)}_\theta(\vx_{\lambda_{t_{i-1}}}, \lambda_{t_{i-1}})$ for $n=0, \dots, k$.

The simplest approximation is $k=1$, and is equivalent to DDIM~\citep{song2020denoising} as follows:
\begin{equation}\label{eqn:ddim}
    \vx_{t_i} = \frac{\sigma_{t_i}}{\sigma_{t_{i-1}}}\vx_{t_{i-1}} - \alpha_{t_i}(e^{-h_i} - 1)\vx_\vtheta(\vx_{t_{i-1}}, t_{i-1}).
\end{equation}
For more precise approximation (hence for smaller number of steps), $k=2$ is a good choice:
\begin{equation}\label{eqn:dpm++2M}
\begin{aligned}
    \vx_{t_{i}} &= \frac{\sigma_{t_{i}}}{\sigma_{t_{i-1}}} \vx_{t_{i-1}} - \alpha_{t_{i}}\left(e^{-h_i} - 1\right)\biggl(\\ 
    &  \left(1 + \frac{1}{2r_i}\right)\vx_\theta(\vx_{t_{i-1}},t_{i-1}) 
        - \frac{1}{2r_i}\vx_\theta(\vx_{t_{i-2}},t_{i-2})\biggr).
\end{aligned}
\end{equation}
This is called as DPM-Solver++(2M)~\citep{lu2022dpm++}, where `2M' denotes second-order multistep. DPM-Solver++(2M) uses the previous value (\textit{i.e.}, $\vx_{t_{i-2}}$). Although DPM-Solver++(2M) enables fast sampling within only 10 to 20 steps, the nature of multistep methods becomes a tough obstacle for doing exact inversion. This will be covered in detail in \cref{sec:method-1}.

\subsection{Na\"ive DDIM inversion}
DDIM inversion implies obtaining $\vx_{t_{i-1}}$ given $\vx_{t_{i}}$, so $\vx_\vtheta(\vx_{t_{i-1}}, t_{i-1})$ as in \cref{eqn:ddim} is not explicitly obtainable (as $\vx_{t_{i-1}}$ is unknown yet).
To avoid the computational overhead of the implicit method, the na\"ive DDIM inversion takes the simplest way of using $\vx_\vtheta(\vx_{t_{i}}, t_{i-1})$ instead of $\vx_\vtheta(\vx_{t_{i-1}}, t_{i-1})$. Each step of the na\"ive DDIM inversion is expressed as follows:
\begin{equation}\label{eqn:naive_ddim_inv}
    \hat\vx_{t_{i-1}} = \frac{\sigma_{t_{i-1}}}{\sigma_{t_{i}}} \left( \vx_{t_{i}} + \alpha_{t_i}(e^{-h_i} - 1)\vx_\vtheta(\vx_{t_{i}}, t_{i-1}) \right).
\end{equation}
This method can be interpreted as another forward Euler method starting from $t=0$; hence this is the exact inversion of sampling via the backward Euler, as shown in \cref{tab:1}. 
Nevertheless, the na\"ive DDIM inversion is widely used for many applications such as image editing~\citep{hertz2022prompt,Kim_2022_CVPR} as they have short runtimes.

\
\section{Proposed Method}\label{sec:proposed method}

\subsection{Exact Inversion of DDIM}\label{sec:method-1}

\paragraph{Backward Euler method:} We employ the backward Euler method for exact inversion of DDIM. \Cref{alg:inv_ddim} shows the proposed exact inversion of DDIM. For initialization, we perform the na\"ive DDIM inversion (line \ref{alg1:naiveDDIM} of \cref{alg:inv_ddim}). For iterations (lines \ref{alg1:correctorstart}-\ref{alg1:correctorend} and \Call{Update}{$\hat\vz_{t_{i-1}}; \hat\vz_{t_{i}}, \vz'_{t_{i}}$} of \cref{alg:inv_ddim}), we use either gradient descent:
\begin{equation*}
\text{Taking gradient step on }\nabla_{\hat\vz_{t_{i-1}}} \lVert \hat\vz_{t_{i}} - \vz'_{t_{i}} \rVert_2^2,
\end{equation*}
or the forward step method:
\begin{equation*}
    \hat\vz_{t_{i-1}} = \hat\vz_{t_{i-1}} - \rho (\vz'_{t_{i}} - \hat\vz_{t_{i}}),
\end{equation*}
where $\vz'_{t_{i}} \gets \frac{\sigma_{t_{i}}}{\sigma_{t_{i-1}}} \hat\vz_{t_{i-1}} - \alpha_{t_{i}}(e^{-h_i} - 1)\vz_\theta(\hat\vz_{t_{i-1}},t_{i-1})$.

\begin{algorithm}[!b]
    \small
    \centering
    \caption{Inversion of DDIM via the backward Euler.}\label{alg:inv_ddim}
    \begin{algorithmic}[1]
    \Require initial value $\vx$, time steps $\{t_i\}_{i=0}^M$, data prediction model $\vz_\vtheta$, \Call{Update}{}, \Call{$\gD^\dagger$}{} in \cref{sec:method-1}.
        \State Denote $h_i \coloneqq \lambda_{t_{i}} - \lambda_{t_{i-1}}$
        for $i=1,\dots,M$.
        \State $\hat\vz_{t_M}\gets \gD^\dagger(\vx_0)$ \textbf{if} LDM \textbf{else} $\vx_0$
        \For{$i\gets M$ to $1$}
        \State  $\hat\vz_{t_{i-1}} \gets \frac{\sigma_{t_{i-1}}}{\sigma_{t_{i}}} \left( \hat\vz_{t_{i}} + \alpha_{t_{i}}\left(e^{-h_i} - 1\right)\vz_\theta(\hat\vz_{t_{i}},t_{i-1})\right)$ \label{alg1:naiveDDIM} 
        \Repeat \label{alg1:correctorstart}
        \State  $\vz'_{t_{i}} \gets \frac{\sigma_{t_{i}}}{\sigma_{t_{i-1}}} \hat\vz_{t_{i-1}} - \alpha_{t_{i}}(e^{-h_i} - 1)\vz_\theta(\hat\vz_{t_{i-1}},t_{i-1})$ \label{alg1:naiveDDI}
        \State \Call{Update}{$\hat\vz_{t_{i-1}}; \hat\vz_{t_{i}}, \vz'_{t_{i}}$ }
        \Until converged \label{alg1:correctorend}
        \EndFor
        \State \Return $\hat\vz_{t_0}$
    \end{algorithmic}
\end{algorithm}

\paragraph{Gradient descent or the forward step method vs FPI:} One may try employing FPI rather than gradient descent or the forward step method. However, in \cite{Pan_2023_ICCV}, it is observed that the accuracy of reconstruction (measured by LPIPS and SSIM) significantly decreases when the classifier-free guidance $\omega$ is larger than $1$. In this paragraph, we briefly explain why FPI is vulnerable to large classifier-free guidance. In our setting (\cref{eqn:ddim}), the FPI operator $F$ can be defined as: 
\begin{equation}
    F(\cdot) := \frac{\sigma_{t_{i-1}}}{\sigma_{t_i}} \alpha_{t_i}(e^{-h_i} - 1)\vx_\vtheta(\cdot, t_{i-1}) + \frac{\sigma_{t_{i-1}}}{\sigma_{t_i}}\hat\vx_{t_i}.
\end{equation}
To ensure the convergence of FPI, at the very least, $F$ needs to be nonexpansive, and a sufficient condition for being nonexpansive is that $\vx_\vtheta(\cdot, t_{i-1})$ is $(\nicefrac{\sigma_{t_{i-1}}\alpha_{t_i}(e^{-h_i} - 1)}{\sigma_{t_i}} )^{-1}$-Lipschitz continuous. Considering the classifier-free guidance $\omega > 1$, the model should be $(|\omega|+|1-\omega|)^{-1}(\nicefrac{\sigma_{t_{i-1}}\alpha_{t_i}(e^{-h_i} - 1)}{\sigma_{t_i}} )^{-1}$-Lipschitz continuous.
This suggests that the inversion via FPI is likely to fail when the classifier-free guidance $\omega$ is large. In contrast, 
the forward step method (gradient descent) can adjust step sizes (learning rates). When the step size is reduced, it takes more time to converge, but is more likely to converge. This property enhances the robustness of our approach with large classifier-free guidance (\cref{sec:exp}). In fact, it is widely known that gradient descent or the forward step method is more stable than FPI~\cite{ryu2022large}. 

\paragraph{Decoder inversion:}
As LDMs use latent variables in the diffusion process, they necessarily require a decoder ($\gD$) that can convert latent variable ($\vz_0$) to image ($\vx_0$). Previous studies~\cite{wallace2023edict, Pan_2023_ICCV} used the encoder ($\gE$) for the inversion of the decoder. However, since the encoder is not the exact inverse of the decoder, it induces reconstruction errors (so \cite{wallace2023edict, Pan_2023_ICCV} set $\lVert \gD (\gE (\vx_0)) - \vx_0\rVert$ as a lower bound for reconstruction errors). 
For reducing this error, we perform the exact inversion of the \emph{decoder}. As in many GAN inversion studies~\cite{abdal2019image2stylegan, abdal2020image2stylegan++, xia2022gan}, we employ the gradient descent as: 
\begin{algorithm}[!h]
    \small
    \centering
    \begin{algorithmic}[1]
    \Function{$\gD^\dagger$}{$\vx$}\label{dec_inv} \hfill \textit{// \small Decoder inversion}
         \State $\vz \gets \gE(\vx)$
         \Repeat \text{ gradient} step on $\nabla_{\vz} \lVert \vx - \gD(\vz) \rVert_2^2 $
         \Until converged
         \State \Return $\vz$
    \EndFunction 
    \end{algorithmic}
\end{algorithm}

\noindent We use \cref{alg:inv_ddim} in \cref{sec:exp:recon} and \ref{sec:exp:edit}.

\subsection{Exact Inversion of High-order DPM-Solvers}
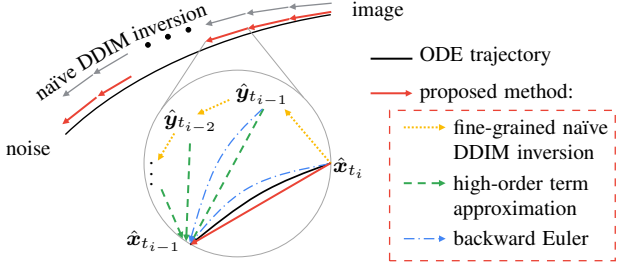
\begin{figure}[!t]
    \centering
    \footnotesize
    \begin{tikzpicture}[  every node/.style={ 
    font=\footnotesize,          
    text width=\linewidth,      
    align=left, anchor=south west        
  }]
    \node[anchor=south west, inner sep=0] (image) at (0,0) {\includegraphics[width=\linewidth]{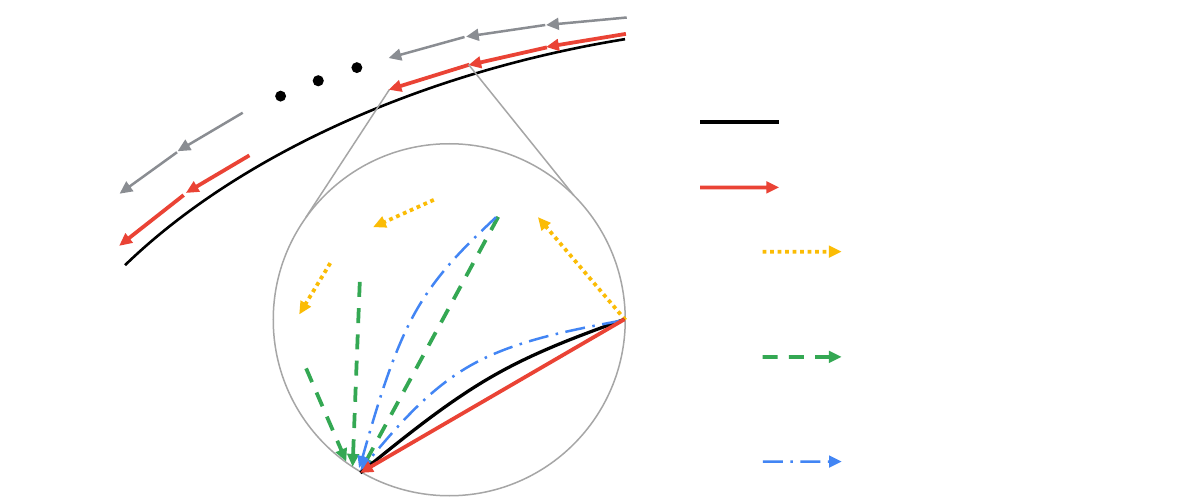}};
    \node[] at (3.05,1.95) {$\hat\vy_{t_{i-1}}$};
    \node[] at (2.1,1.6) {$\hat\vy_{t_{i-2}}$};
    \node[] at (1.6,0.0) {$\hat\vx_{t_{i-1}}$};
    \node[] at (4.35,1.) {$\hat\vx_{t_{i}}$};
    \node[] at (1.9, 0.9) {$\vdots$};

    \node[] at (0.,1.3) {noise};
    \node[] at (4.6,3.1) {image};
    \node[rotate=35, inner sep=0, text width=0cm] at (0.55,2.1) {na\"ive};
    \node[rotate=30, inner sep=0, text width=0cm] at (1.1,2.5) {DDIM};
    \node[rotate=20, inner sep=0, text width=0cm] at (1.8,2.9) {inversion};

    \node[] at (5.5,2.5) {ODE trajectory};
    \node[] at (5.5,2.0) {proposed method:};
    \node[] at (5.95, 1.25) {fine-grained na\"ive \\ DDIM inversion};
    \node[] at (5.95, 0.45) {high-order term \\ approximation};
    \node[] at (5.95, 0.1) {backward Euler};
    \definecolor{myred}{RGB}{234, 67, 53}
    \draw[draw=myred, dashed, line width=0.5pt] (5.2,0) rectangle (8.2,2);
    
    \end{tikzpicture}
\caption{An abstract of our algorithm for exact inversion of high-order DPM-solvers. Since $\hat\vx_{t_{i-1}}, \hat\vx_{t_{i-2}}, \dots$ are needed for high-order terms but unobtainable, we estimate them via the fine-grained na\"ive DDIM inversion ($\hat\vy_{t_{i-1}}, \hat\vy_{t_{i-2}}, \dots$). Then we use the backward Euler method with high-order term approximation.}
\label{fig:algabs}
\end{figure}

In this subsection, we propose an exact inversion method for high-order DPM-solvers. Our motivation for this idea is that values prior to $t_{i-1}$ (\textit{i.e.}, $\vx_{t_{i-2}}, \vx_{t_{i-3}}, \dots$), which cannot be estimated at the current time, have been used for higher-order terms in \cref{eq:dpm_taylor_x0}, \textit{i.e.}, 
\begin{equation}\label{eqn:high-order}
\sigma_{t_i}\sum_{n=1}^{k-1}
            \vphantom{\int_{\lambda_{t_{i-1}}}^{\lambda_{t_i}} e^{\lambda}\frac{(\lambda - \lambda_{t_{i-1}})^n}{n!}d\lambda}
            \vx^{(n)}_\theta(\vx_{\lambda_{t_{i-1}}}, \lambda_{t_{i-1}}) \int_{\lambda_{t_{i-1}}}^{\lambda_{t_i}} e^{\lambda}\frac{(\lambda - \lambda_{t_{i-1}})^n}{n!}d\lambda.  
\end{equation}
Their impact on the overall computation is expected to be relatively small. So we estimate these values (\textit{i.e.}, $\vx_{t_{i-1}}, \vx_{t_{i-2}}, \dots$) using a slightly less precise method, such as the na\"ive DDIM inversion with a finer step size (the yellow lines in \cref{fig:algabs}). After that, we find $\hat\vx_{t_{i-1}}$ by the backward Euler method (the blue lines in \cref{fig:algabs}), as the high-order terms (\cref{eqn:high-order}) are treated as constant (the green lines in \cref{fig:algabs}). \Cref{fig:algabs} shows an abstract of our algorithm for exact inversion of forward linear multistep methods.

To illustrate this with DPM-Solver++(2M) (\cref{eqn:dpm++2M}), we provide \cref{alg:2}. 
Using our key idea, we first obtain $\hat\vy_{t_{i-1}}$ and $\hat\vy_{t_{i-2}}$ as substitutes for $\hat\vx_{t_{i-1}}$ and $\hat\vx_{t_{i-2}}$ using a fine-grained na\"ive DDIM inversion. Then we use $\hat\vy_{t_{i-1}}$ and $\hat\vy_{t_{i-2}}$ to find $\hat\vx_{t_{i-1}}$ via the backward Euler method with high-order term approximation as follows:
\begin{equation} \label{eq:alg2-1}
    \vd'_i \gets 
        \vz_\theta(\hat\vz_{t_{i-1}},t_{i-1}) + \underbrace{\frac{\vz_\theta(\hat\vy_{t_{i-1}},t_{i-1}) 
        - \vz_\theta(\hat\vy_{t_{i-2}},t_{i-2}) }{2 r_i}}_{\mathclap{\text{high-order term approximation}}},
\end{equation}
where $r_i =  \frac{ \lambda_{t_{i-1}} - \lambda_{t_{i-2}}}{\lambda_{t_{i}} - \lambda_{t_{i-1}}}$,
and these operations are repeated until convergence is achieved. We employ \cref{alg:2} in \cref{sec:exp:recon} and \ref{sec:exp:WM}.

\begin{algorithm}[h]
    \small
    \centering
    \caption{Inversion of DPM-Solver++(2M)}\label{alg:2}
    \begin{algorithmic}[1]
    \Require initial value $\vx_0$, time steps $\{t_i\}_{i=0}^M$, data prediction model $\vz_\vtheta$, \Call{Update}{}, \Call{$\gD^\dagger$}{} in \cref{sec:method-1}.
        \State Denote $h_i \coloneqq \lambda_{t_{i}} - \lambda_{t_{i-1}}$
        and $r_i \coloneqq \frac{h_{i-1}}{h_{i}}$ 
        for $i=1,\dots,M$.
        \State $\hat\vz_{t_M}\gets \gD^\dagger(\vx_0)$ \textbf{if} LDM \textbf{else} $\vx_0$
        \For{$i\gets M$ to $2$}
        $\hat\vy_{t_{i}} \gets \hat\vz_{t_{i}}$
        \For{$j\gets 1$ to $2J$} \label{alg2:finerstart}
        \State  $\hat\vy_{t_{i-\nicefrac{j}{J}}}
        \gets \frac{\sigma_{t_{i-\nicefrac{j}{J}}}}{\sigma_{t_{i-\nicefrac{(j-1)}{J}}}} ( \hat\vy_{t_{i-\nicefrac{j}{J}}} + \alpha_{t_{i-\nicefrac{j}{J}}}(e^{-h_{i-\nicefrac{j}{J}}} - 1)\vz_\theta(\hat\vy_{t_{i-\nicefrac{(j-1)}{J}}},t_{i-\nicefrac{j}{J}}))$ 
        \EndFor \label{alg2:finerend}
        \State  $\hat\vz_{t_{i-1}} \gets \hat\vy_{t_{i-1}}$ \label{alg2:Eulerstart}
        \Repeat 
        \State $\vd'_i \gets 
        \vz_\theta(\hat\vz_{t_{i-1}},t_{i-1}) + \frac{1}{2 r_i}(\vz_\theta(\hat\vy_{t_{i-1}},t_{i-1}) 
        - \vz_\theta(\hat\vy_{t_{i-2}},t_{i-2}) )$ 
        \State  $\vz'_{t_{i}} \gets \frac{\sigma_{t_{i}}}{\sigma_{t_{i-1}}} \hat\vz_{t_{i-1}} - \alpha_{t_{i}}\left(e^{-h_i} - 1\right)\vd'_i$
        \State \Call{Update}{$\hat\vz_{t_{i-1}}; \hat\vz_{t_{i}}, \vz'_{t_{i}}$ }
        \Until converged \label{alg2:Eulerend}
        \EndFor
        \State  $\hat\vz_{t_{0}} \gets \frac{\sigma_{t_{0}}}{\sigma_{t_{1}}} \left( \hat\vz_{t_{1}} + \alpha_{t_{1}}\left(e^{-h_1} - 1\right)\vz_\theta(\hat\vz_{t_{1}},t_{0})\right)$ 
        \Repeat 
        \State  $\vz'_{t_{1}} \gets \frac{\sigma_{t_{1}}}{\sigma_{t_{0}}} \hat\vz_{t_{0}} - \alpha_{t_{1}}\left(e^{-h_1} - 1\right)\vz_\theta(\hat\vz_{t_{0}},t_{0})$ 
        \State \Call{Update}{$\hat\vz_{t_{0}}; \hat\vz_{t_{1}}, \vz'_{t_{1}}$ }
        \Until converged
        \State \Return $\hat\vz_{t_0}$
    \end{algorithmic}
\end{algorithm}

\
\section{Experiments}\label{sec:exp}
\subsection{Reconstruction}\label{sec:exp:recon}
In this subsection, we perform the reconstruction of noise and image to evaluate the exact invertibility of the proposed methods. For simplicity, let $\vx_0 = \mathrm{DPM} (\vx_T)$. Let $\mathrm{DPM}^\dagger: \mathbb{R}^D \rightarrow \mathbb{R}^D$ be the inversion of $\mathrm{DPM}$. Let $\hat\vx_T = \mathrm{DPM}^\dagger (\vx_0)$ and $\hat\vx_0 = \mathrm{DPM}(\hat\vx_T)$. Exact inversion of \emph{noise} refers to $\vx_T = \hat\vx_T$, and thus, the goal is to minimize $\mathrm{NMSE}(\vx_T, \hat\vx_T) = \lVert \vx_T - \hat\vx_T \rVert_2^2 / \lVert \vx_T \rVert_2^2$. Similarly, exact inversion of the \emph{image} refers to $\vx_0 = \hat\vx_0$, and the objective is to minimize $\mathrm{NMSE}(\vx_0, \hat\vx_0)$.
For practical utility, we used LDM~\cite{rombach2022high} with the classifier-free guidance $\omega=3.0$. To evaluate algorithm performance independently, unaffected by decoder inversion or classifier-free guidance, we also use an unconditional pixel-space DPM~\cite{song2020denoising} trained on the ImageNet64 dataset\footnote{https://github.com/LuChengTHU/dpm-solver/tree/main/examples/ddpm\_and\_guided-diffusion}.

\begin{figure*}[t]
    \setlength{\tabcolsep}{2pt}
    \centering
    \begin{tabular}{@{}c@{ }c@{ }c@{}c@{}c@{}c@{}}
    \multicolumn{1}{@{}c@{}}{\multirow{2}{*}{\rotatebox[origin=c]{90}{\centering \small Generation}}}& \multirow{2}{*}{\rotatebox[origin=c]{90}{\small Inversion}} & \multicolumn{2}{@{}c@{ }}{Pixel-space DPM} & \multicolumn{2}{c}{LDM}\\
    \cmidrule(l){3-4} \cmidrule(l){5-6}
        &            & \begin{tabular}{@{}c@{}} Image \\ \small (Recon. / Error $\times 2$) \end{tabular}  & \begin{tabular}{@{}c@{}} Noise \\ \small (Recon. / Error $\times 2$) \end{tabular} & \begin{tabular}{@{}c@{}} Image \\ \small (Recon. / Error $\times 2$) \end{tabular}  & \begin{tabular}{@{}c@{}} Noise \\ \small (Recon. / Error $\times 2$) \end{tabular}\\
     \cmidrule(l){3-4} \cmidrule(l){5-6}
     \multirow{2}{*}{\rotatebox[origin=c]{90}{ \small DDIM 50 steps}} &   \cellcolor{bGray}\rotatebox[origin=c]{90}{\small na\"ive / 1000}                        & \begin{tabular}{c}\includegraphics[trim=68 0 0 0, clip, width=0.230\linewidth]{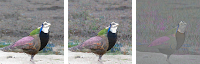}\end{tabular}      & \begin{tabular}{c}\includegraphics[trim=68 0 0 0, clip, width=0.230\linewidth]{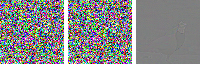}\end{tabular} & \begin{tabular}{c}\includegraphics[trim=544 0 0 0, clip, width=0.230\linewidth]{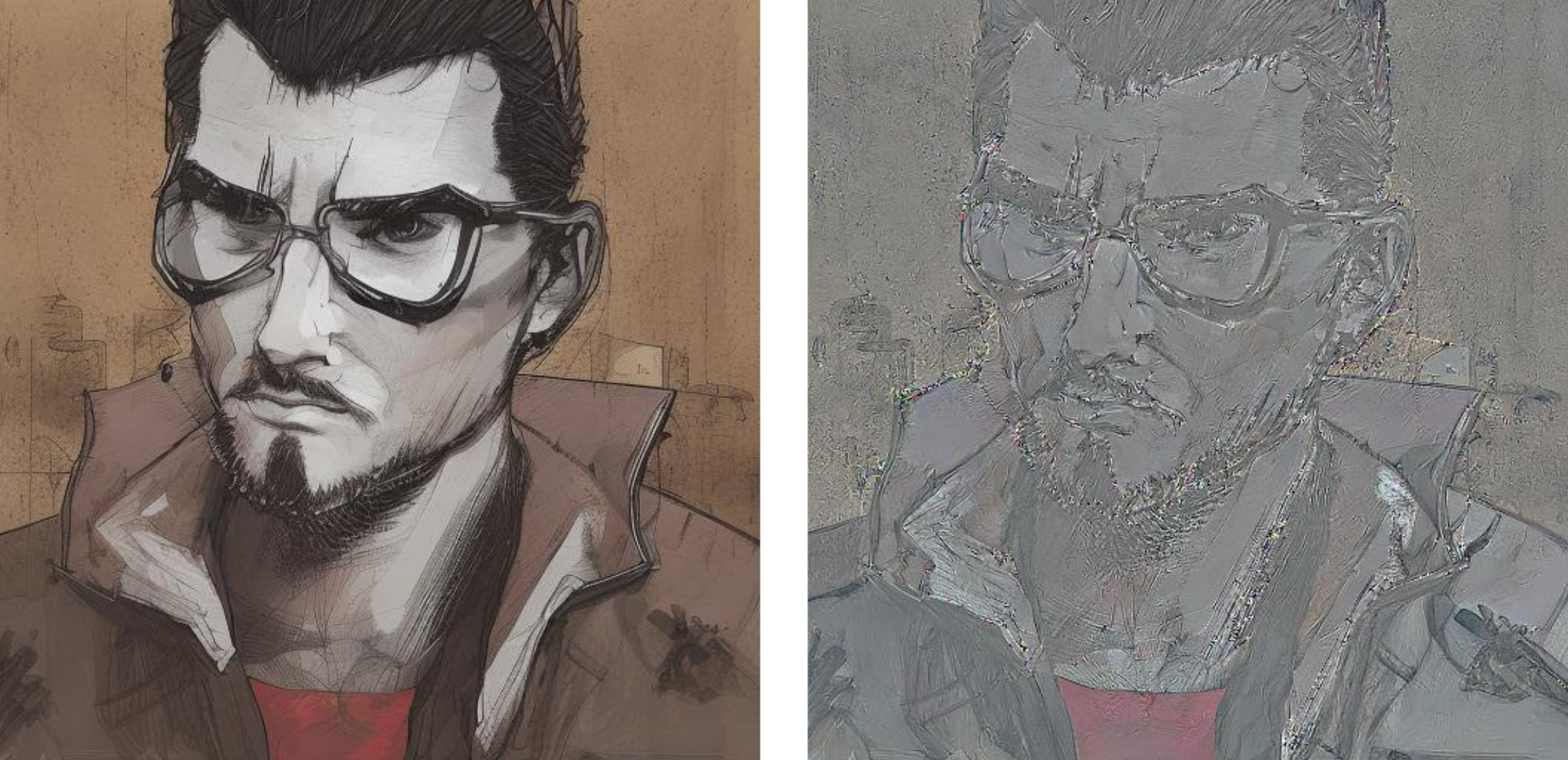}\end{tabular}      & \begin{tabular}{c}\includegraphics[trim=544 0 0 0, clip, width=0.230\linewidth]{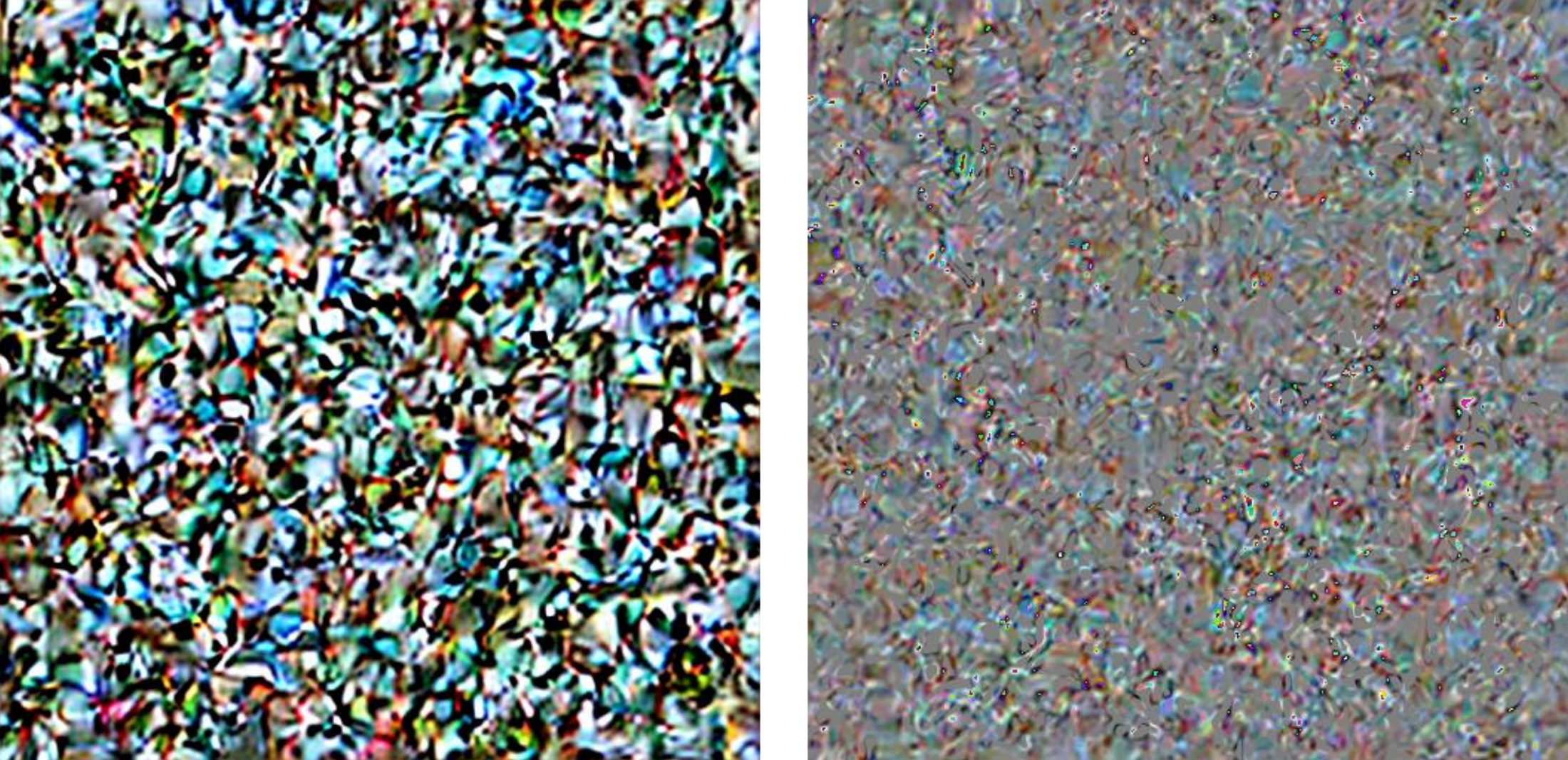}\end{tabular}     \\
     & \cellcolor{bBlue}\rotatebox[origin=c]{90}{\small Alg. \ref{alg:inv_ddim} / 50}                          & \begin{tabular}{c}\includegraphics[trim=68 0 0 0, clip, width=0.230\linewidth]{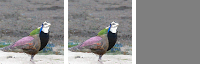}\\[-\dp\strutbox]\end{tabular}      & \begin{tabular}{c}\includegraphics[trim=68 0 0 0, clip, width=0.230\linewidth]{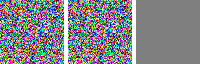}\\[-\dp\strutbox]\end{tabular}  & \begin{tabular}{c}\includegraphics[trim=544 0 0 0, clip, width=0.230\linewidth]{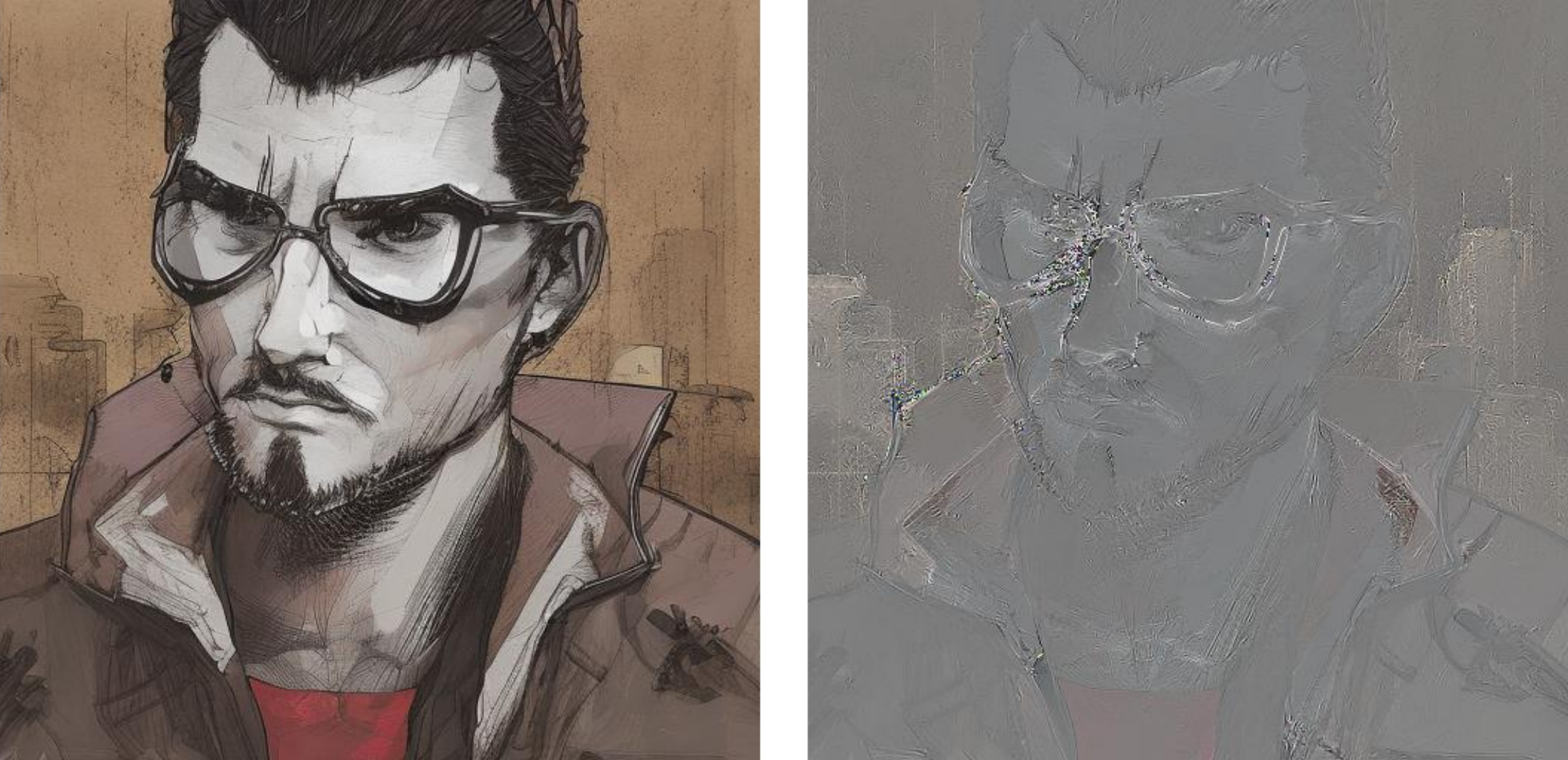}\\[-\dp\strutbox]\end{tabular}      & \begin{tabular}{c}\includegraphics[trim=544 0 0 0, clip, width=0.230\linewidth]{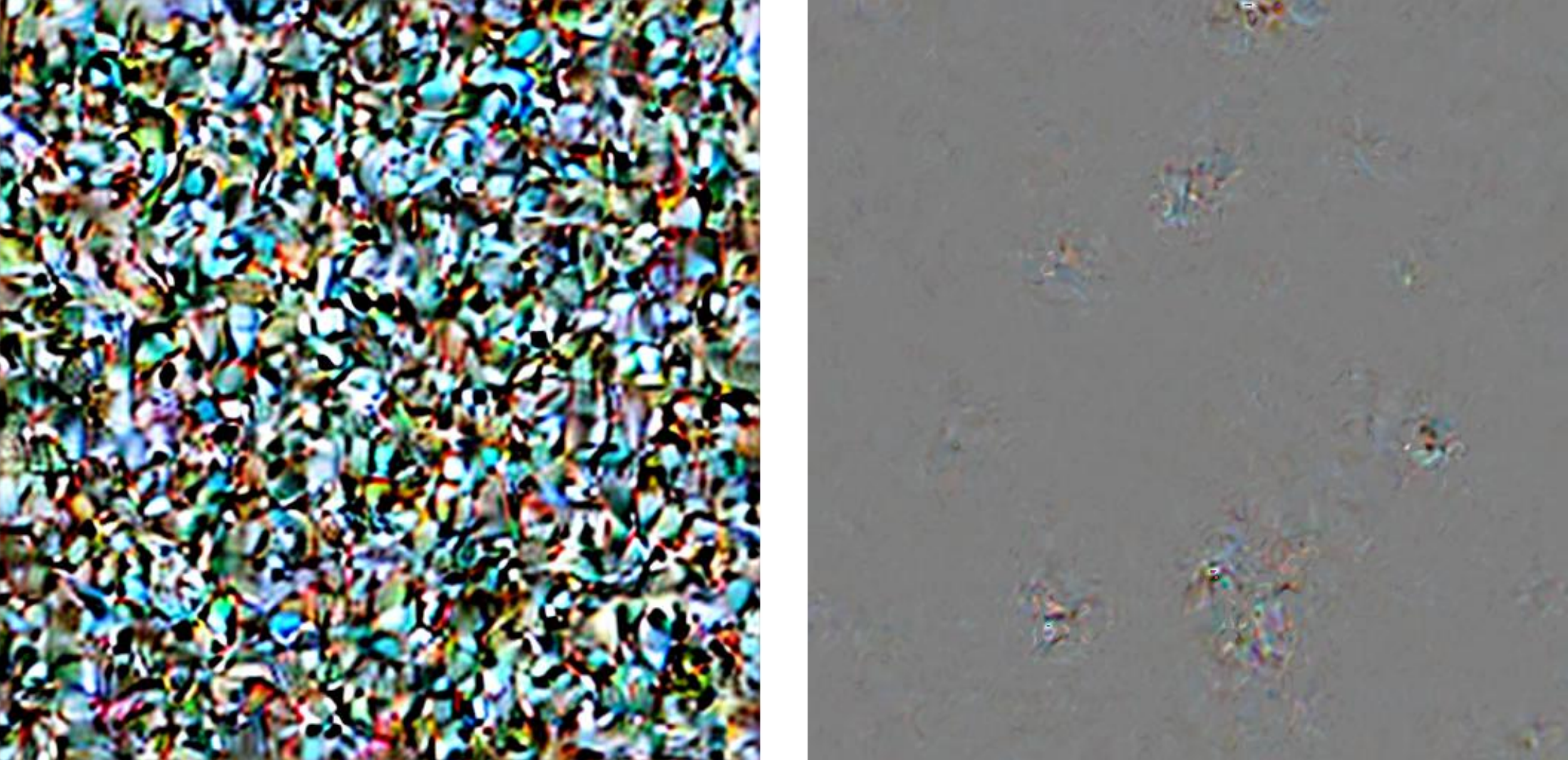}\\[-\dp\strutbox]\end{tabular}     \\
      \cmidrule(l){3-4} \cmidrule(l){5-6}
     \multirow{3}{*}{\rotatebox[origin=c]{90}{\small DPM-Solver++(2M) 10 steps}} & \rotatebox[origin=c]{90}{
          
     \cellcolor{bGray}\small na\"ive / 1000 }         & \begin{tabular}{c}\includegraphics[trim=68 0 0 0, clip, width=0.230\linewidth]{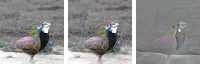}\end{tabular}      & \begin{tabular}{c}\includegraphics[trim=68 0 0 0, clip, width=0.230\linewidth]{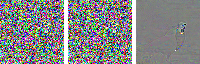}\end{tabular}   & \begin{tabular}{c}\includegraphics[trim=544 0 0 0, clip, width=0.230\linewidth]{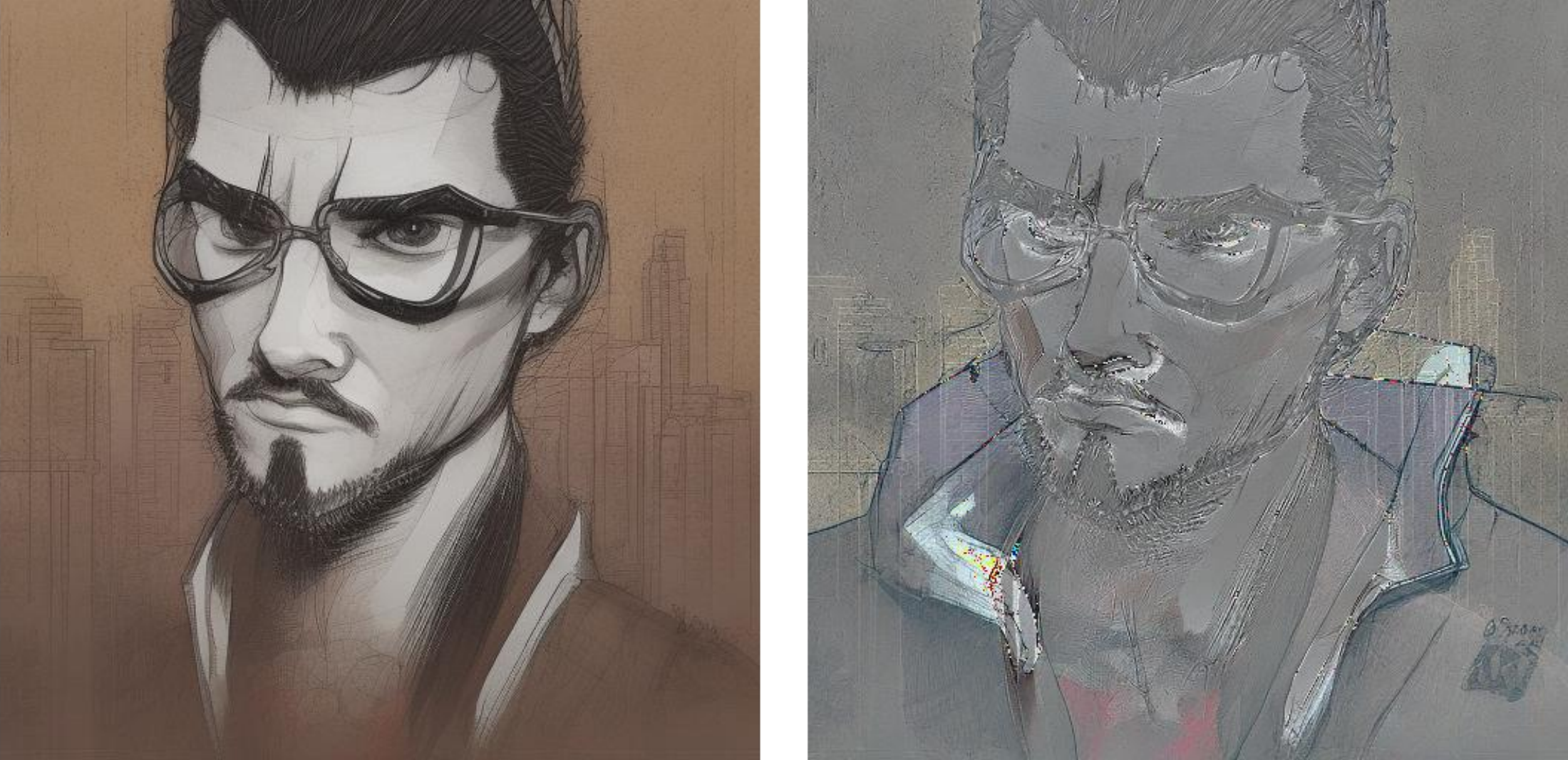}\end{tabular}      & \begin{tabular}{c}\includegraphics[trim=544 0 0 0, clip, width=0.230\linewidth]{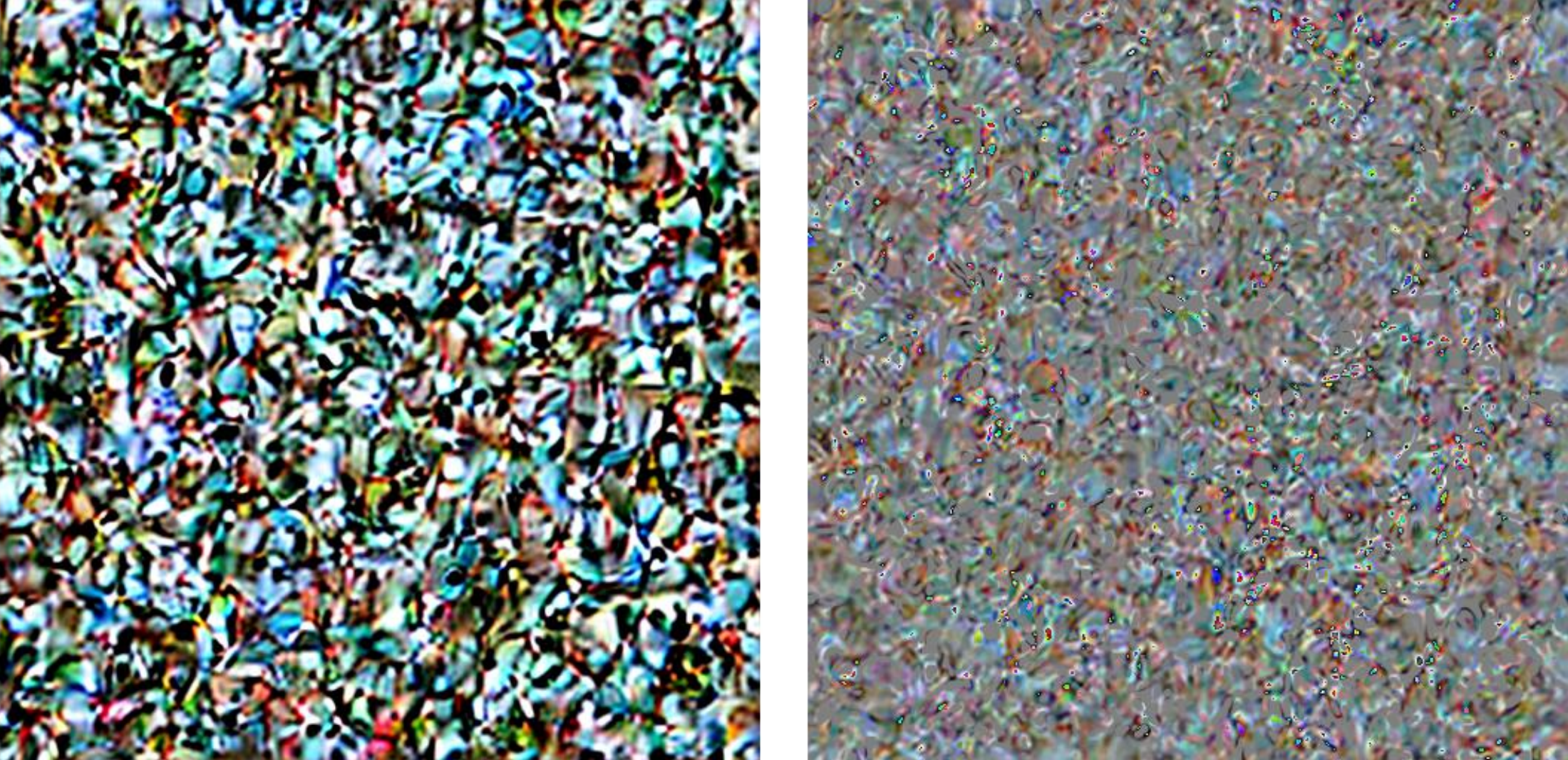}\end{tabular}    \\
     & \cellcolor{bBlue}{\rotatebox[origin=c]{90}{\small Alg. \ref{alg:inv_ddim} / 50}}                              & \begin{tabular}{c}\includegraphics[trim=68 0 0 0, clip, width=0.230\linewidth]{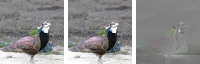}\end{tabular}      & \begin{tabular}{c}\includegraphics[trim=68 0 0 0, clip, width=0.230\linewidth]{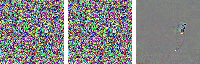}\end{tabular}  & \begin{tabular}{c}\includegraphics[trim=544 0 0 0, clip, width=0.230\linewidth]{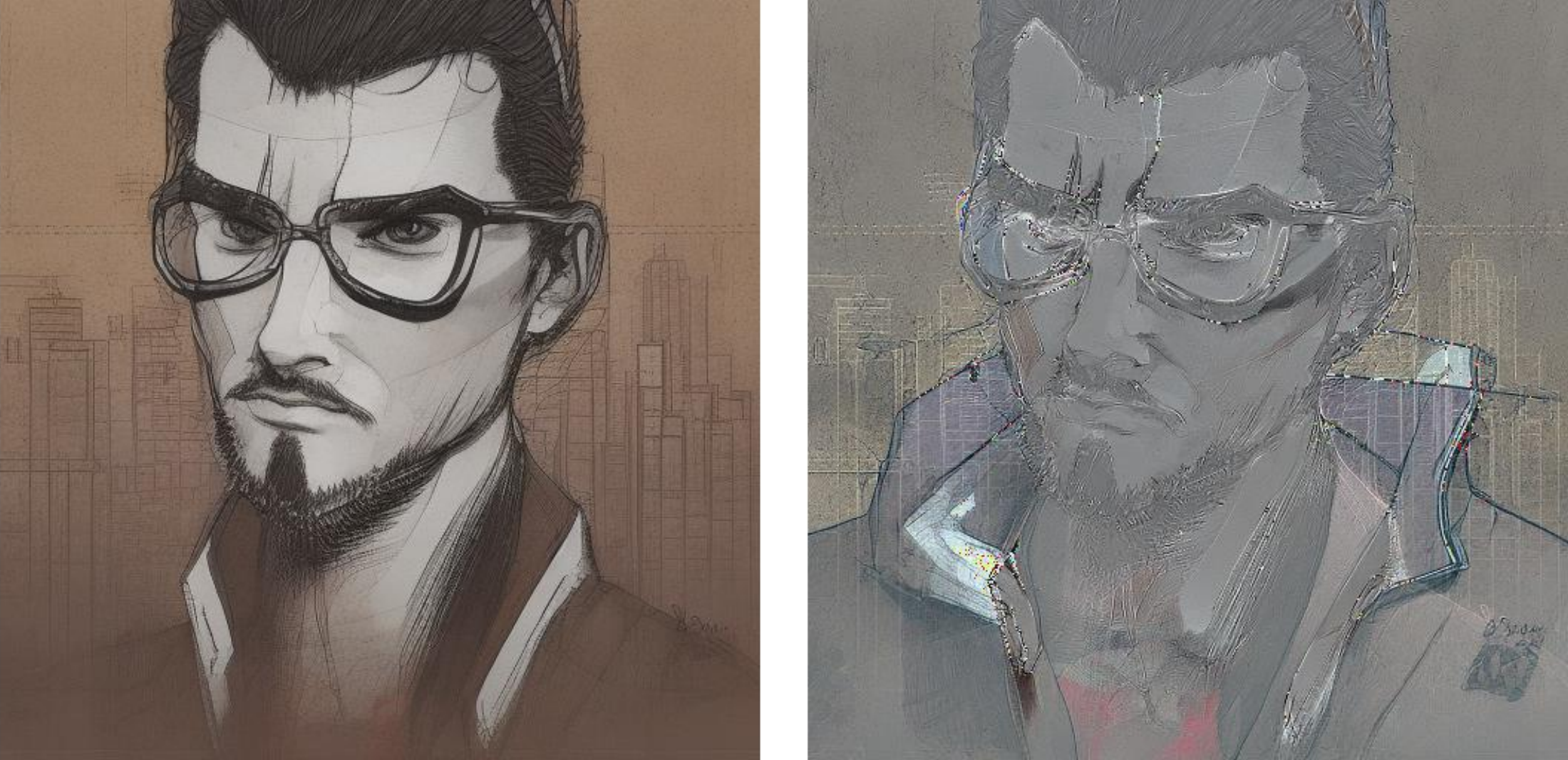}\end{tabular}      & \begin{tabular}{c}\includegraphics[trim=544 0 0 0, clip, width=0.230\linewidth]{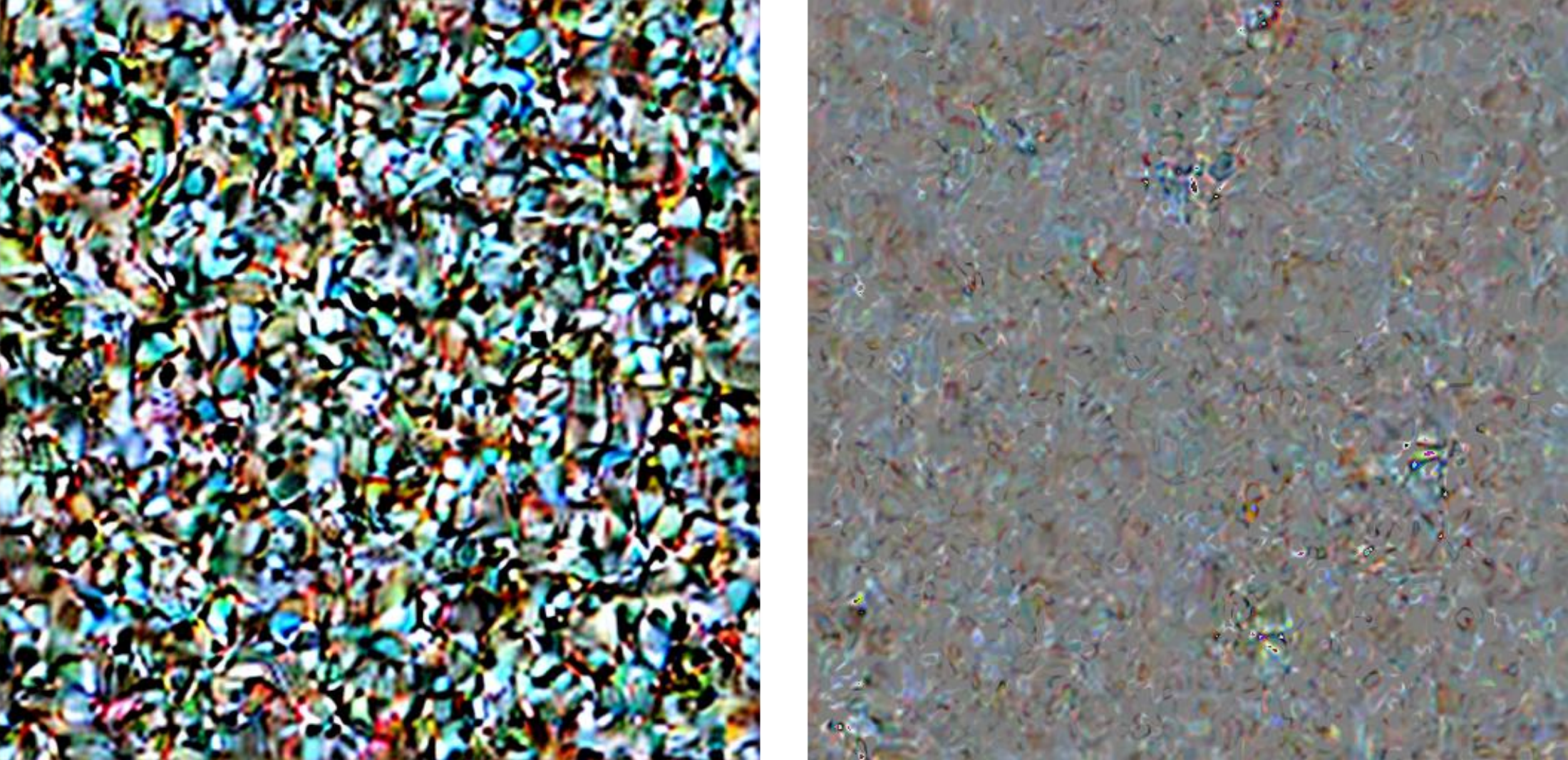}\end{tabular}     \\ 
     & \rotatebox[origin=c]{90}{\cellcolor{bRed}{\small Alg. \ref{alg:2} / 10}}                             & \begin{tabular}{c}\includegraphics[trim=68 0 0 0, clip, width=0.230\linewidth]{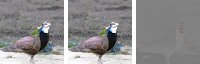}\\[-\dp\strutbox]\end{tabular}      & \begin{tabular}{c}\includegraphics[trim=68 0 0 0, clip, width=0.230\linewidth]{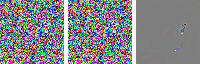}\\[-\dp\strutbox]\end{tabular}    & \begin{tabular}{c}\includegraphics[trim=544 0 0 0, clip, width=0.230\linewidth]{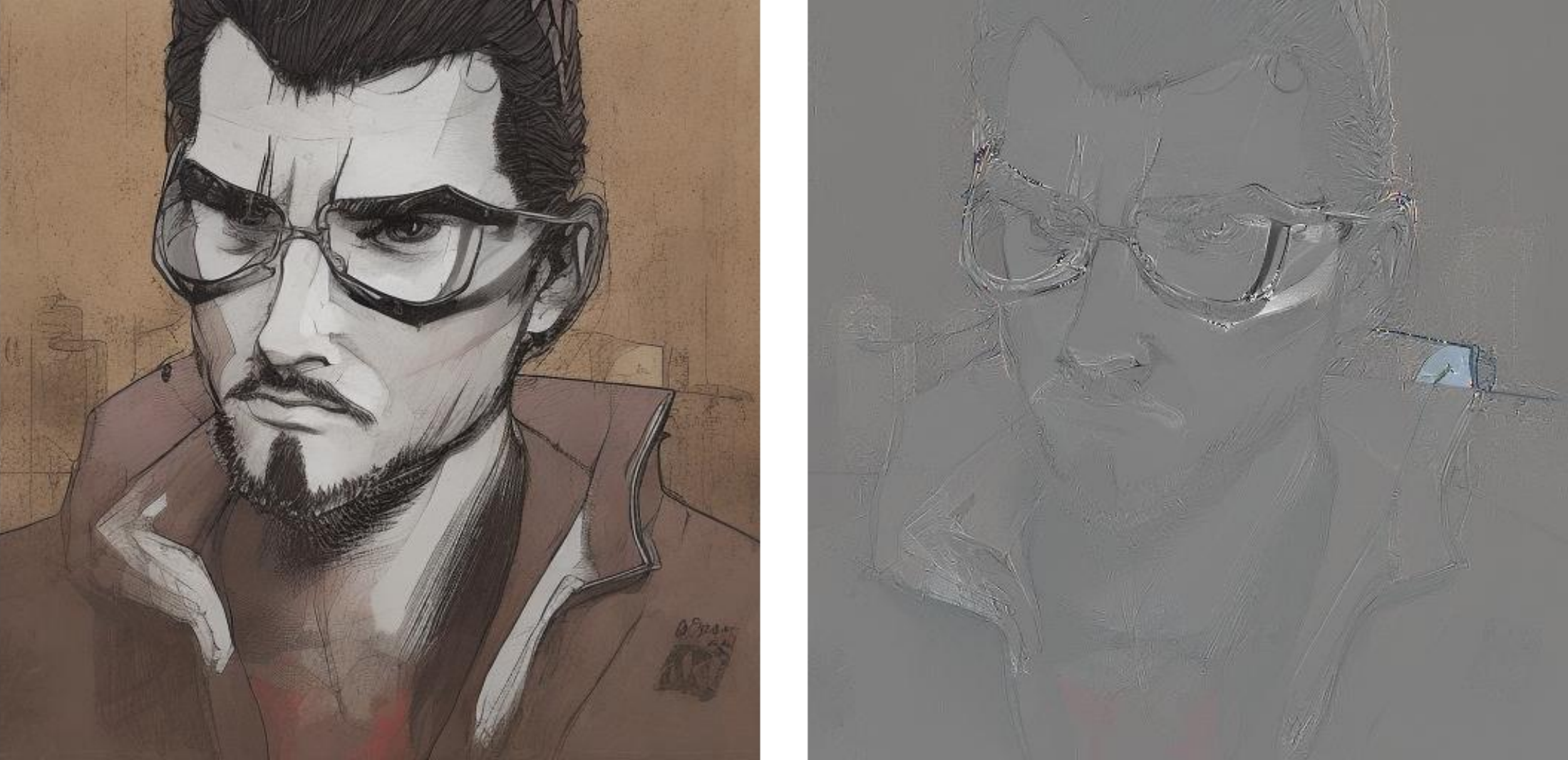}\\[-\dp\strutbox]\end{tabular}      & \begin{tabular}{c}\includegraphics[trim=544 0 0 0, clip, width=0.230\linewidth]{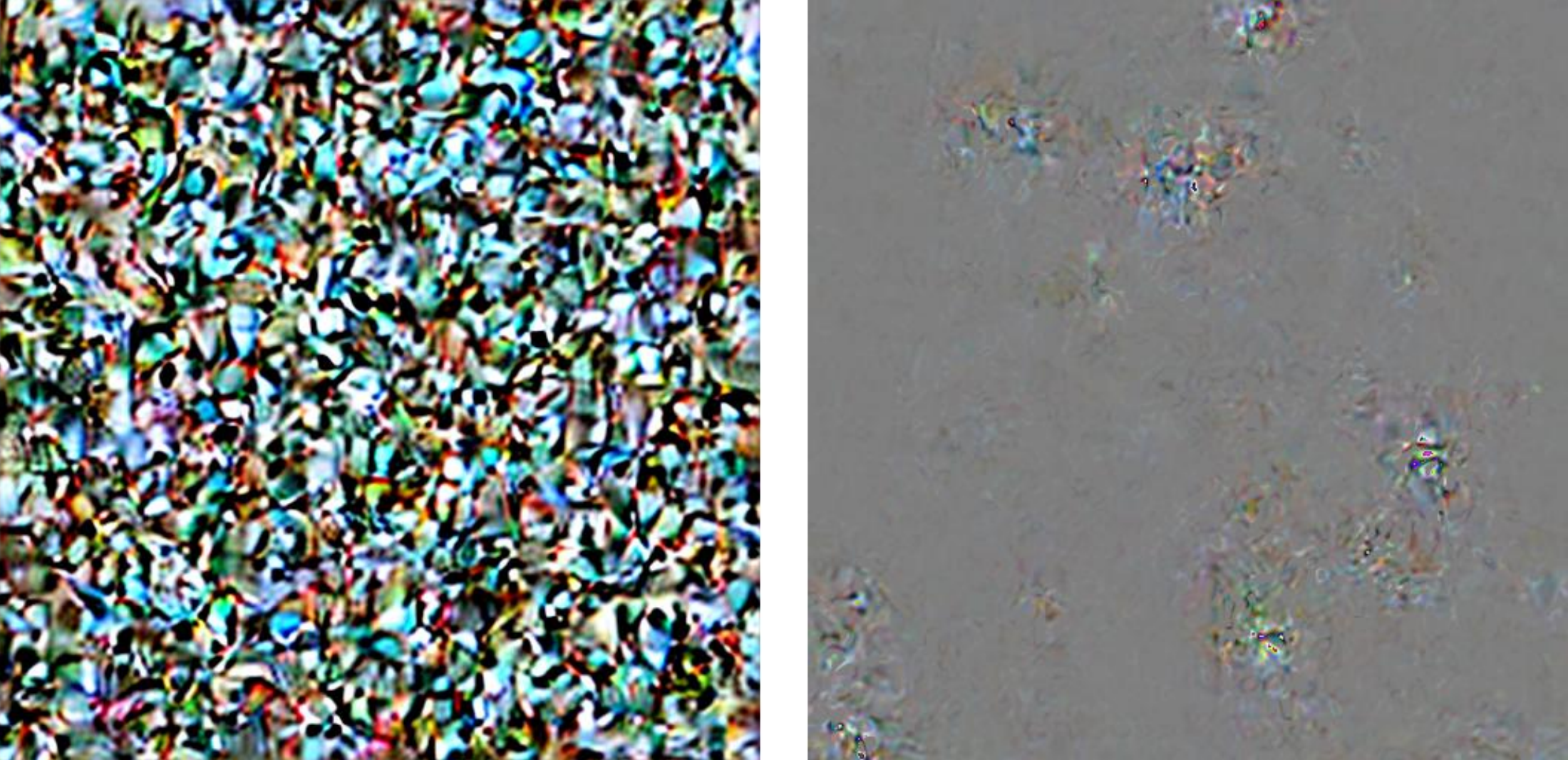}\\[-\dp\strutbox] \end{tabular}   \\
    \end{tabular}
    \caption{Our Algs. \ref{alg:inv_ddim} and \ref{alg:2} significantly reduce reconstruction errors, whether it's for images or noise, DDIM or high-order DPM-solvers, or pixel-space DPM or LDM. The generation / inversion method varies for each row, \textit{e.g.}, `na\"ive / 1000' indicates that we performed the na\"ive DDIM inversion (\cref{eqn:naive_ddim_inv}) for 1000 steps. `Alg. \ref{alg:inv_ddim} / 50' and `Alg. \ref{alg:2} / 10' attempt exact inversion with 50 steps of DDIM and 10 steps of DPM-Solver++(2M), respectively. Achieving exact inversion in LDM is challenging due to information loss from the autoencoder and instability caused by a classifier-free guidance of 3.0. Nonetheless, our algorithm produces good results also in LDM.}
    \label{fig:recon_qualitative}
\end{figure*}

\begin{figure*}[t]
  \centering 
  \begin{subfigure}[t]{0.20\linewidth} 
    \includegraphics[width=\textwidth]{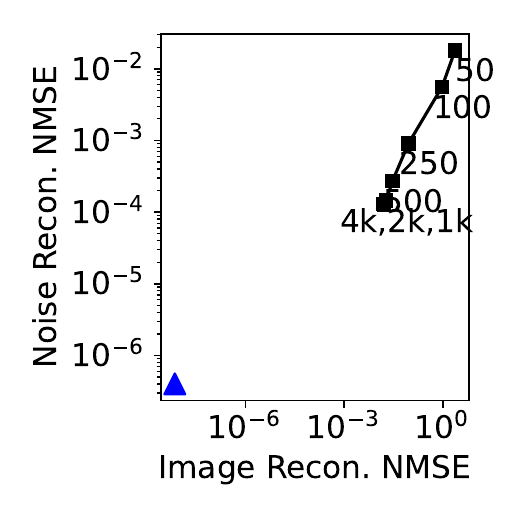}
    \captionsetup{justification=centering}
    \caption{DDIM 50 steps,\\ Pixel-space DPM}
    \label{fig:p_recon_imagenet_ddim}
  \end{subfigure}
  \begin{subfigure}[t]{0.20\linewidth} 
    \includegraphics[width=\textwidth]{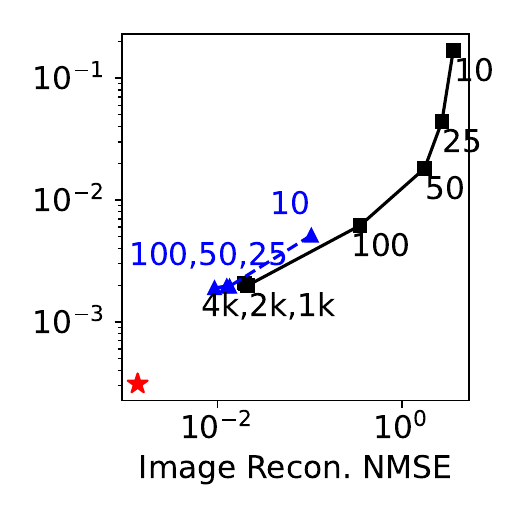}
    \captionsetup{justification=centering}
    \caption{DPM-Solver++(2M) 10 steps, Pixel-space DPM}
    \label{fig:p_recon_imagenet_dpm}
  \end{subfigure}
  \begin{subfigure}[t]{0.20\linewidth} 
    \includegraphics[width=\textwidth]{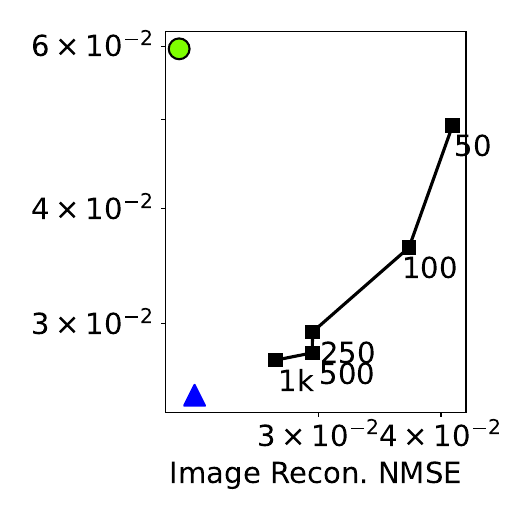}
    \captionsetup{justification=centering}
    \caption{DDIM 50 steps, \\LDM}
    \label{fig:l_recon_imagenet_ddim}
  \end{subfigure}
  \begin{subfigure}[t]{0.20\linewidth} 
    \includegraphics[width=\textwidth]{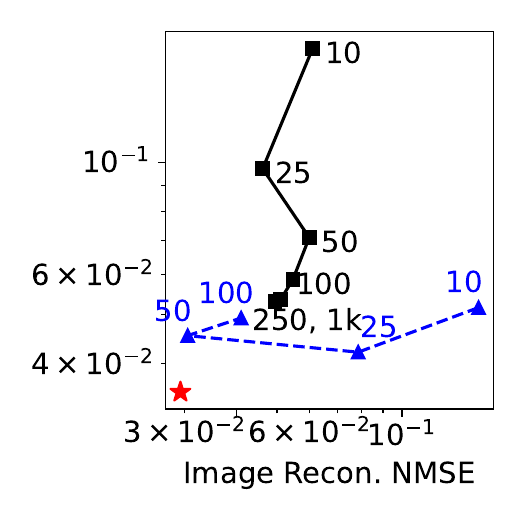}
    \captionsetup{justification=centering}
    \caption{DPM-Solver++(2M) \\ 10 steps, LDM }
    \label{fig:l_recon_imagenet_dpm}
  \end{subfigure}
  \begin{subfigure}[t]{0.18\linewidth} 
    \raisebox{3mm}{\includegraphics[width=\textwidth]{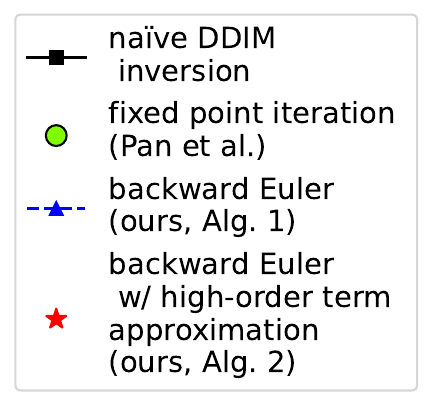}}
  \end{subfigure}
  \caption{Our algorithms reconstruct better than the na\"ive DDIM inversion. When the number of steps in the na\"ive DDIM inversion is increased, the reconstruction error can be reduced, but it becomes saturated (black). Since DPM-solvers are incorrect in the aspects of the diffusion ODE, correcting their errors can further reduce the reconstruction errors. \ref{fig:p_recon_imagenet_ddim} and \ref{fig:l_recon_imagenet_ddim} were generated with DDIM using 50 steps, so \cref{alg:inv_ddim} based on the backward Euler (blue) minimizes the reconstruction errors, while \ref{fig:p_recon_imagenet_dpm} and \ref{fig:l_recon_imagenet_dpm} were generated with DPM-Solver++(2M) using 10 steps, making \cref{alg:2}, which approximates high-order terms, the best performer (red). \citet{Pan_2023_ICCV}'s method using FPI exhibits poor performance on noise reconstruction in \ref{fig:l_recon_imagenet_ddim}, because of its weakness at large classifier-free guidance ($>1$). }
  \label{fig:recon_quantitative}
\end{figure*}

Experimental results show that our Algs. \ref{alg:inv_ddim} and \ref{alg:2} significantly reduce reconstruction errors than the na\"ive DDIM inversion, whether it's for images or noise, DDIM or high-order DPM-solver, or pixel-space DPM or LDM (see \cref{fig:recon_qualitative} and \cref{fig:recon_quantitative} for qualitative and quantitative results, respectively). In \cref{fig:l_recon_imagenet_ddim}, we also show that inversion with FPI (`AIDI\_E' of \citet{Pan_2023_ICCV}) exhibits poor performance in noise reconstruction, as we noted in \cref{sec:method-1}. 

Some may argue that fine-grained na\"ive DDIM inversion should perform well as it converges to the diffusion ODE trajectory (\textit{i.e.}, \cref{eq:exact_solution_x0}). However, that is not the case, as DPM-solvers make a \emph{discretized} trajectory. Even if we make the na\"ive DDIM inversion finer to closely follow the ODE solution, it cannot further reduce the reconstruction error, as seen in the black lines in \cref{fig:recon_quantitative}. 
Therefore, we must use implicit methods like our algorithms to address it.

\begin{figure*}[t]
    \small
    \centering
    \setlength{\tabcolsep}{0pt}
    \begin{tabular}{@{}c@{ }c@{ }c@{ }c@{ }c@{ }c@{ }c@{ }c@{}}
     \begin{tabular}{@{}c@{}}Embedded\\ watermark\end{tabular} & \begin{tabular}{@{}c@{}}Generated by\\ DPM-Solver++\end{tabular} & \multicolumn{2}{c@{ }}{\cellcolor{bGray} \begin{tabular}{@{}c@{ }} na\"ive DDIM inversion \\ (Recon. / Error)\end{tabular}} &  \multicolumn{2}{c@{ }}{\cellcolor{bGray} \begin{tabular}{@{}c@{}} na\"ive DDIM inversion w/ $\mathcal{D}^\dagger$ \\ (Recon. / Error) \end{tabular}} &  \multicolumn{2}{c@{}}{\cellcolor{bRed} \begin{tabular}{@{}c@{}}\Cref{alg:2} (ours) \\ (Recon. / Error) \end{tabular}}  \\
    
    \includegraphics[width=0.120\linewidth]{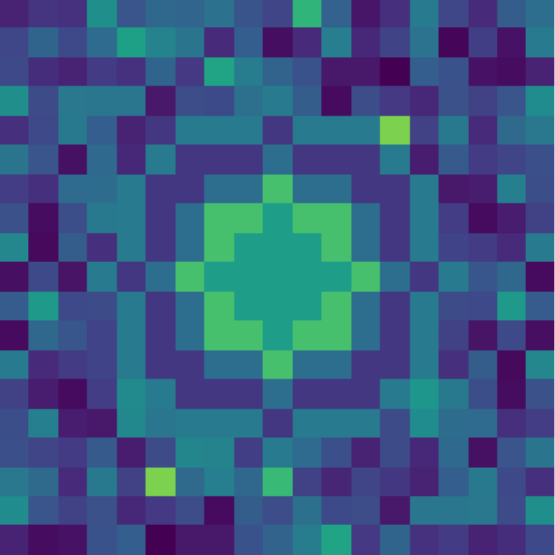}
     & \includegraphics[width=0.120\linewidth]{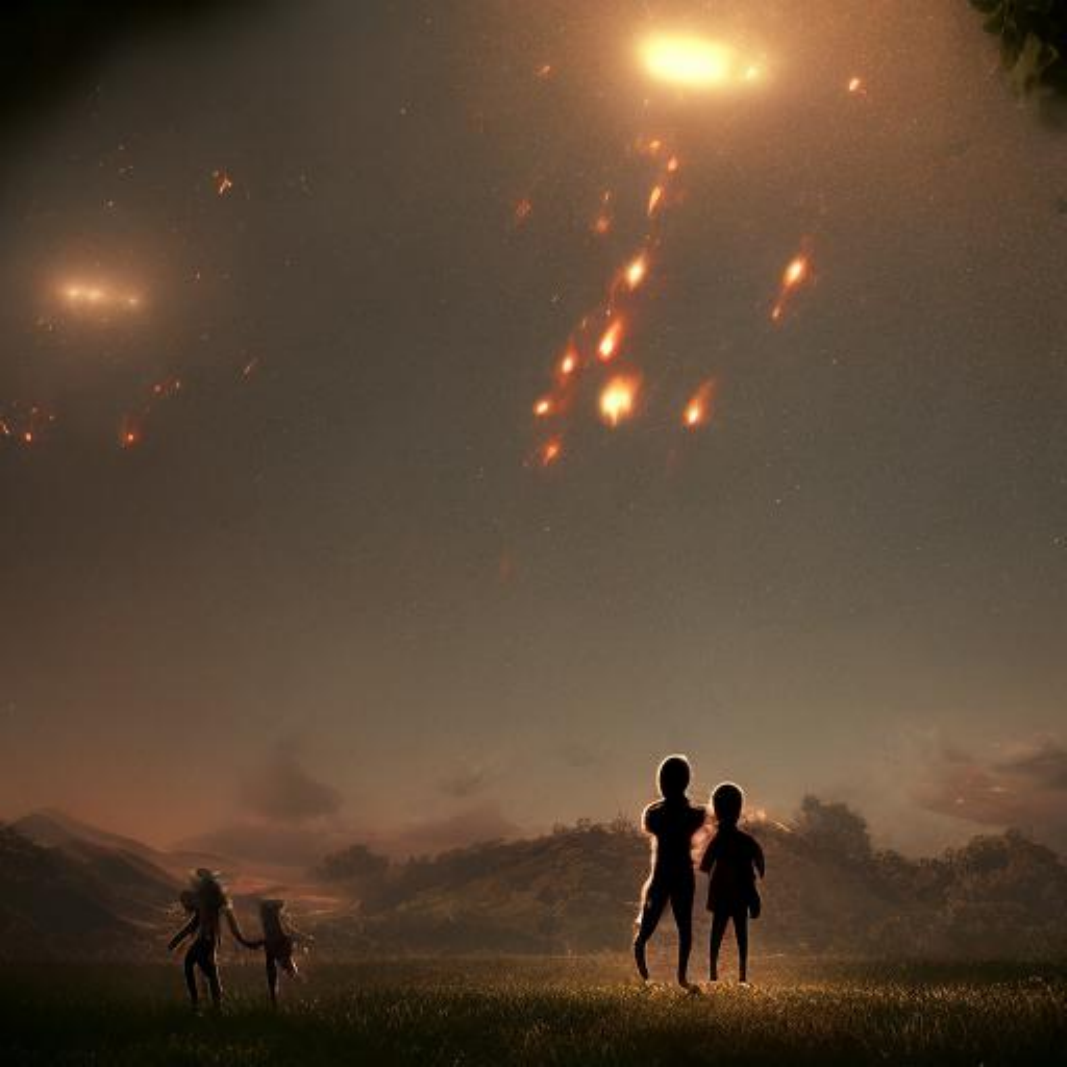} & \includegraphics[width=0.120\linewidth]{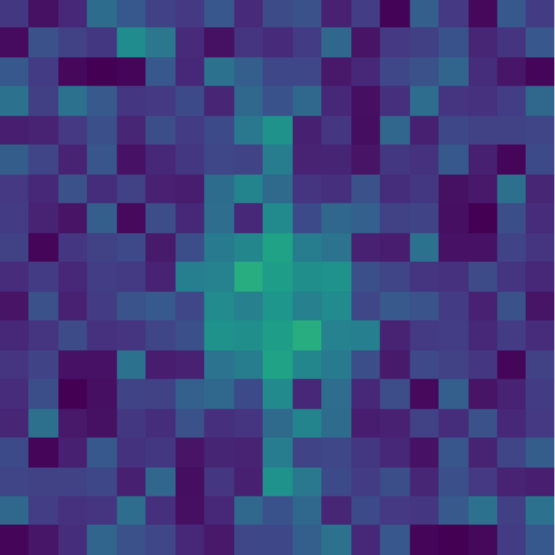} & \includegraphics[width=0.120\linewidth]{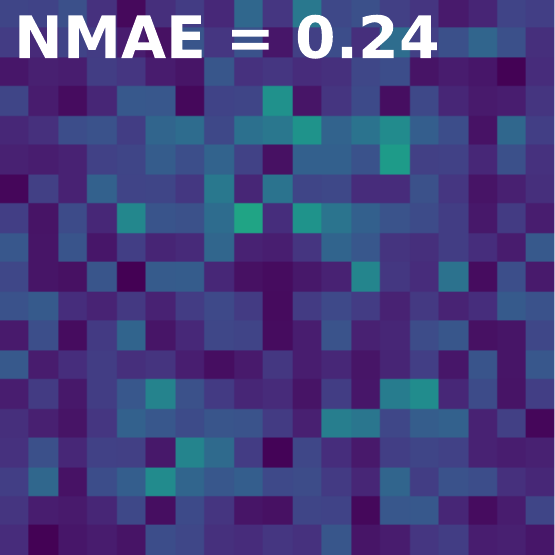} & \includegraphics[width=0.120\linewidth]{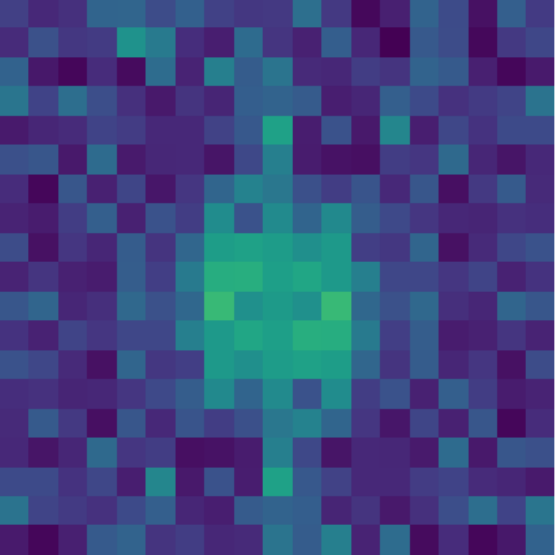} & \includegraphics[width=0.120\linewidth]{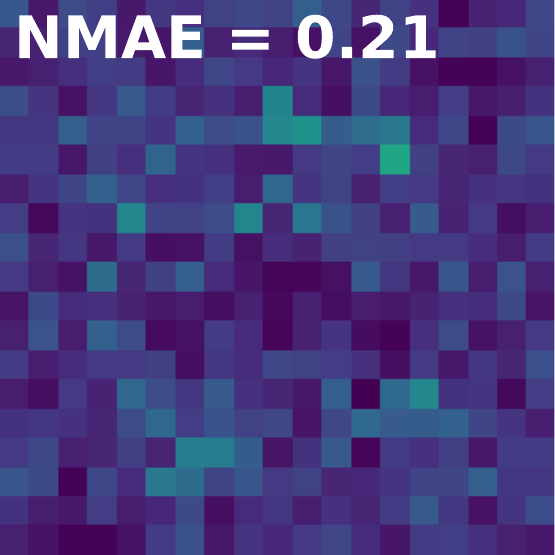} & \includegraphics[width=0.120\linewidth]{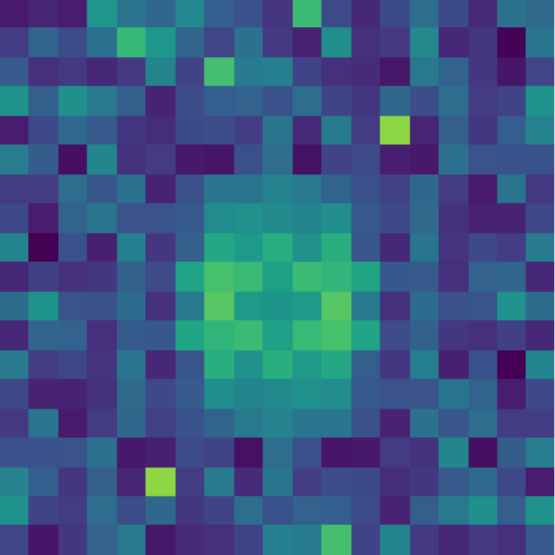} & \includegraphics[width=0.120\linewidth]{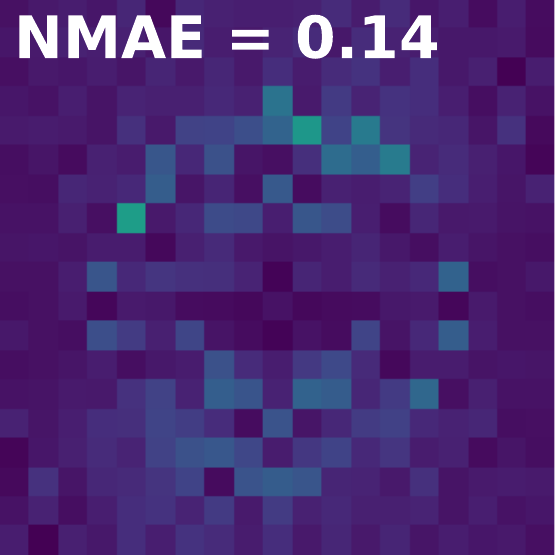} \\
     
     \includegraphics[width=0.120\linewidth]{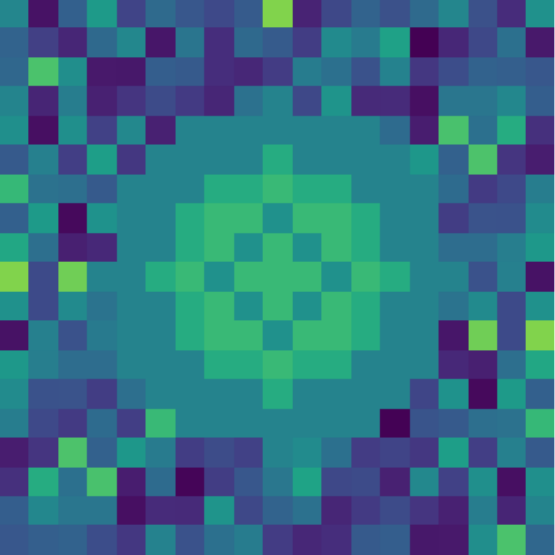}
     & \includegraphics[width=0.120\linewidth]{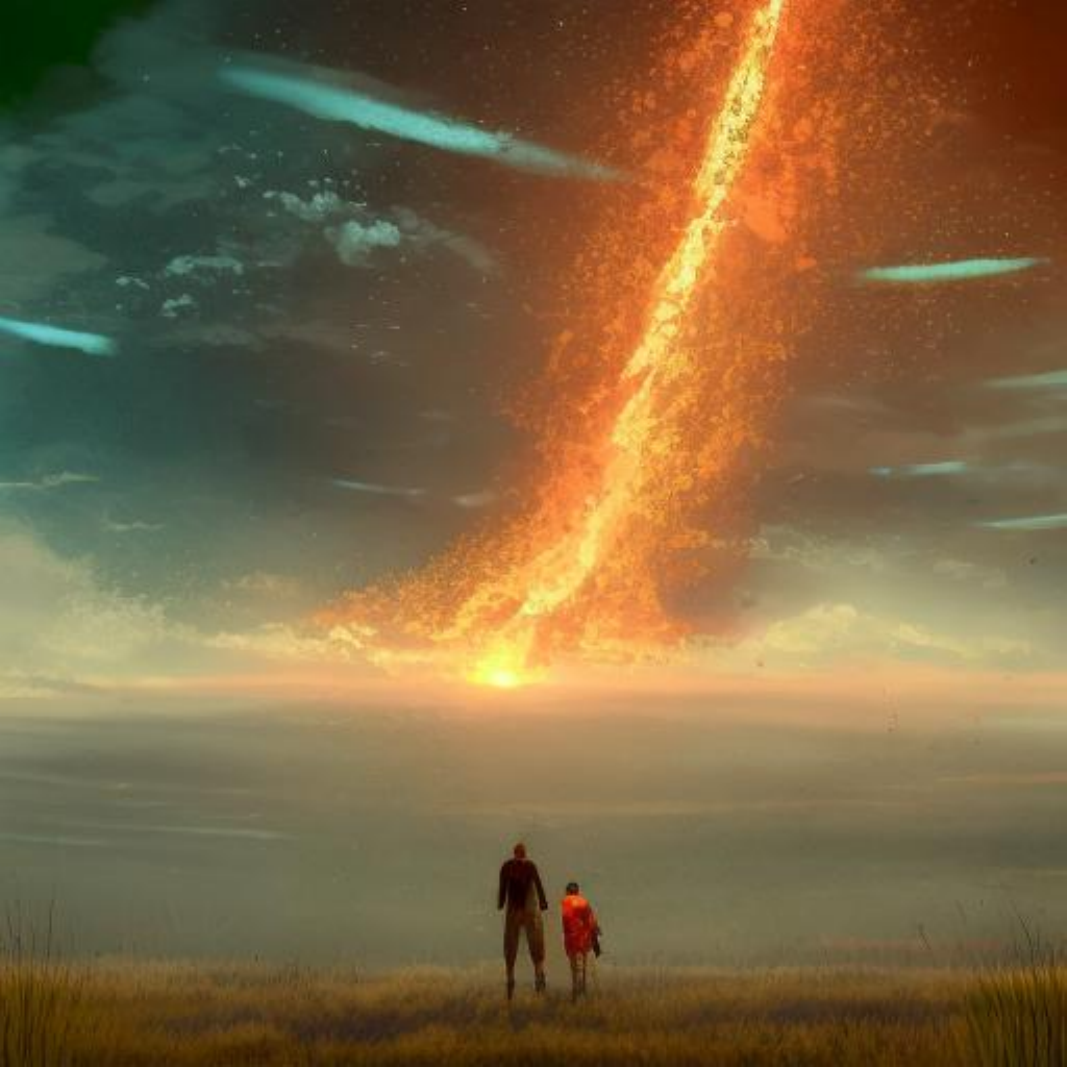} & \includegraphics[width=0.120\linewidth]{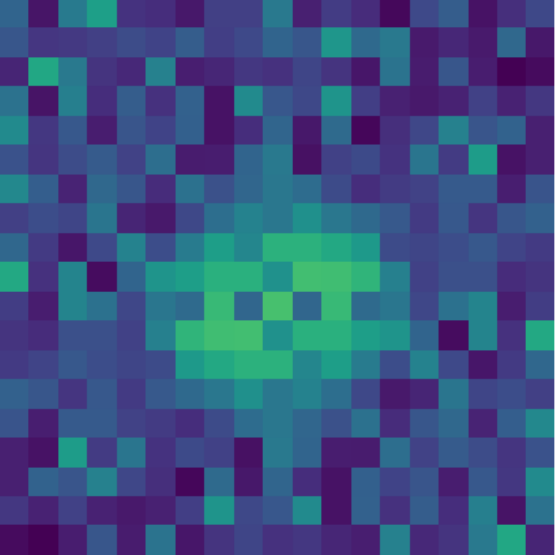} & \includegraphics[width=0.120\linewidth]{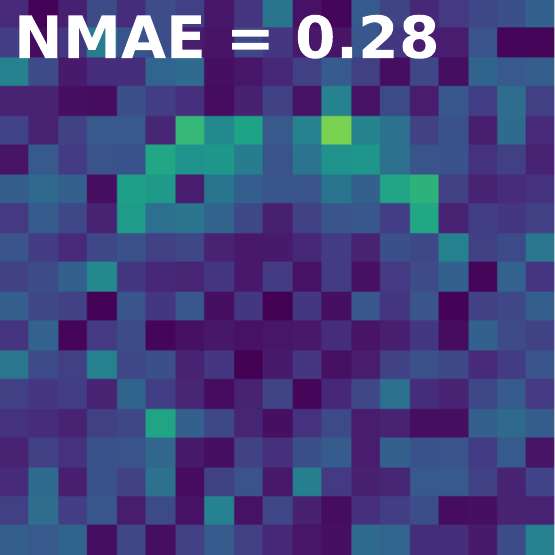} & \includegraphics[width=0.120\linewidth]{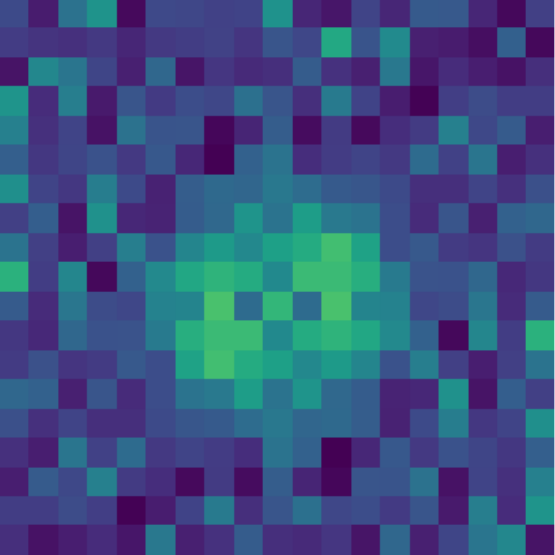} & \includegraphics[width=0.120\linewidth]{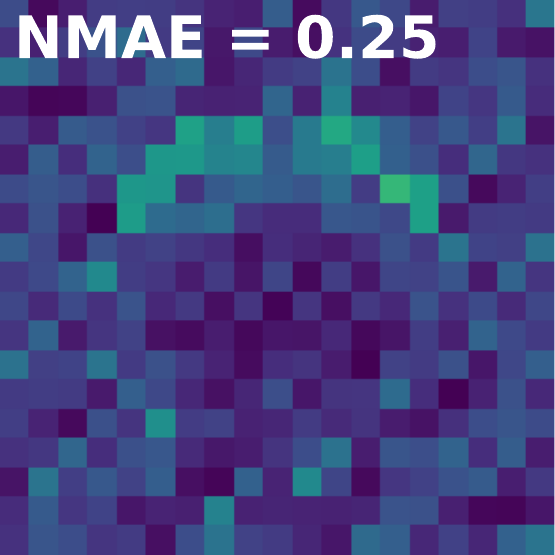} & \includegraphics[width=0.120\linewidth]{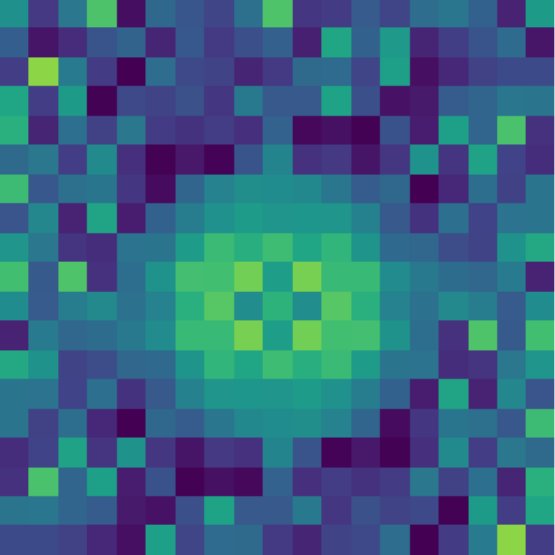} & \includegraphics[width=0.120\linewidth]{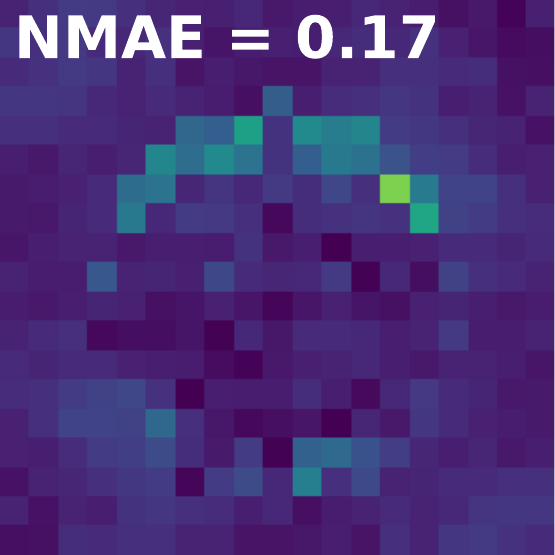} \\
     
     \includegraphics[width=0.120\linewidth]{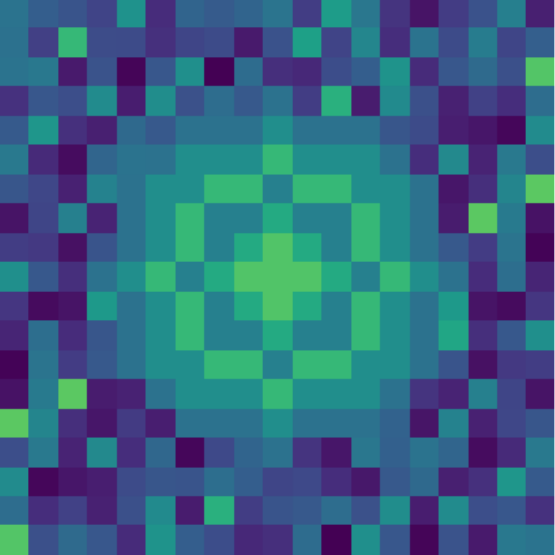}
     & \includegraphics[width=0.120\linewidth]{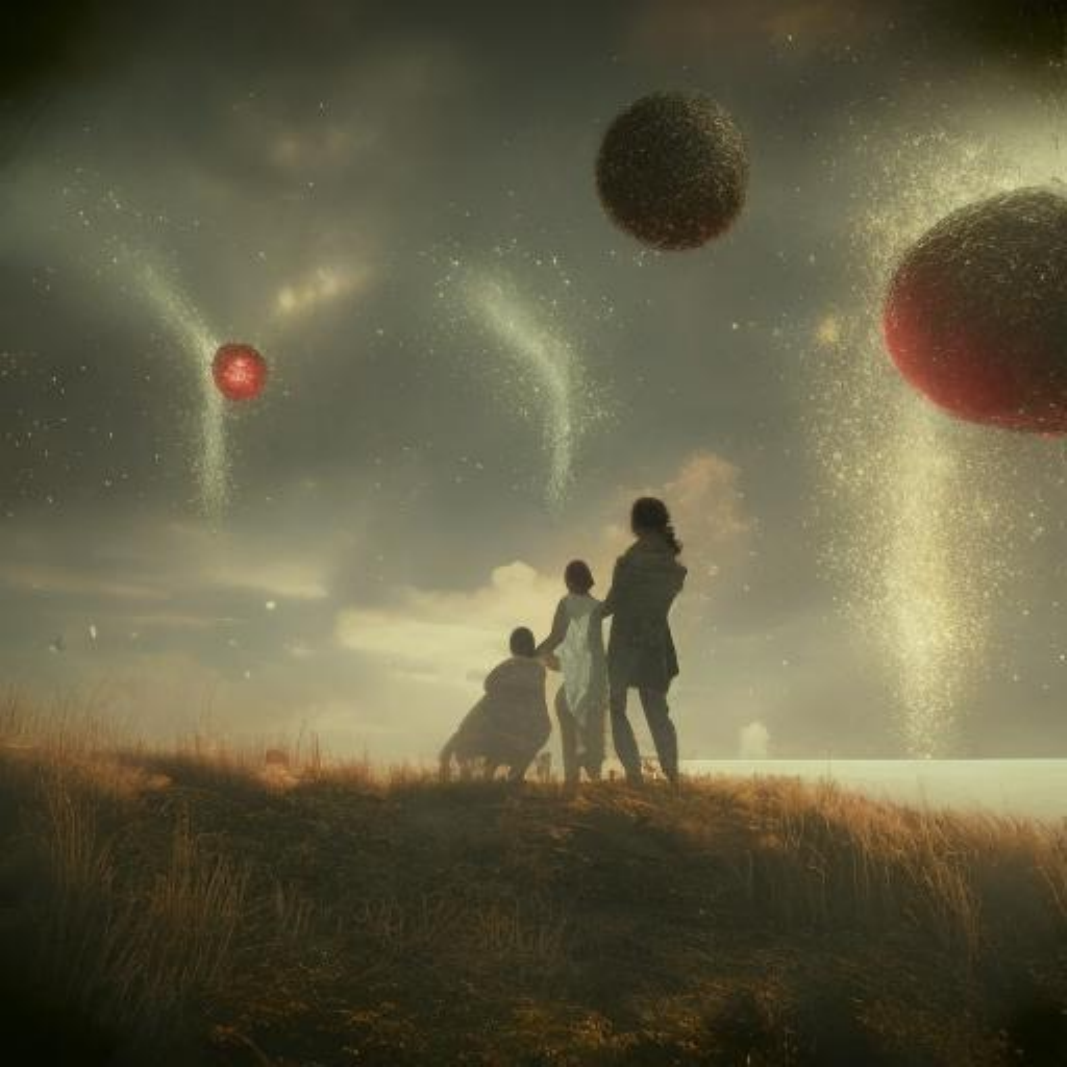} & \includegraphics[width=0.120\linewidth]{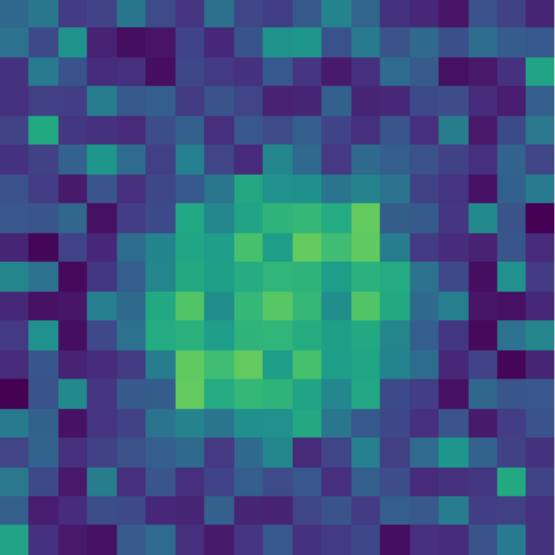} & \includegraphics[width=0.120\linewidth]{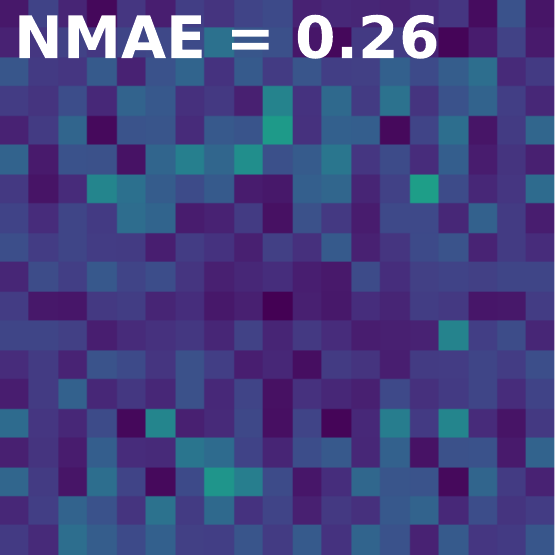} & \includegraphics[width=0.120\linewidth]{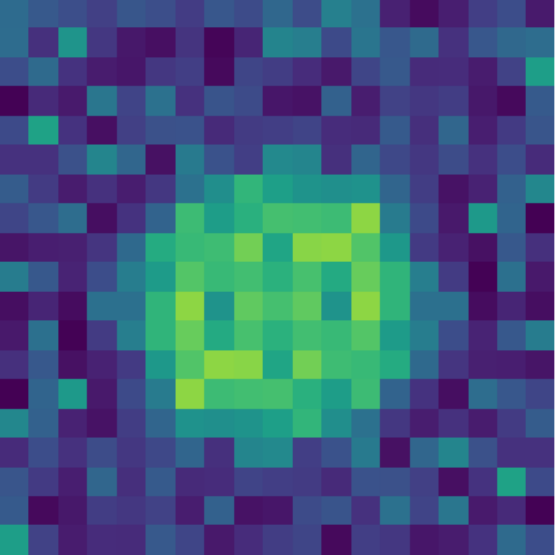} & \includegraphics[width=0.120\linewidth]{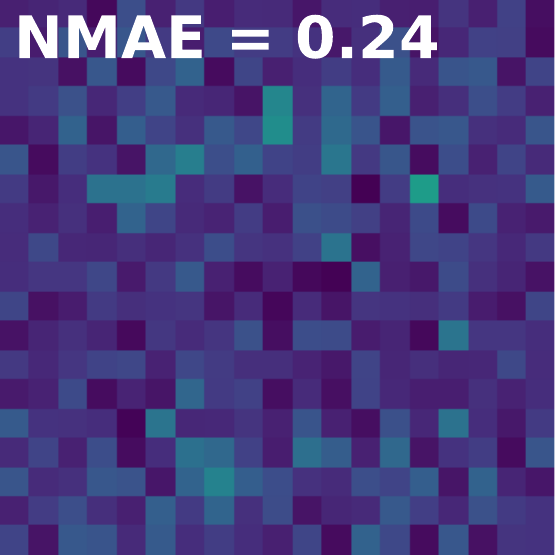} & \includegraphics[width=0.120\linewidth]{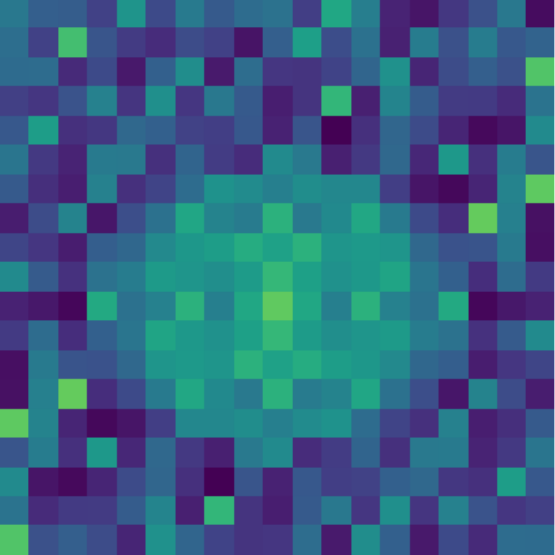} & \includegraphics[width=0.120\linewidth]{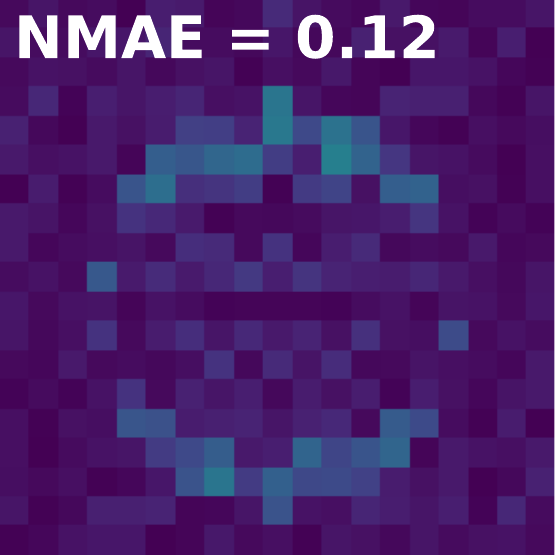} \\
    \end{tabular}
    \caption{Our \cref{alg:2} enables accurate reconstruction of Tree-ring watermarks~\cite{wen2023tree} in the Fourier space of the initial noise ($\vz_T)$. The Tree-ring watermark is embedded in the Fourier space of the initial noise in the shape of tree-rings and can be utilized for copyright tracing (column 1). Then, the image is generated starting from the watermarked noise. The practical approach is to accelerate image generation using methods like DPM-Solver++(2M)~\cite{lu2022dpm++} (column 2). Using \cref{alg:2} (columns 7-8) for watermark reconstruction results in lower errors compared to employing na\"ive DDIM inversion (columns 3-6). We provide NMAE on each error map.}
    \label{fig:watermarks}
\end{figure*}

\begin{figure*}[t]
    \small
    \centering   
    \begin{subfigure}[]{0.3\textwidth}
    \centering
        \begin{tikzpicture}
            \begin{axis}[
                    width=0.9\linewidth,  
                    height=0.625\linewidth,  
                    colormap={custom}{
                        color(0cm)=(white);
                        color(1cm)=(yellow);
                        color(2cm)=(green);
                        color(3cm)=(green);
                    },
                    xlabel=Predicted,
                    xlabel style={yshift=0pt},
                    ylabel=Actual,
                    ylabel style={yshift=3pt},
                    xticklabels={WM 1, WM 2, WM 3},
                    xtick={0,...,2},
                    xtick style={draw=none},
                    yticklabels={WM 1, WM 2, WM 3},
                    ytick={0,...,2},
                    ytick style={draw=none},
                    enlargelimits=false,
                    xticklabel style={},
                    nodes near coords={\pgfmathprintnumber\pgfplotspointmeta},
                    nodes near coords style={
                        yshift=-7pt
                    },
                ]
                \addplot[
                    matrix plot,
                    mesh/cols=3,
                    point meta=explicit,draw=gray
                ] table [meta=C] {
                    x y C
                    0 0 99
                    1 0 1
                    2 0 0

                    0 1 60
                    1 1 16
                    2 1 24

                    0 2 40
                    1 2 0
                    2 2 60

                };
            \end{axis}
        \end{tikzpicture}
    \caption{Na\"ive DDIM inversion}
    \label{fig:confusion_naive}
    \end{subfigure}
    \hfill
    \begin{subfigure}[]{0.3\textwidth}
        \centering
        \begin{tikzpicture}
            \begin{axis}[
                    width=0.9\linewidth,  
                    height=0.625\linewidth,  
                    colormap={custom}{
                        color(0cm)=(white);
                        color(1cm)=(yellow);
                        color(2cm)=(green);
                        color(3cm)=(green);
                    },
                    xlabel=Predicted,
                    xlabel style={yshift=0pt},
                    ylabel=Actual,
                    ylabel style={yshift=3pt},
                    xticklabels={WM 1, WM 2, WM 3},
                    xtick={0,...,2},
                    xtick style={draw=none},
                    yticklabels={WM 1, WM 2, WM 3},
                    ytick={0,...,2},
                    ytick style={draw=none},
                    enlargelimits=false,
                    xticklabel style={},
                    nodes near coords={\pgfmathprintnumber\pgfplotspointmeta},
                    nodes near coords style={
                        yshift=-7pt
                    },
                ]
                \addplot[
                    matrix plot,
                    mesh/cols=3,
                    point meta=explicit,draw=gray
                ] table [meta=C] {
                    x y C
                    0 0 99
                    1 0 1
                    2 0 0

                    0 1 44
                    1 1 36
                    2 1 20

                    0 2 26
                    1 2 0
                    2 2 74

                };
            \end{axis}
        \end{tikzpicture}
    \caption{na\"ive DDIM inversion w/ $\mathcal{D}^\dagger$ in Sec. \ref{sec:method-1}}
    \label{fig:confusion_naive+}
    \end{subfigure}
    \hfill
    \begin{subfigure}[]{0.375\textwidth}
    \centering
        \begin{tikzpicture}
            \begin{axis}[
                    width=0.72\linewidth,  
                    height=0.50\linewidth,  
                    colormap={custom}{
                        color(0cm)=(white);
                        color(1cm)=(yellow);
                        color(2cm)=(green);
                        color(3cm)=(green);
                    },
                    xlabel=Predicted,
                    xlabel style={yshift=0pt},
                    ylabel=Actual,
                    ylabel style={yshift=3pt},
                    xticklabels={WM 1, WM 2, WM 3},
                    xtick={0,...,2},
                    xtick style={draw=none},
                    yticklabels={WM 1, WM 2, WM 3},
                    ytick={0,...,2},
                    ytick style={draw=none},
                    enlargelimits=false,
                    xticklabel style={},
                    colorbar,
                    nodes near coords={\pgfmathprintnumber\pgfplotspointmeta},
                    nodes near coords style={
                        yshift=-7pt
                    },
                ]
                \addplot[
                    matrix plot,
                    mesh/cols=3,
                    point meta=explicit,draw=gray
                ] table [meta=C] {
                    x y C
                    0 0 98
                    1 0 2
                    2 0 0

                    0 1 32
                    1 1 58
                    2 1 10

                    0 2 23
                    1 2 0
                    2 2 77

                };
            \end{axis}
        \end{tikzpicture}
    \caption{\Cref{alg:2} (ours)}
    \label{fig:confusion_alg2}
    \end{subfigure}
    
    \caption{Our algorithm's strong reconstruction performance allows for the classification of tree-ring watermarks as well. For copyright tracing, it is possible to generate images by embedding different unique watermarks. Three distinct watermarks (WM 1,2, and 3) are displayed in the first column of \cref{fig:watermarks}. In the confusion matrices, `Predicted' corresponds to the watermark with the smallest $l_1$ difference among the three watermarks. In Figs. \ref{fig:confusion_naive} and \ref{fig:confusion_naive+}, the na\"ive DDIM inversion encounters difficulties in detecting WM 2. In contrast (\cref{fig:confusion_alg2}), our \cref{alg:2} performs well in detecting WM 2.}
    \label{fig:confusions}
\end{figure*}

\begin{figure*}[!t]
    \setlength{\tabcolsep}{1pt}
    \centering
    \begin{tabular}{@{}l@{  }c@{}c@{  }c@{}c@{}}
    Method & Edited & Error map ($\times 5$) & Edited & Error map ($\times 5$) \\
     \begin{tabular}{@{}l@{}} Oracle \\ \cite{patashnik2023localizing} \end{tabular} &
    \begin{tabular}{c}\includegraphics[ clip, width=0.225\linewidth]{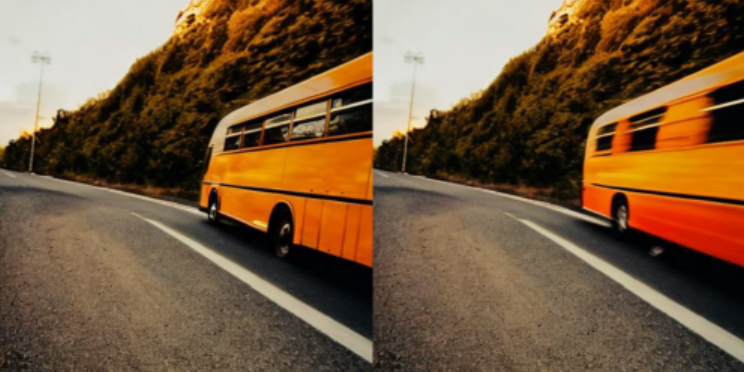}\end{tabular} & \begin{tabular}{c}\includegraphics[clip, width=0.225\linewidth]{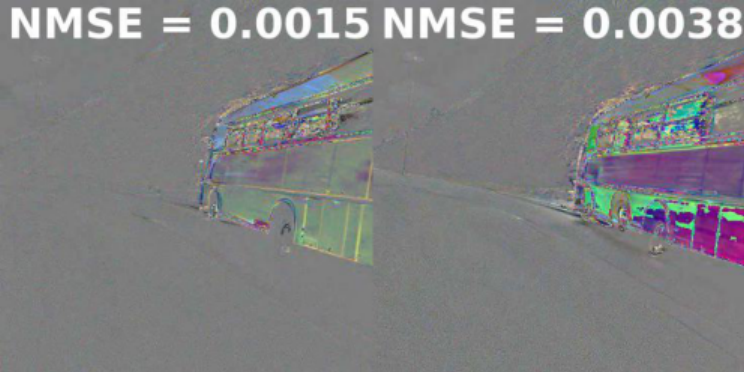}\end{tabular}& \begin{tabular}{c}\includegraphics[clip, width=0.225\linewidth]{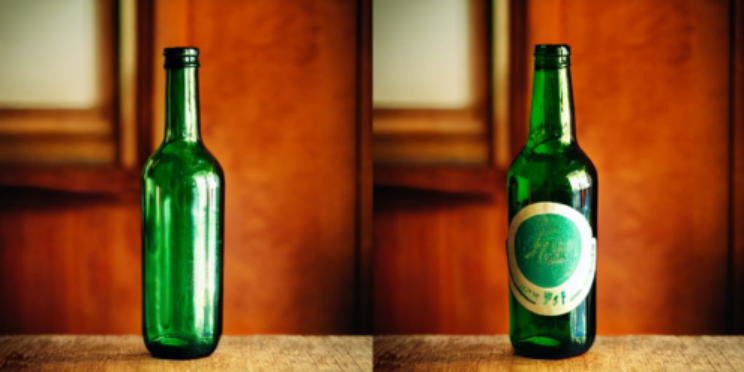}\end{tabular} & \begin{tabular}{c}\includegraphics[clip, width=0.225\linewidth]{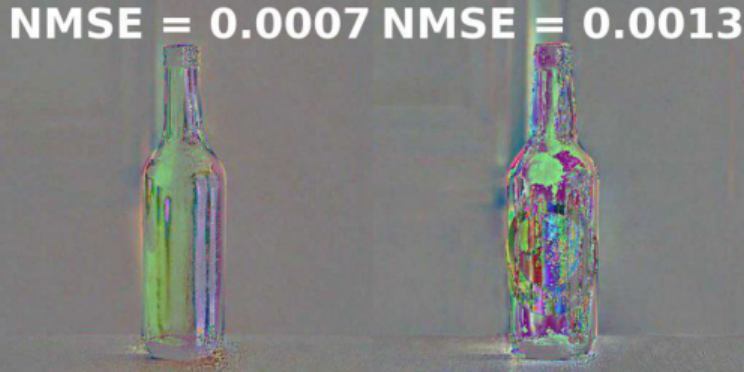}\end{tabular}\\
    \cellcolor{bGray} \begin{tabular}{@{}l@{}} na\"ive \\ DDIM \\ inversion \end{tabular} & \begin{tabular}{c}\includegraphics[clip, width=0.225\linewidth]{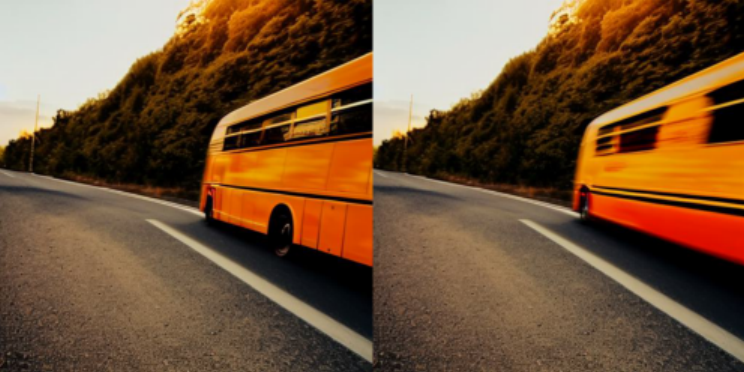}\end{tabular}&\begin{tabular}{c}\includegraphics[clip, width=0.225\linewidth]{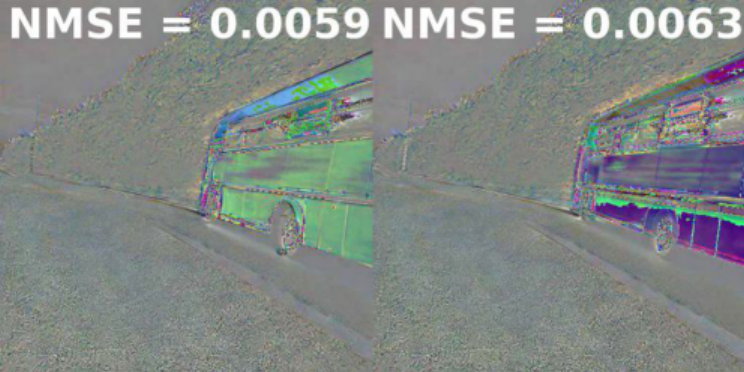}\end{tabular}& \begin{tabular}{c}\includegraphics[clip, width=0.225\linewidth]{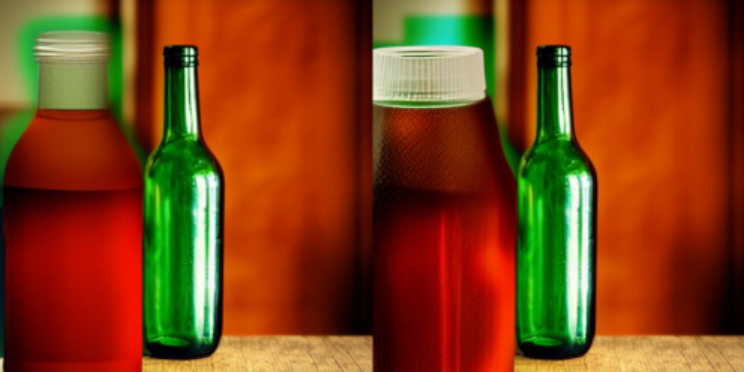}\end{tabular} & \begin{tabular}{c}\includegraphics[clip, width=0.225\linewidth]{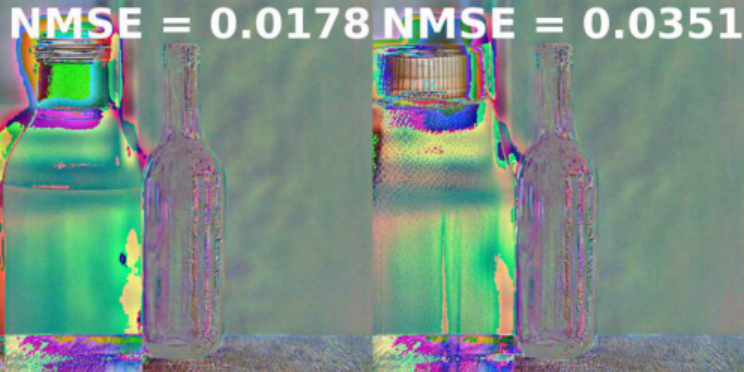}\end{tabular} \\ 
    \cellcolor{bGray} \begin{tabular}{@{}l@{}} na\"ive \\ DDIM \\ inversion \\ w/ $\gD^\dagger$ \end{tabular} & \begin{tabular}{c}\includegraphics[clip, width=0.225\linewidth]{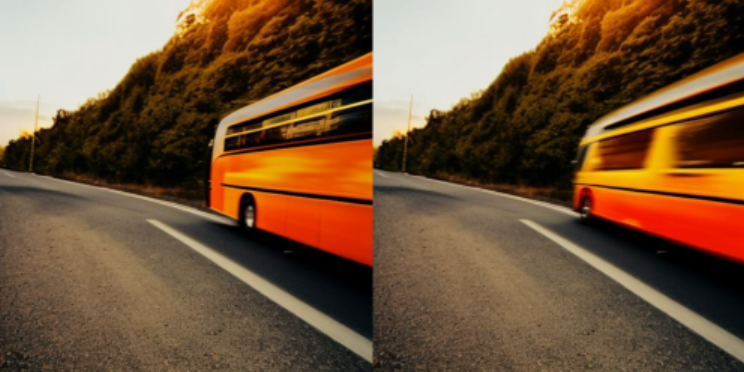}\end{tabular}&\begin{tabular}{c}\includegraphics[clip, width=0.225\linewidth]{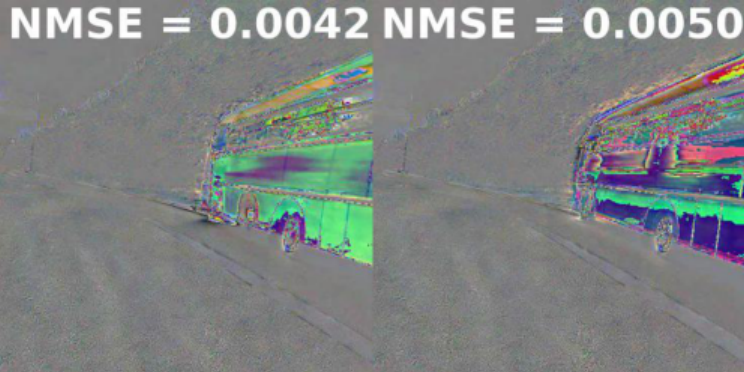}\end{tabular} & \begin{tabular}{c}\includegraphics[clip, width=0.225\linewidth]{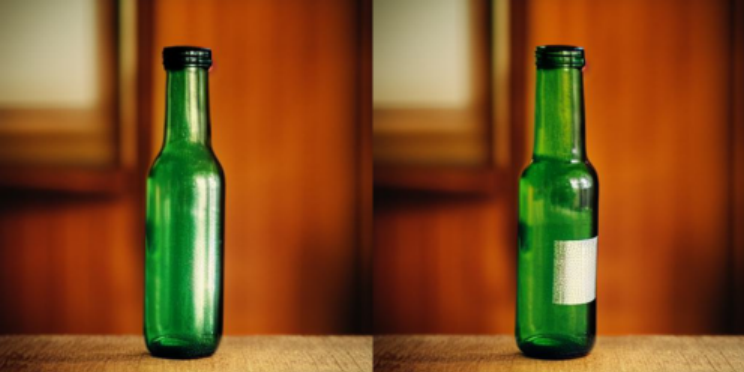}\end{tabular} & \begin{tabular}{c}\includegraphics[clip, width=0.225\linewidth]{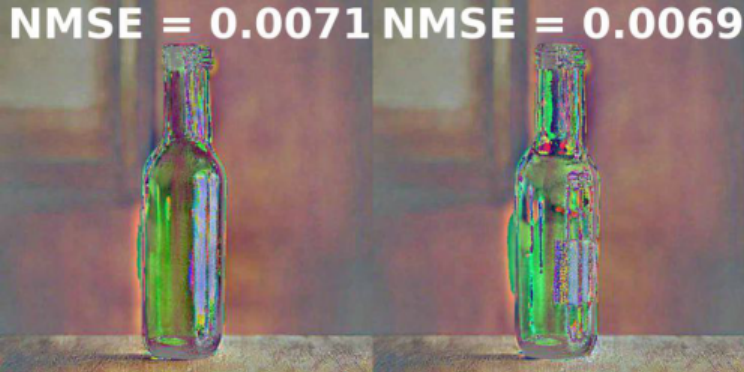}\end{tabular} \\
    \cellcolor{bBlue} \begin{tabular}{@{}c@{}} Alg. \ref{alg:inv_ddim} \\ (Ours) \end{tabular}   & \begin{tabular}{c}\includegraphics[ clip, width=0.225\linewidth]{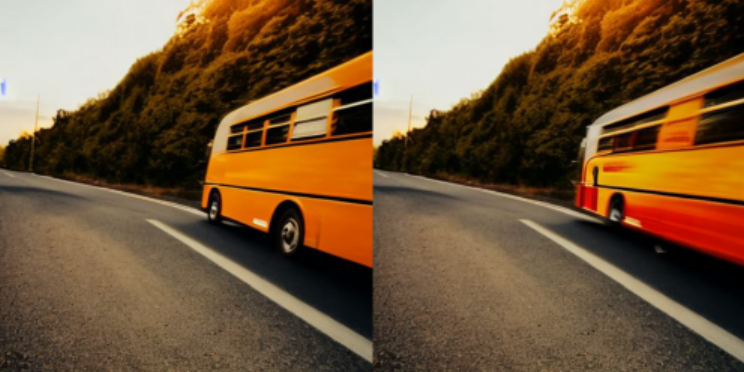}\\[-\dp\strutbox]\end{tabular}&\begin{tabular}{c} \includegraphics[clip, width=0.225\linewidth]{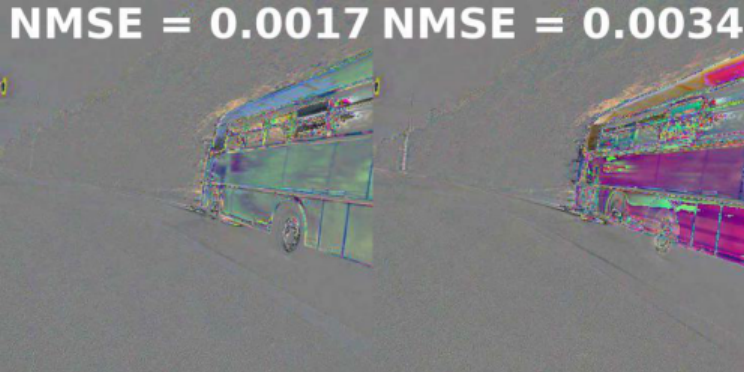}\\[-\dp\strutbox]\end{tabular}& \begin{tabular}{c}\includegraphics[clip, width=0.225\linewidth]{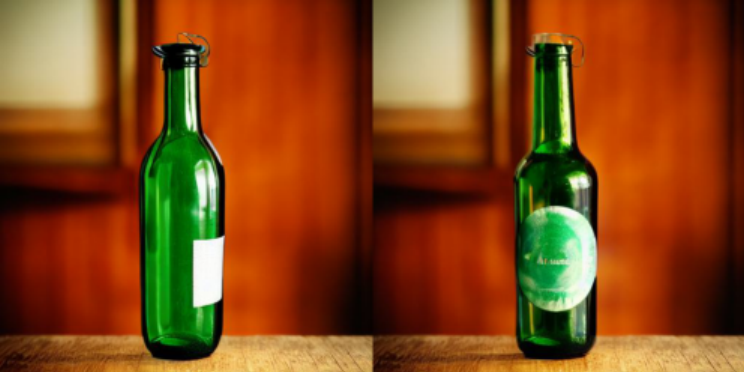}\\[-\dp\strutbox]\end{tabular} & \begin{tabular}{c}\includegraphics[clip, width=0.225\linewidth]{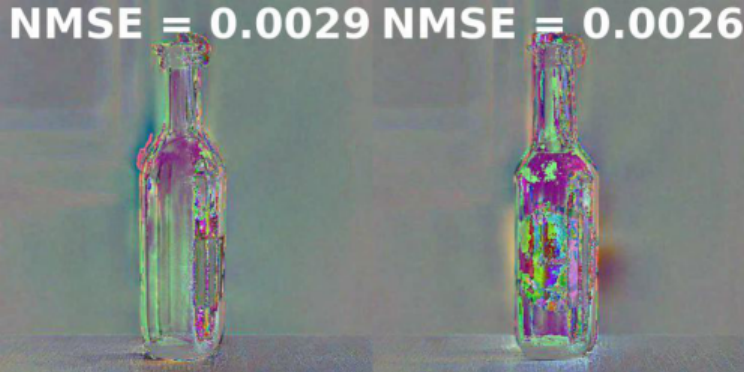}\\[-\dp\strutbox]\end{tabular}  \\
    \end{tabular}
    \caption{Our \cref{alg:inv_ddim} enables the preservation of the background and upholds high diversity of editing, even though the image’s original trajectory (\textit{i.e.}, $(\vz_{t_i})_{i=0}^M$) is unknown. The first row (Oracle) shows the result when the entire generating trajectory is provided, while in the subsequent rows, only the generated image (\textit{i.e.}, $\vx_0$) is given. In the latter cases,  we estimate the trajectory through each inversion method and perform editing based on the inversion results. While $\gD^\dagger$ (\textit{i.e.}, decoder inversion) enhances overall performance when employed with the na\"ive DDIM inversion, using the backward Euler as \cref{alg:inv_ddim} is necessary to achieve background-preserved edits at a level similar to that of the oracle. We provide NMSE of background on each error map.}
    \label{fig:editing}
\end{figure*}

\subsection{Application: Tree-ring watermark}\label{sec:exp:WM}
\citet{wen2023tree} proposed a new method for watermarking diffusion-generated images. It is invisible to human observers and robust to image manipulations. It works by embedding a watermark into the Fourier transform of the initial noise vector for image generattion. The watermark can be detected by inversion (to recover the initial noise vector) and comparing the Fourier transform to the expected watermark pattern. It can protect the intellectual property of the diffusion model and track diffusion-generated images' provenance. In \cref{sec:exp:WM}, we demonstrate that our proposed methods can enhance watermark detection.

In this subsection, we demonstrate the improved detection of watermarks~\cite{wen2023tree} by employing our algorithm, even when the images were generated using high-order DPM-solvers. Furthermore, with improved reconstruction, our algorithm can perform \emph{classification} as well. We used LDM~\cite{rombach2022high}, DPM-Solver++(2M) 10 steps, with classifier-free guidance $\omega=3.0$ to generate images. We embedded three different watermarks as in the first column of \cref{fig:watermarks}. \Cref{fig:watermarks} provides qualitative results of watermark detection, where the images were generated with the same prompt and different watermarks. Our \cref{alg:2} exhibits the best reconstruction performance.

\Cref{fig:confusions} shows quantitative results of watermark classification, where 100 images were generated for each watermark.  The $l_1$ norm is used for classification, as same in the detection~\cite{wen2023tree}. Our \cref{alg:2} exhibits the best performance in classification, as well as in the reconstruction.

\subsection{Application: Background-preserving editing}\label{sec:exp:edit}
One of the most common applications is image editing~\citep{hertz2022prompt,Kim_2022_CVPR,meng2022sdedit, nichol2021glide}: to manipulate an image based on a new condition while preserving information from the original image. \citet{patashnik2023localizing} proposed methods to localize the variations exclusively on the object while preserving the background. They suggested a prompt-mixing technique that switches the original and new prompt during the denoising process. Additionally, they introduced two localization techniques: self-attention map injection and blending the original latent image with the generated one. These techniques allowed them to utilize the information included in the original latents, the image structure and detailed appearance of the desired region (\textit{e.g.}, background, objects to preserve). In \cref{sec:exp:edit}, we experimentally demonstrate our proposed methods enable the background-preserving editing, without the need for the original latents.

Here, we experimentally show our \cref{alg:inv_ddim} enables the background-preserving editing proposed by \citet{patashnik2023localizing}, even though we don't know the whole denoising process of the original image (\textit{i.e.}, trajectory, $(\vz_{t_i})_{i=0}^M$). Note that \citet{patashnik2023localizing} employed oracle for the originally generated image, but any DDIM inversion methods (\textit{i.e.}, they knew the trajectory). \Cref{fig:editing} displays the results of performing background-preserving image editing~\cite{patashnik2023localizing}, where the original trajectory ($(\vz_{t_i})_{i=0}^M$) is estimated using the na\"ive DDIM inversion and our \cref{alg:inv_ddim}. Note that the classifier-free guidance $\omega$ is set to $7.5$, demonstrating the robustness of our method.

\section{Limitations}
The proposed method comes with a significantly larger computational time compared to na\"ive DDIM inversion. Additionally, it assumes prior knowledge of the prompt in the case of LDMs. Estimating the prompt and initial noise jointly is left as future work. 

\section{Conclusion}
We have presented exact inversion methods of DPM-solvers, to seek the initial noise of generated images. Our methods work by backward Euler implemented with gradient descent or the forward step method, which is robust to large classifier-free guidance. For the inversion of high-order DPM-solvers, we approximate high-order terms using the na\"ive DDIM inversion. Our method performs exact inversion well in various scenarios. It can be used to enhance the performance of noise-space watermark detection and even enable watermark classification. Additionally, our method can perform background-preserving editing effectively without requiring knowledge of the original image's generation trajectory. Our proposed method is widely applicable to standard DPMs, thus can encourage to create new DPM applications where exact inversion is essential.

\section*{Acknowledgements}
This work was supported in part by the National Research Foundation of Korea (NRF) grants funded by the Korea government (MSIT) (NRF-2022R1A4A1030579, NRF-2022M3C1A309202211) and Creative-Pioneering Researchers Program through Seoul National University. The authors acknowledged the financial support from the BK21 FOUR program of the Education and Research Program for Future ICT Pioneers, Seoul National University.

{
    \small
    \bibliographystyle{ieeenat_fullname}
    \bibliography{refs}
}
\clearpage
\newpage

\crefname{section}{Sec.}{Secs.}
\Crefname{section}{Section}{Sections}
\Crefname{table}{Table}{Tables}
\crefname{table}{Tab.}{Tabs.}
\renewcommand{\thesection}{S\arabic{section}}
\renewcommand{\thefigure}{S\arabic{figure}}
\renewcommand{\thetable}{S\arabic{table}}
\renewcommand{\theequation}{S\arabic{equation}}
\renewcommand{\thetheorem}{S\arabic{theorem}}
\setcounter{section}{0}
\setcounter{figure}{0}
\setcounter{table}{0}
\setcounter{equation}{0}
\setcounter{theorem}{0}

\section{Notes on instability of fixed point iteration in large classifier-free guidance}
In Section 4.1 of the main paper, we explained why the fixed point iteration (FPI)-based method~\cite{Pan_2023_ICCV} loses stability in the context of large classifier-free guidance. Here, we formulate and prove this as \cref{proposition1}. \Cref{proposition1} suggests that FPI may not converge when the classifier-free guidance is large.
\nobrackets
\begin{proposition}[Instability of FPI in large classifier-free guidance]\label{proposition1}
Let $\omega \geq 1$, $t_{i-1}$, $C$, and $\varnothing$ be appropriate inputs for 
\begin{equation}\label{eq:p1}
    \vx_\vtheta(\cdot, t_{i-1}) = \omega \bar\vx_\vtheta(\cdot, t_{i-1},C) - (1-\omega) \bar\vx_\vtheta(\cdot, t_{i-1},\varnothing).
\end{equation}
    If $\bar\vx_\vtheta(\cdot, t_{i-1}, C)$ is Lipschitz continuous with the constant
\begin{equation}\label{eq:p2}
    \frac{1}{|\omega|+|1-\omega|} \cdot \frac{\sigma_{t_i}}{\sigma_{t_{i-1}}\alpha_{t_i}(e^{-h_i} - 1)},
\end{equation}
then
\begin{equation}\label{eq:p3}
    F(\cdot) := \frac{\sigma_{t_{i-1}}}{\sigma_{t_i}} \alpha_{t_i}(e^{-h_i} - 1)\vx_\vtheta(\cdot, t_{i-1}) + \frac{\sigma_{t_{i-1}}}{\sigma_{t_i}}\hat\vx_{t_i}
\end{equation} 
is nonexpansive (\textit{i.e.}, $1$-Lipschitz continuous).
\end{proposition}

To prove \cref{proposition1}, we use the following \cref{lemma1} from page 5 in \cite{ryu2022large}: 
\begin{lemma}\label{lemma1}
    Let $a ,b \in \sR$. If $f,g:\gX \rightarrow \gY$ be $L_f,L_g$-Lipschitz continuous, respectively, then $af+bg$ is $|a|L_f + |b|L_g$-Lipschitz continuous.
\end{lemma}

Now we start the proof of \cref{proposition1}.
\begin{proof}[Proof of \cref{proposition1}]
By the assumption, \cref{eq:p1} and $\cref{lemma1}$, $\vx_\vtheta(\cdot, t_{i-1})$ is $\frac{\sigma_{t_i}}{\sigma_{t_{i-1}}\alpha_{t_i}(e^{-h_i} - 1)}$-Lipschitz continuous. By \cref{eq:p3} and $\cref{lemma1}$, $F(\cdot)$ is nonexpansive.
\end{proof}

\section{Experimental details}\label{sec:dtl}
\subsection{Reconstruction}\label{sec:dtl:recon}
\subsubsection{Pixel-space DPM}
For pixel-space DPM, we used gradient descent without momentum, with $l_2$ loss and a learning rate of 0.1. To dynamically adjust the learning rate, when the minimum loss for the last 5 iterations did not improve, the learning rate was halved, but not below a minimum of 0.001.
The number of iterations was a maximum of 500. The original model operates in half float (\textit{i.e.}, 16-bit), so it was used as is during gradient descent. We used $J=100$ in Algorithm 2. We experimented with 100 images.

\subsubsection{Latent diffusion model (LDM)}
For LDM, we generated images 
with classifier-free guidance of 3.0, and the same prompt was employed for the inversion process. We used the forward step method, with $l_2$ loss. The initial step size was set as 0.5 for Algorithm 1, and $10/t$ for Algorithm 2 except for the first step ($t=0$) where it was 1. To dynamically adjust the step size, when the minimum loss for the last 20 iterations did not improve, the step size was halved. For numerical stability, we applied a 20-step warmup (\textit{i.e.}, linearly increasing the step size from 0). We used $J=10$ in Algorithm 2. The number of iterations was a maximum of 500. Every operation was held in float (\textit{i.e.}, 32-bit). We experimented with 100 images.

\paragraph{Decoder inversion}
We applied decoder inversion for every method for a fair comparison, including na\"ive DDIM inversion and \cite{Pan_2023_ICCV}. We used Adam with $l_2$ loss and a learning rate of 0.1. For improved convergence, we applied a cosine learning rate scheduler with 10-step warmup within a total of 100 iterations.

\subsection{Application: Tree-ring watermark}\label{sec:dtl:WM}
The work of tree-ring watermark in \cite{wen2023tree} injects a watermark in the frequency domain of noise and then generates an image with noise obtained by inverse transformation. We tested on detection of watermarks by generating various types of watermarks with the same radius and calculated $l_1$ difference between the original watermark and the watermark obtained by inversion. The radius was set the same to make the comparison consistent and detection difficult. We tested on three watermarks with the shape of a tree-ring and a radius of 6 pixels. For watermark creation, we set a constant and designated pixel values to be random but close to constant to make the difference scale reliable.

\subsection{Application: Background-preserving editing}\label{sec:dtl:edit}
In this experiment, we employed the open-source code of ~\cite{patashnik2023localizing} to compare Algorithm 1 with three other methods: (i) The original code utilizing the latents stored during the generating process (\textit{i.e.}, oracle), (ii) the na\"ive DDIM inversion, and (iii) the na\"ive DDIM inversion with decoder inversion. We used classifier-free guidance of $7.5$ and for decoder inversion, we applied a cosine learning rate scheduler with 50-step warmup within a total of 500 iterations. All other experimental settings remained identical to those of LDM.

\section{Additional results}
\subsection{Reconstruction}\label{sec:add:recon}
\begin{figure*}[t]
    \setlength{\tabcolsep}{2pt}
    \centering
    \begin{tabular}{@{}c@{ }c@{ }c@{}c@{}c@{}c@{}}
    \multicolumn{1}{@{}c@{}}{\multirow{2}{*}{\rotatebox[origin=c]{90}{\centering \small Generation}}}& \multirow{2}{*}{\rotatebox[origin=c]{90}{\small Inversion}} & \multicolumn{2}{@{}c@{ }}{Pixel-space DPM} & \multicolumn{2}{c}{LDM}\\
    \cmidrule(l){3-4} \cmidrule(l){5-6}
        &            & \begin{tabular}{@{}c@{}} Image \\ \small (Recon. / Error $\times 2$) \end{tabular}  & \begin{tabular}{@{}c@{}} Noise \\ \small (Recon. / Error $\times 2$) \end{tabular} & \begin{tabular}{@{}c@{}} Image \\ \small (Recon. / Error $\times 3$) \end{tabular}  & \begin{tabular}{@{}c@{}} Noise \\ \small (Recon. / Error $\times 2$) \end{tabular}\\
     \cmidrule(l){3-4} \cmidrule(l){5-6}
     \multirow{2}{*}{\rotatebox[origin=c]{90}{ \small DDIM 50 steps}} &   \cellcolor{bGray}\rotatebox[origin=c]{90}{\small na\"ive / 1000}                        & \begin{tabular}{c}\includegraphics[trim=68 0 0 0, clip, width=0.230\linewidth]{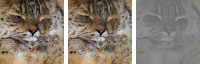}\end{tabular}      & \begin{tabular}{c}\includegraphics[trim=68 0 0 0, clip, width=0.230\linewidth]{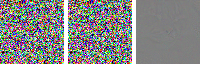}\end{tabular} & \begin{tabular}{c}\includegraphics[trim=544 0 0 0, clip, width=0.230\linewidth]{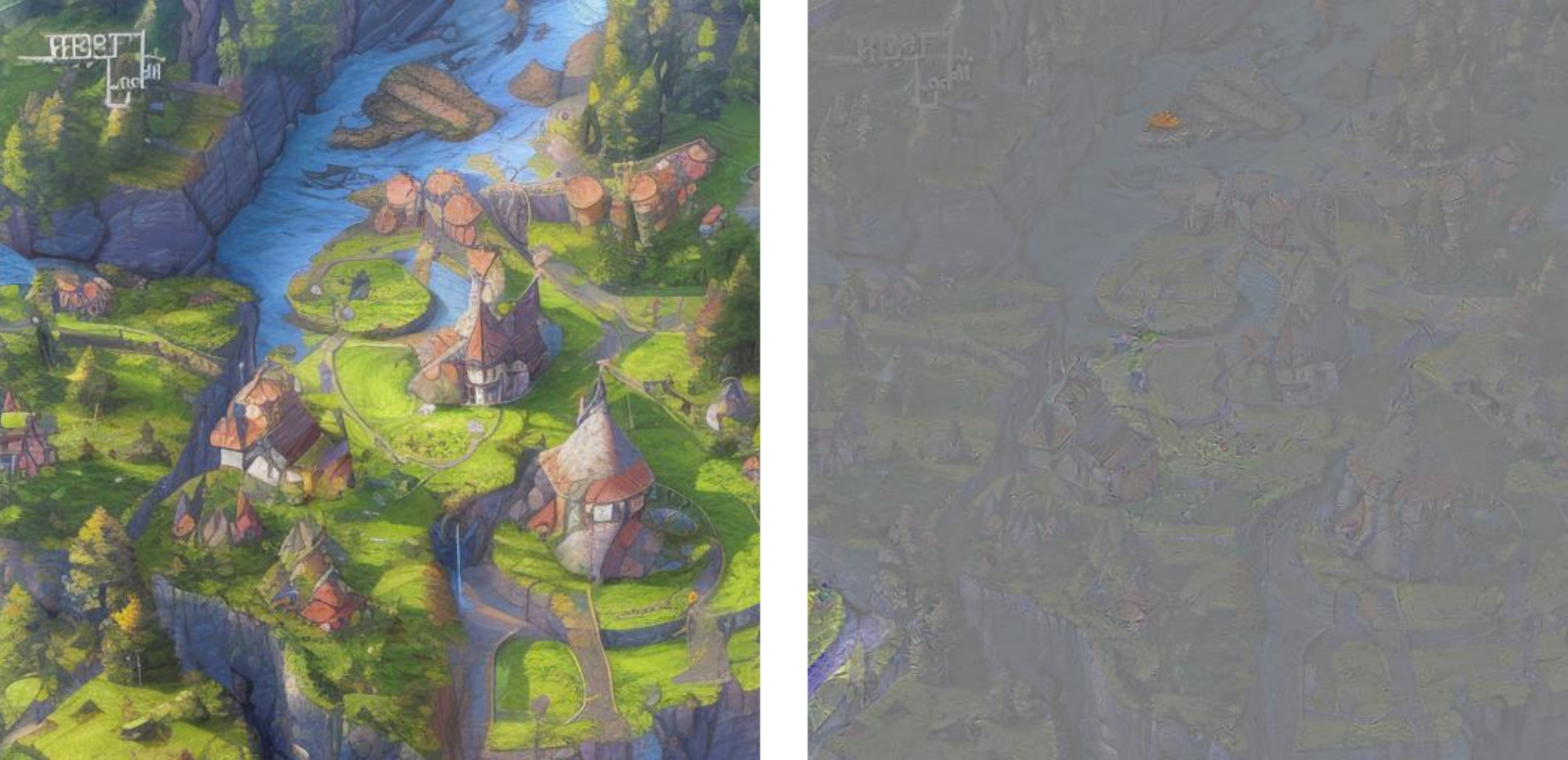}\end{tabular}      & \begin{tabular}{c}\includegraphics[trim=544 0 0 0, clip, width=0.230\linewidth]{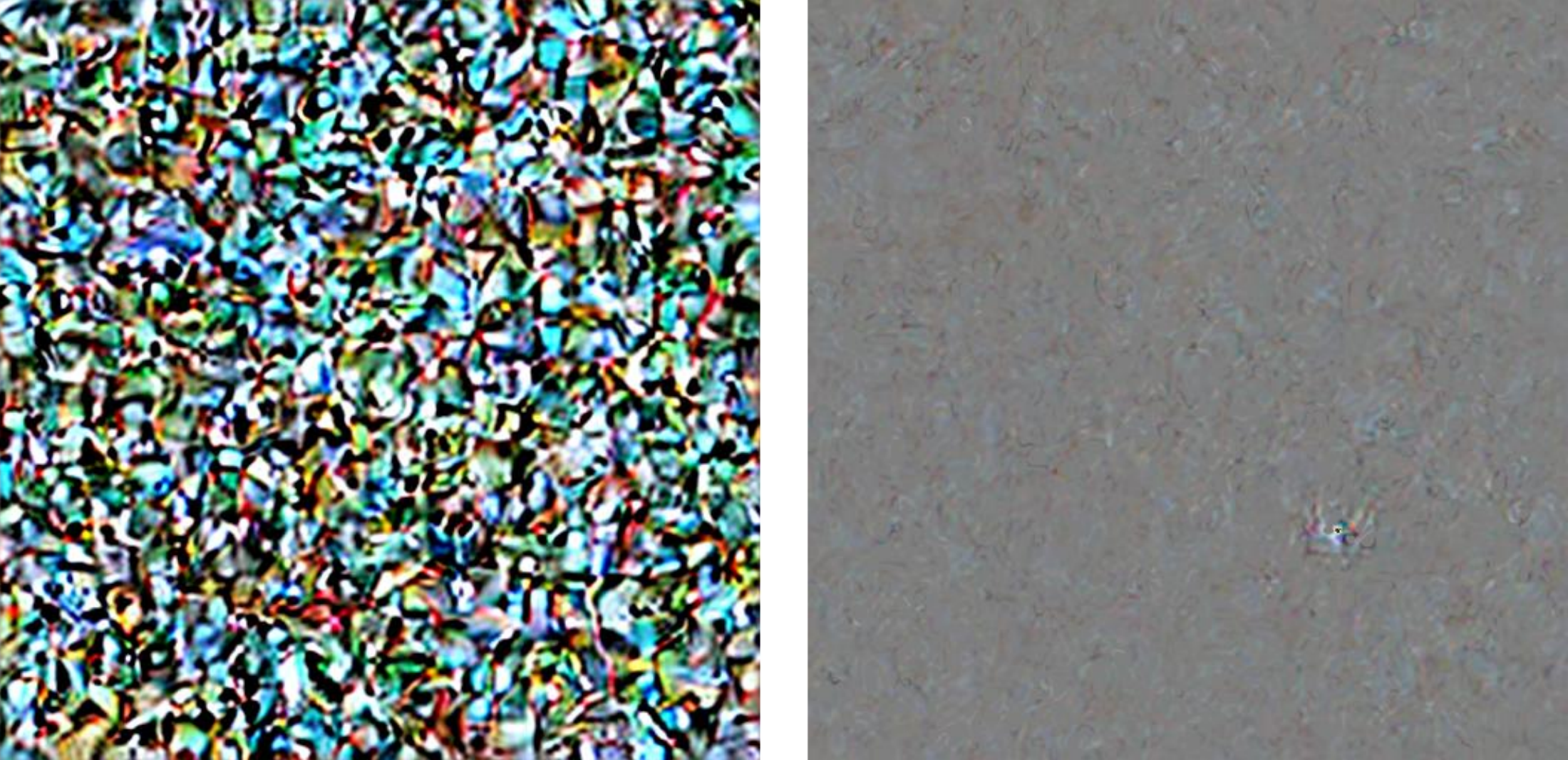}\end{tabular}     \\
     & \cellcolor{bGreen}\rotatebox[origin=c]{90}{\small FPI \cite{Pan_2023_ICCV} / 50}                          &     &  & \begin{tabular}{c}\includegraphics[trim=544 0 0 0, clip, width=0.230\linewidth]{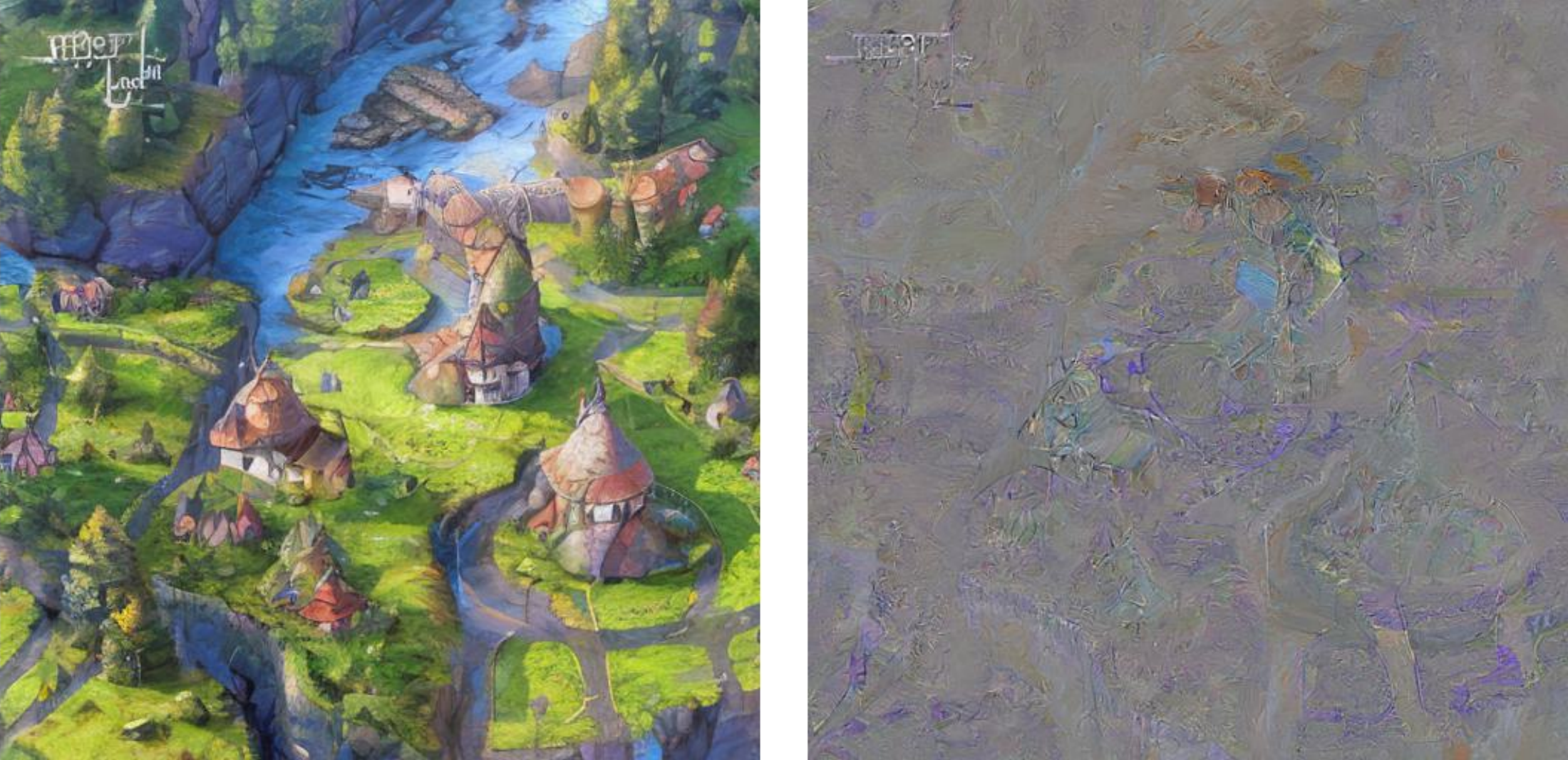}\\\end{tabular}      & \begin{tabular}{c}\includegraphics[trim=544 0 0 0, clip, width=0.230\linewidth]{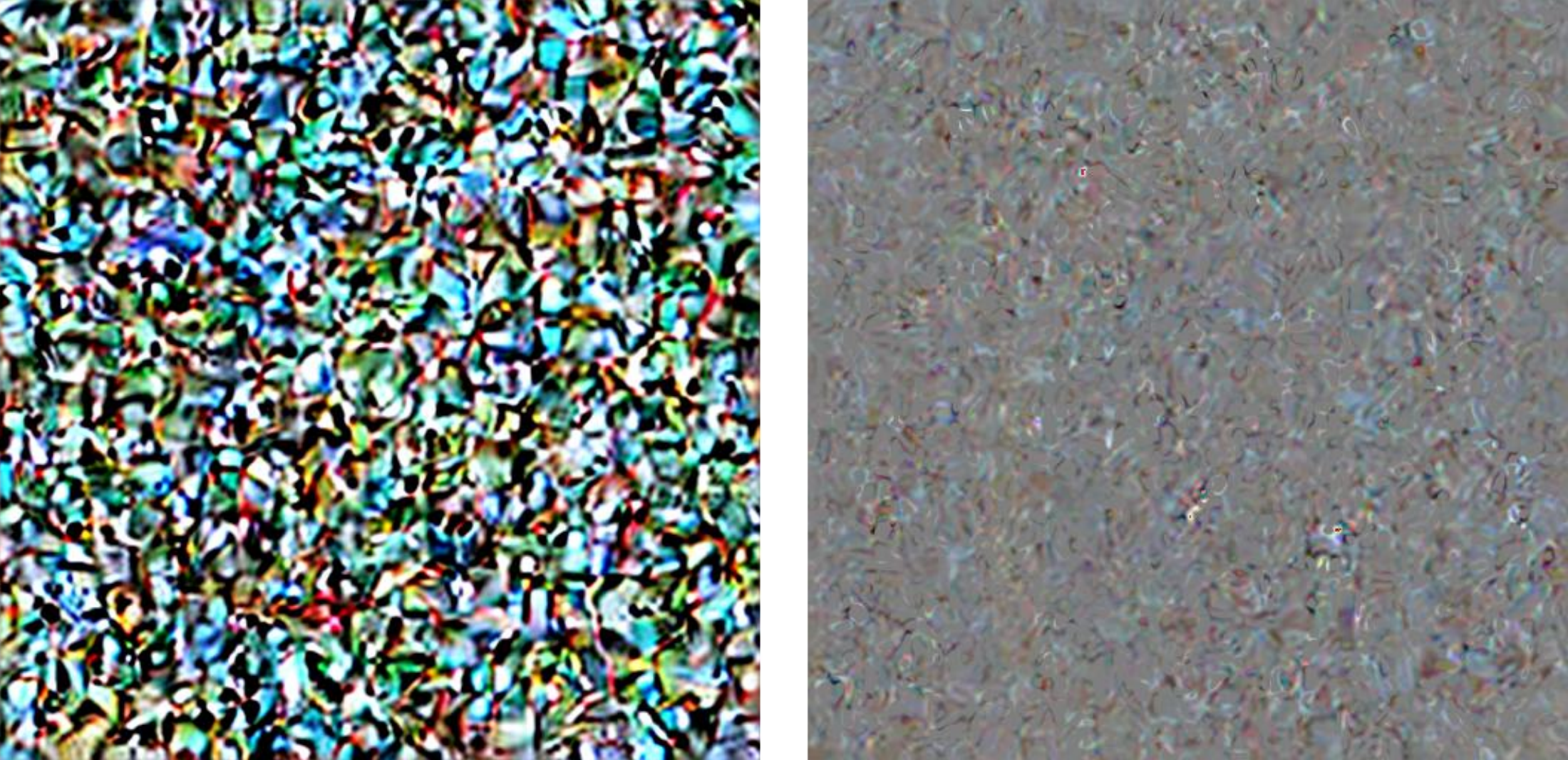}\\\end{tabular}     \\
     & \cellcolor{bBlue}\rotatebox[origin=c]{90}{\small Alg. 1 / 50}                          & \begin{tabular}{c}\includegraphics[trim=68 0 0 0, clip, width=0.230\linewidth]{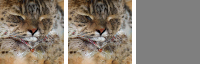}\\[-\dp\strutbox]\end{tabular}      & \begin{tabular}{c}\includegraphics[trim=68 0 0 0, clip, width=0.230\linewidth]{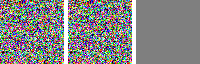}\\[-\dp\strutbox]\end{tabular}  & \begin{tabular}{c}\includegraphics[trim=544 0 0 0, clip, width=0.230\linewidth]{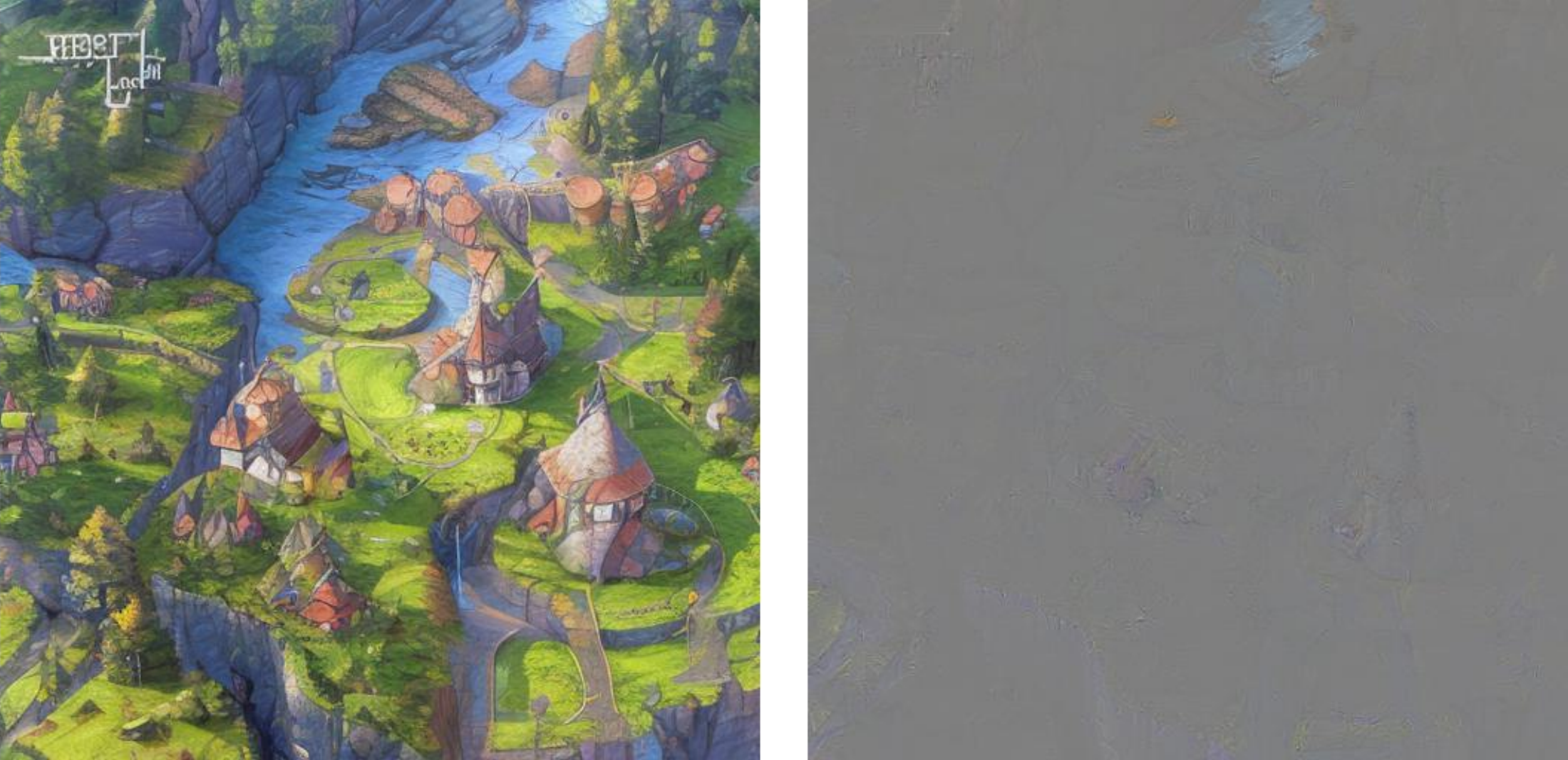}\\[-\dp\strutbox]\end{tabular}      & \begin{tabular}{c}\includegraphics[trim=544 0 0 0, clip, width=0.230\linewidth]{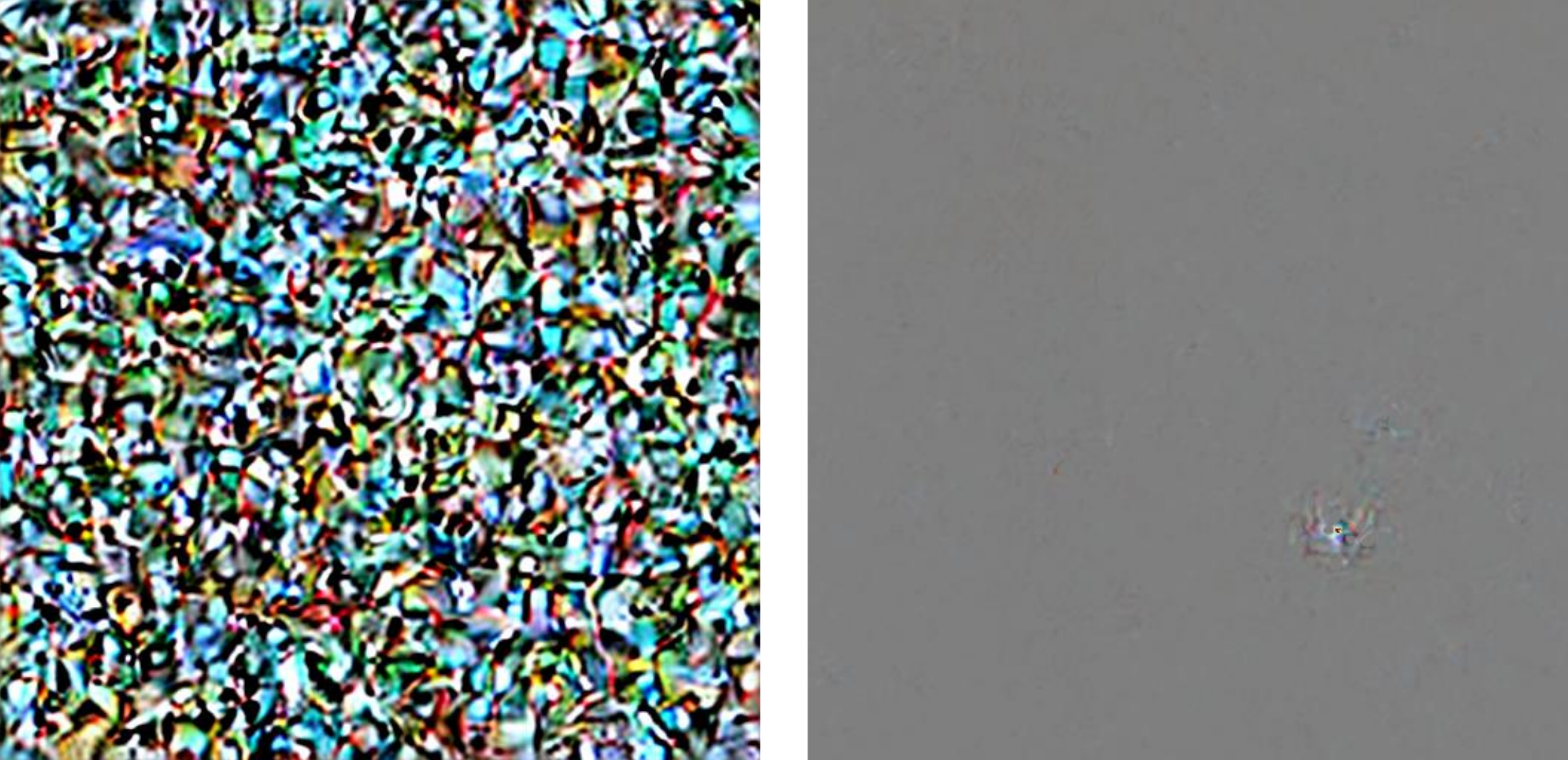}\\[-\dp\strutbox]\end{tabular}     \\
      \cmidrule(l){3-4} \cmidrule(l){5-6}
     \multirow{3}{*}{\rotatebox[origin=c]{90}{\small DPM-Solver++(2M) 10 steps}} & \rotatebox[origin=c]{90}{
          
     \cellcolor{bGray}\small na\"ive / 1000 }         & \begin{tabular}{c}\includegraphics[trim=68 0 0 0, clip, width=0.230\linewidth]{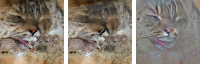}\end{tabular}      & \begin{tabular}{c}\includegraphics[trim=68 0 0 0, clip, width=0.230\linewidth]{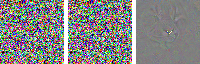}\end{tabular}   & \begin{tabular}{c}\includegraphics[trim=544 0 0 0, clip, width=0.230\linewidth]{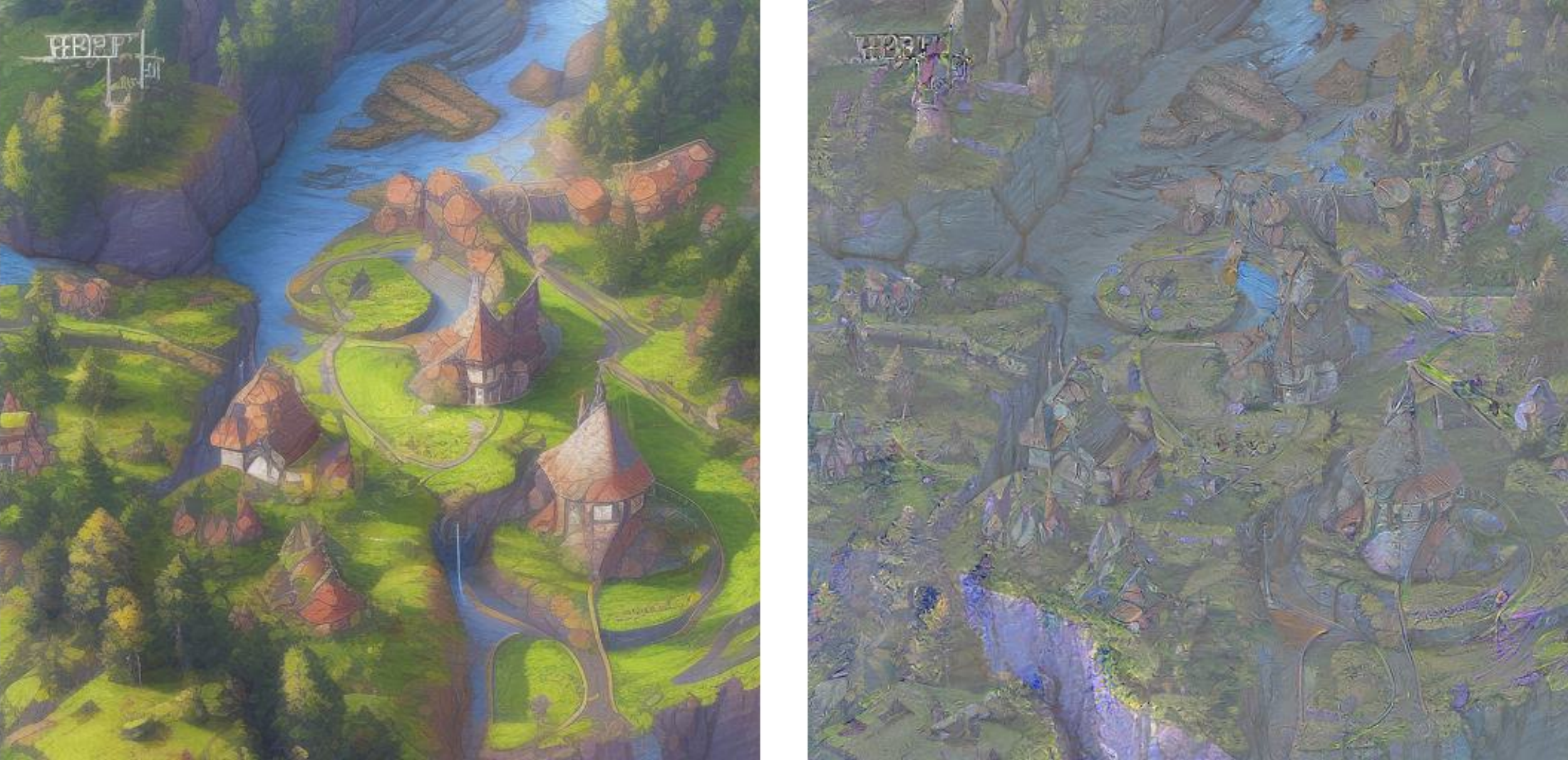}\end{tabular}      & \begin{tabular}{c}\includegraphics[trim=544 0 0 0, clip, width=0.230\linewidth]{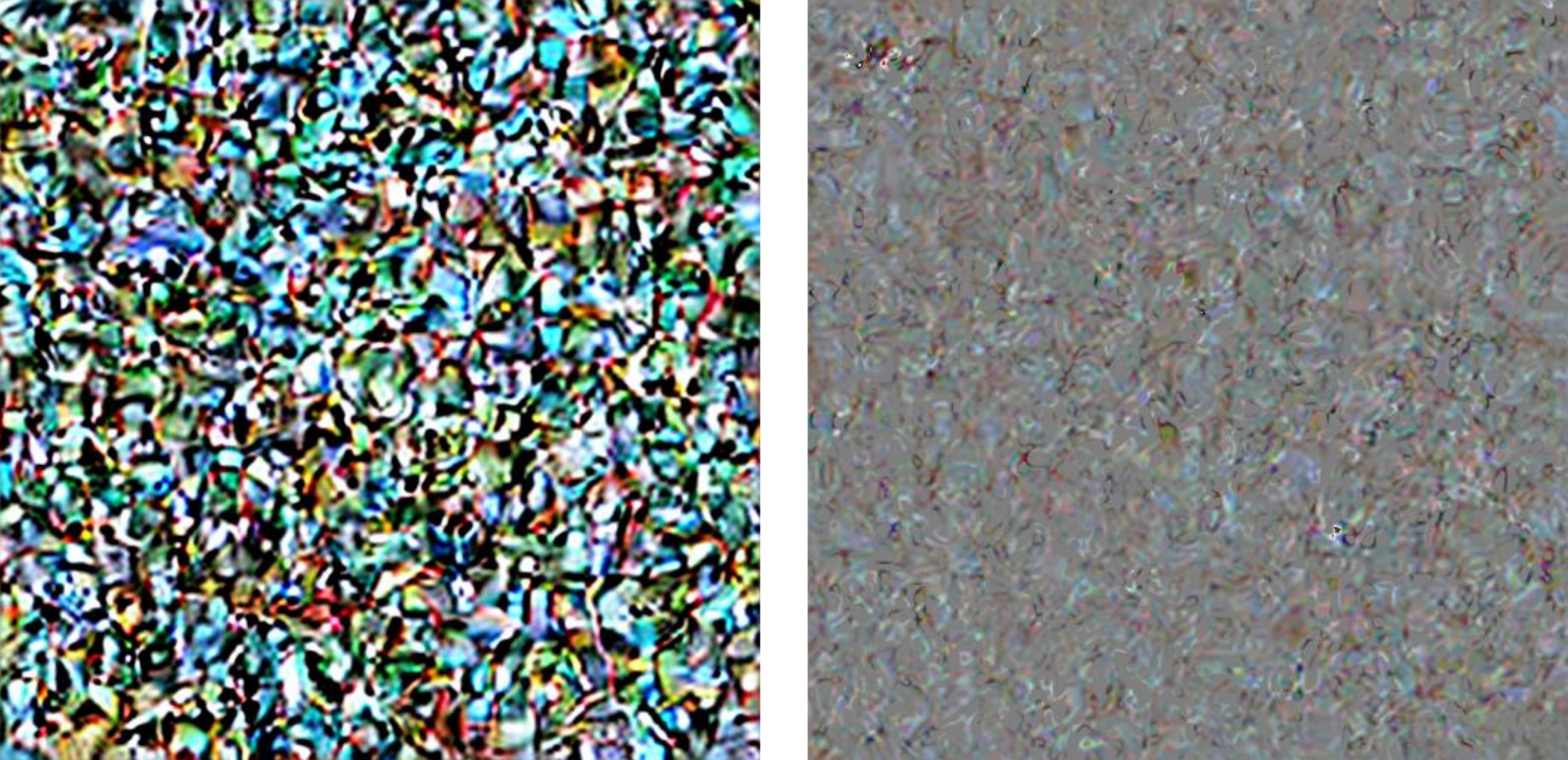}\end{tabular}    \\
     & \cellcolor{bBlue}{\rotatebox[origin=c]{90}{\small Alg. 1 / 50}}                              & \begin{tabular}{c}\includegraphics[trim=68 0 0 0, clip, width=0.230\linewidth]{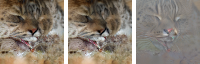}\end{tabular}      & \begin{tabular}{c}\includegraphics[trim=68 0 0 0, clip, width=0.230\linewidth]{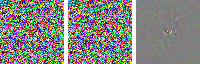}\end{tabular}  & \begin{tabular}{c}\includegraphics[trim=544 0 0 0, clip, width=0.230\linewidth]{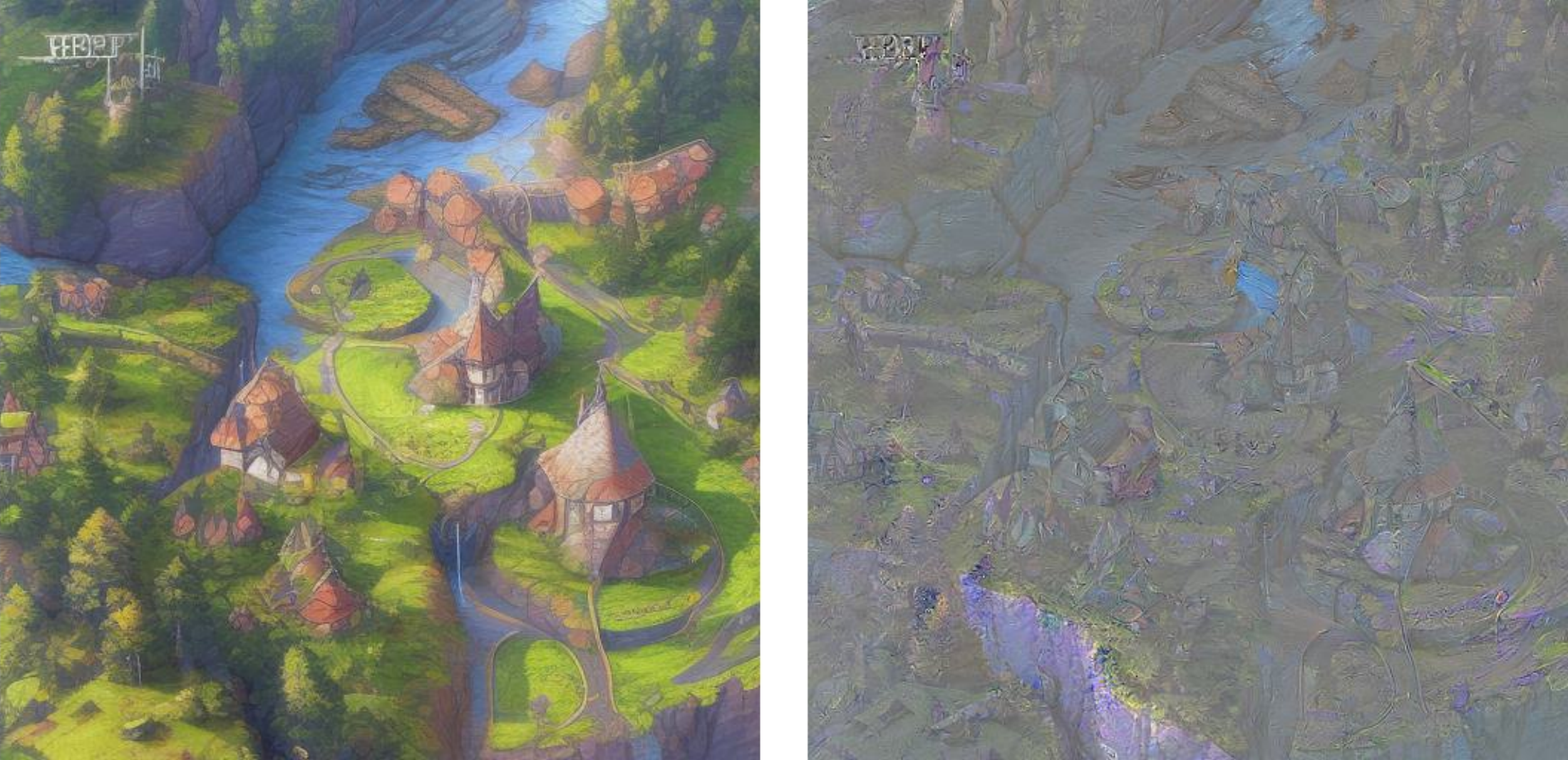}\end{tabular}      & \begin{tabular}{c}\includegraphics[trim=544 0 0 0, clip, width=0.230\linewidth]{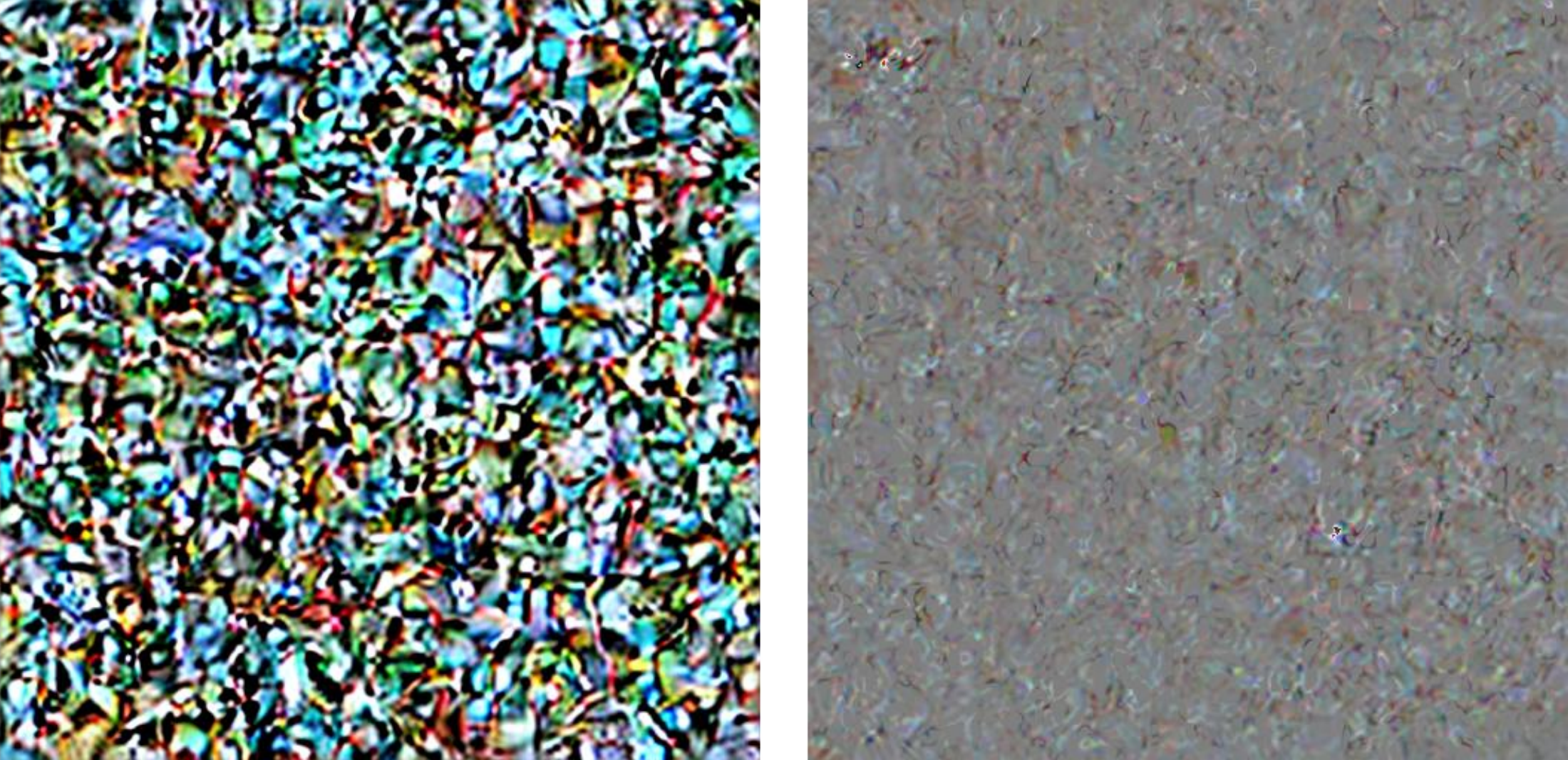}\end{tabular}     \\ 
     & \rotatebox[origin=c]{90}{\cellcolor{bRed}{\small Alg. 2 / 10}}                             & \begin{tabular}{c}\includegraphics[trim=68 0 0 0, clip, width=0.230\linewidth]{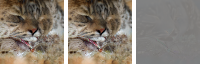}\\[-\dp\strutbox]\end{tabular}      & \begin{tabular}{c}\includegraphics[trim=68 0 0 0, clip, width=0.230\linewidth]{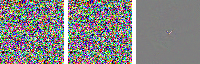}\\[-\dp\strutbox]\end{tabular}    & \begin{tabular}{c}\includegraphics[trim=544 0 0 0, clip, width=0.230\linewidth]{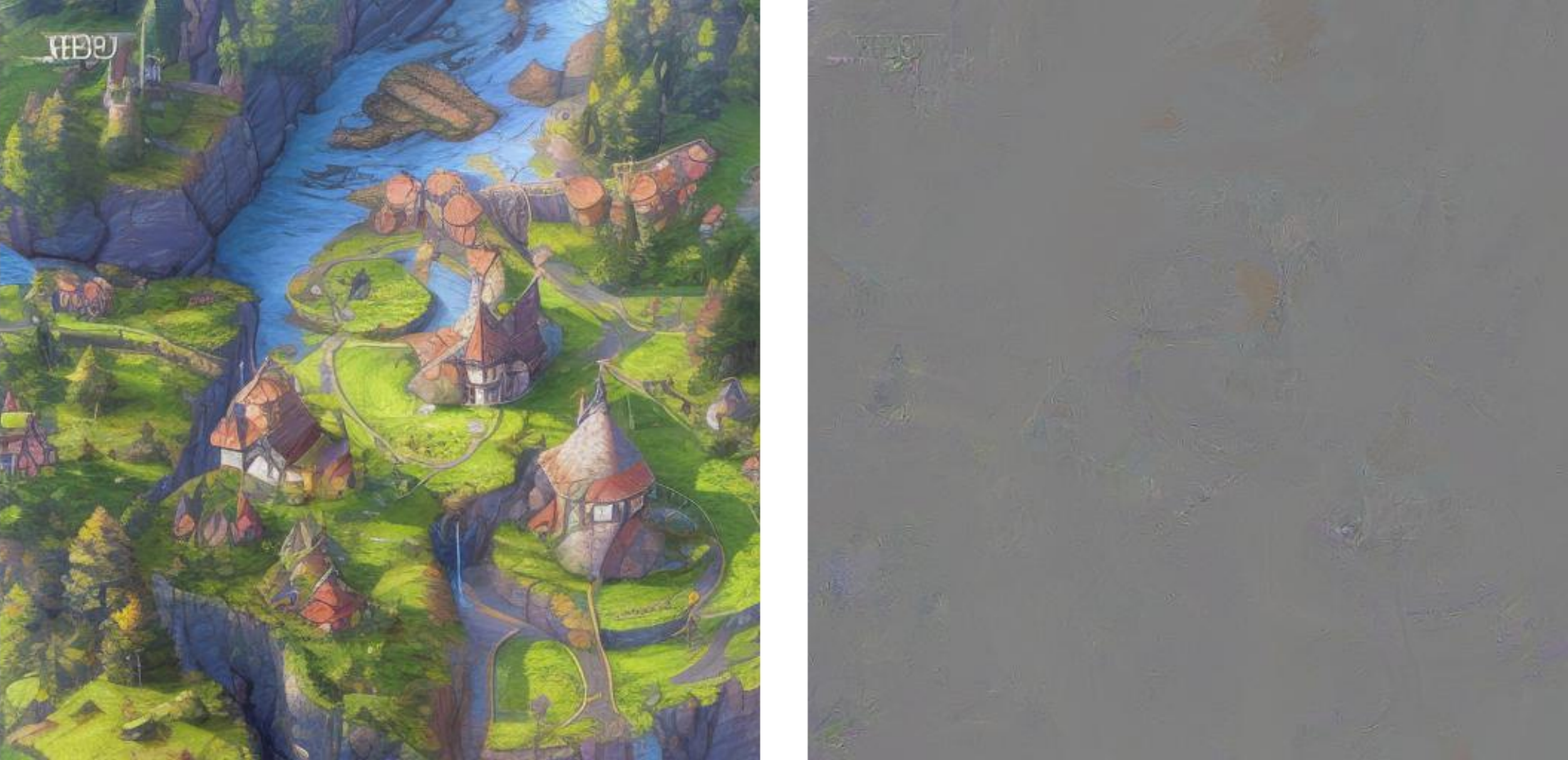}\\[-\dp\strutbox]\end{tabular}      & \begin{tabular}{c}\includegraphics[trim=544 0 0 0, clip, width=0.230\linewidth]{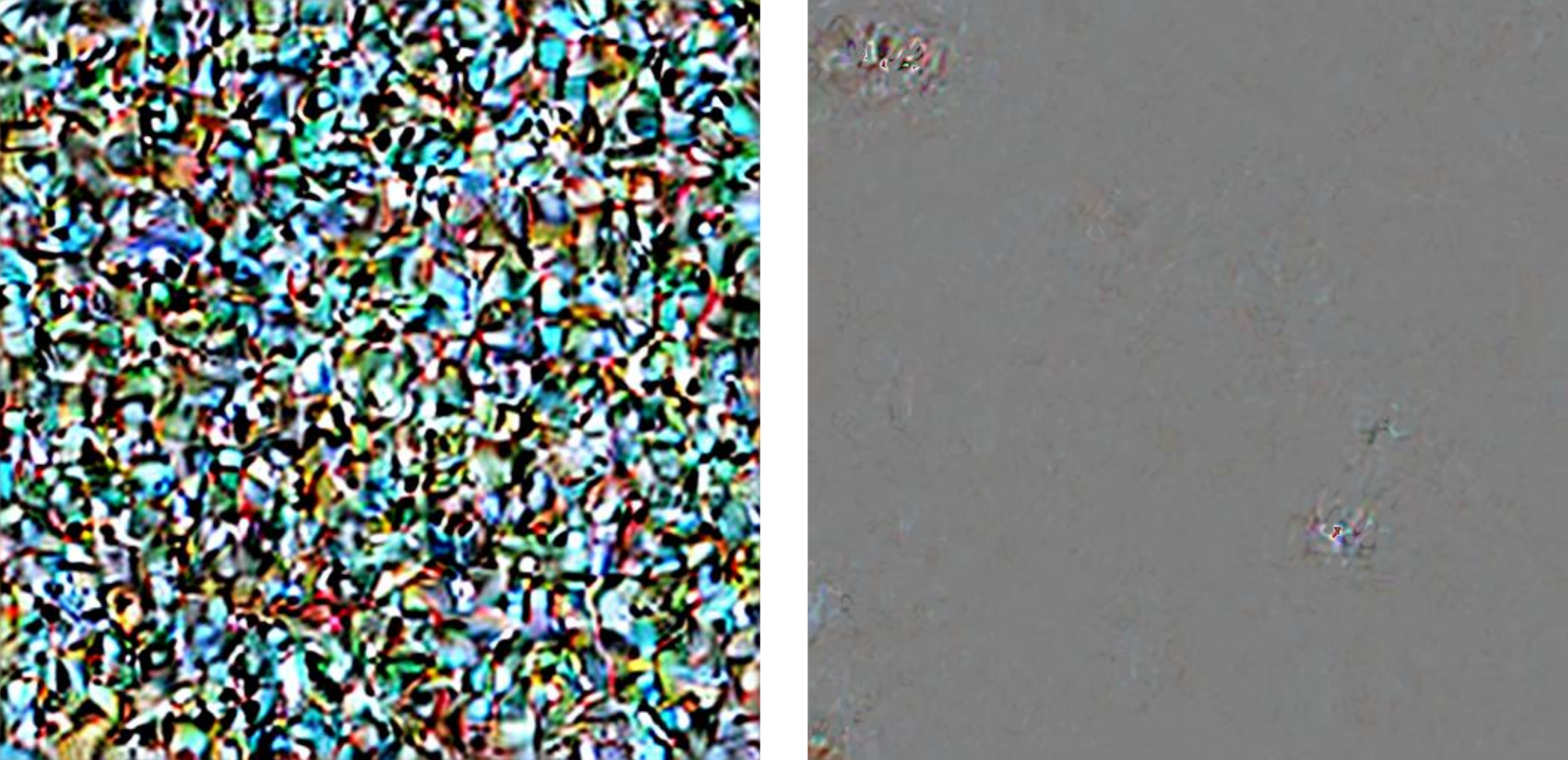}\\[-\dp\strutbox] \end{tabular}   \\
    \end{tabular}
    \caption{Our Algs. 1 and 2 significantly reduce reconstruction errors, whether it's for images or noise, DDIM or high-order DPM-solvers, or pixel-space DPM or LDM. The generation / inversion method varies for each row, \textit{e.g.}, `na\"ive / 1000' indicates that we performed the na\"ive DDIM inversion for 1000 steps. `Alg. 1 / 50' and `Alg. 2 / 10' attempt exact inversion with 50 steps of DDIM and 10 steps of DPM-Solver++(2M), respectively. Achieving exact inversion in LDM is challenging due to information loss from the autoencoder and instability caused by a classifier-free guidance of $3.0$. Nonetheless, our algorithm produces good results (extremely low residual error in Alg. 1 for DDIM on the 3rd row and in Alg. 2 for DPM-solver++(2M) on the 6th row) also in LDM.}
    \label{sfig:recon_qualitative}
\end{figure*}

In Sec. 5.1 of the main paper, we performed the reconstruction of noise and image to evaluate the exact invertibility of the proposed methods.
We provide more qualitative results in \cref{sfig:recon_qualitative}. In \cref{sfig:recon_qualitative}, we addtionally provide the FPI-based method of Pan et al. ~\cite{Pan_2023_ICCV}, namely AIDI-E. 

\subsection{Application: Tree-ring watermark}\label{sec:add:WM}
In Sec. 5.2 of the main paper, we demonstrated the improved detection of watermarks~\cite{wen2023tree} by employing our algorithm, even when the images were generated using high-order DPM-solvers.
We provide an additional example of watermark detection / classification in Figs. \cref{fig:watermarks} and \cref{fig:confusions}. We experimented with different shaped watermarks and prompts from the main paper.

\begin{figure*}[t]
    \small
    \centering
    \setlength{\tabcolsep}{0pt}
    \begin{tabular}{@{}c@{ }c@{ }c@{ }c@{ }c@{ }c@{ }c@{ }c@{}}
     \begin{tabular}{@{}c@{}}Embedded\\ watermark\end{tabular} & \begin{tabular}{@{}c@{}}Generated by\\ DPM-Solver++\end{tabular} & \multicolumn{2}{c@{ }}{\cellcolor{bGray} \begin{tabular}{@{}c@{ }} na\"ive DDIM inversion \\ (Recon. / Error)\end{tabular}} &  \multicolumn{2}{c@{ }}{\cellcolor{bGray} \begin{tabular}{@{}c@{}} na\"ive DDIM inversion w/ $\mathcal{D}^\dagger$ \\ (Recon. / Error) \end{tabular}} &  \multicolumn{2}{c@{}}{\cellcolor{bRed} \begin{tabular}{@{}c@{}}Algorithm 2 (ours) \\ (Recon. / Error) \end{tabular}}  \\
    
    \includegraphics[width=0.120\linewidth]{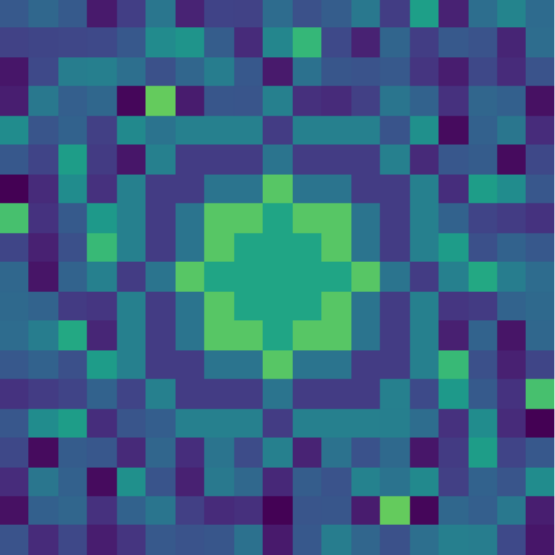}
     & \includegraphics[width=0.120\linewidth]{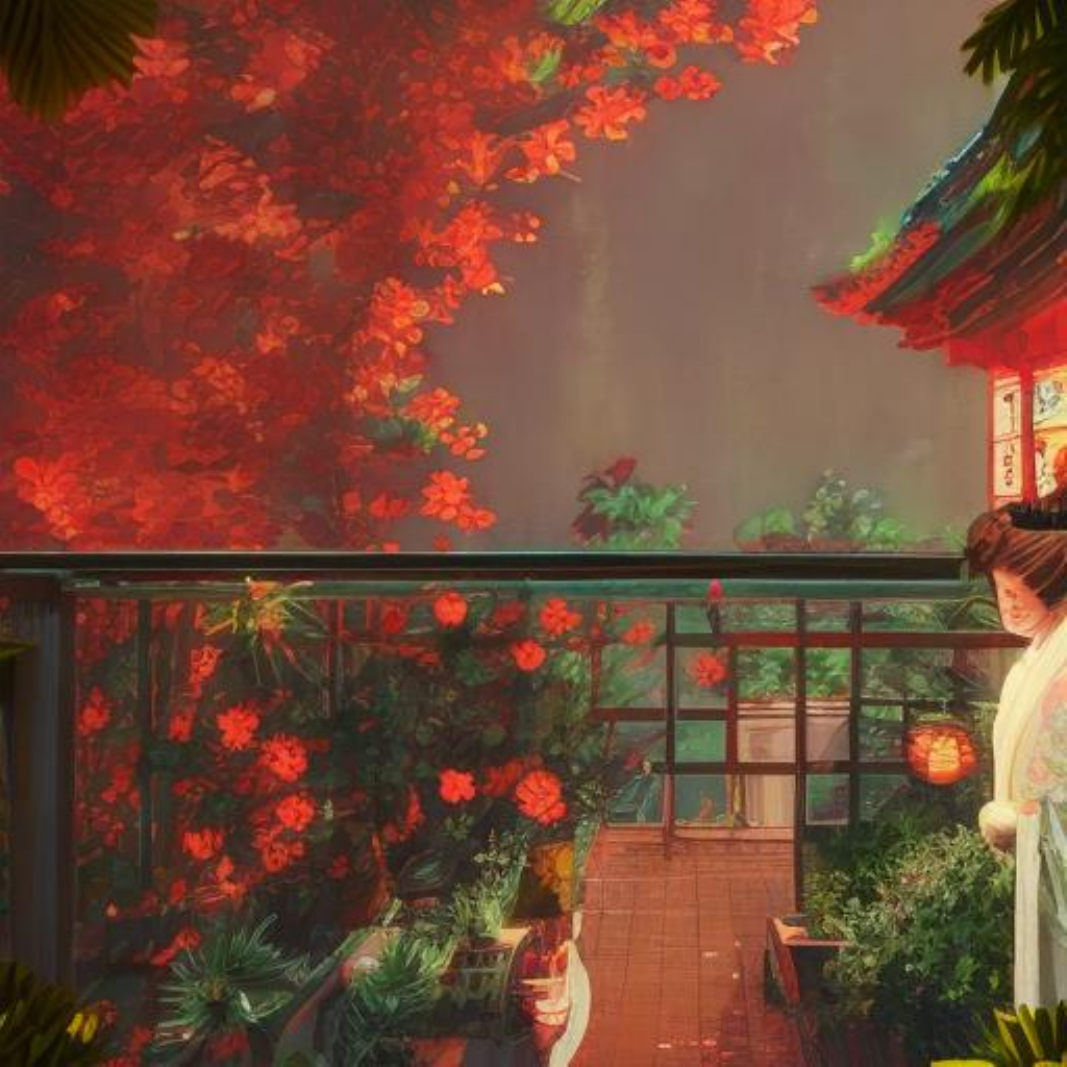} & \includegraphics[width=0.120\linewidth]{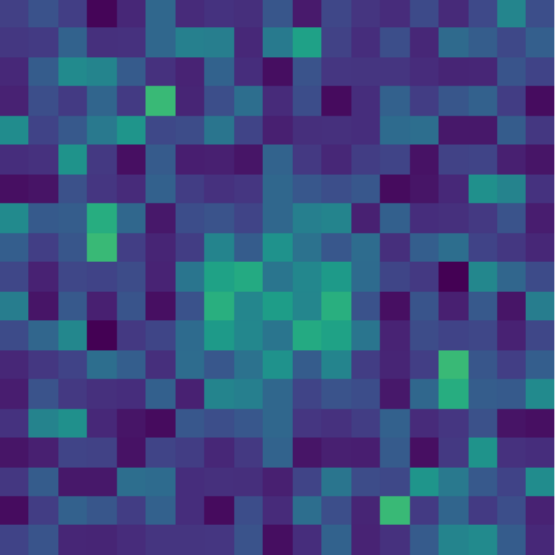} & \includegraphics[width=0.120\linewidth]{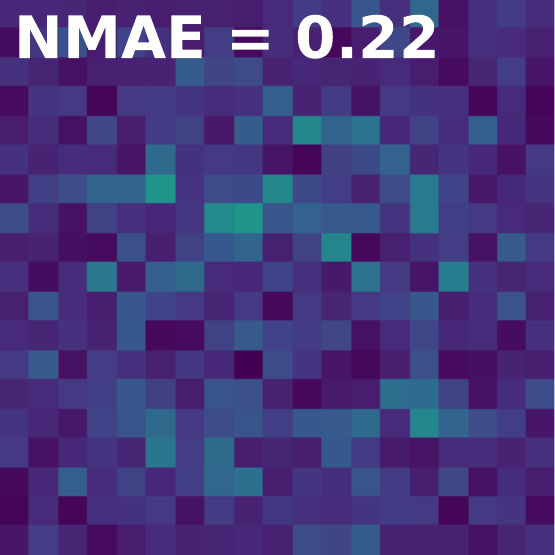} & \includegraphics[width=0.120\linewidth]{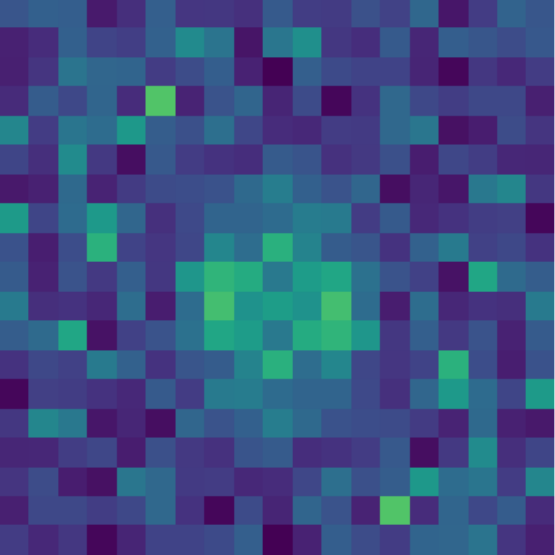} & \includegraphics[width=0.120\linewidth]{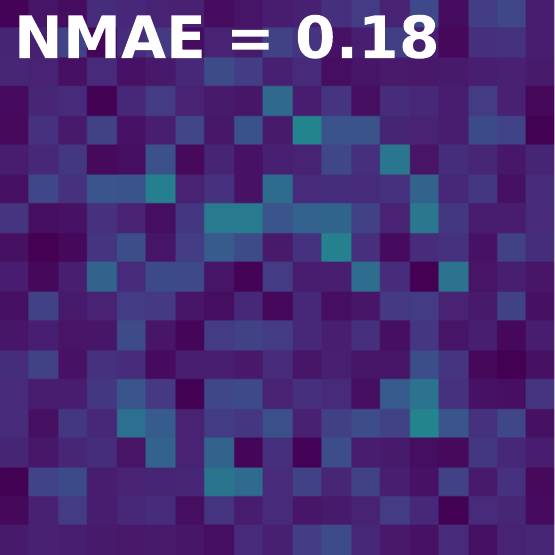} & \includegraphics[width=0.120\linewidth]{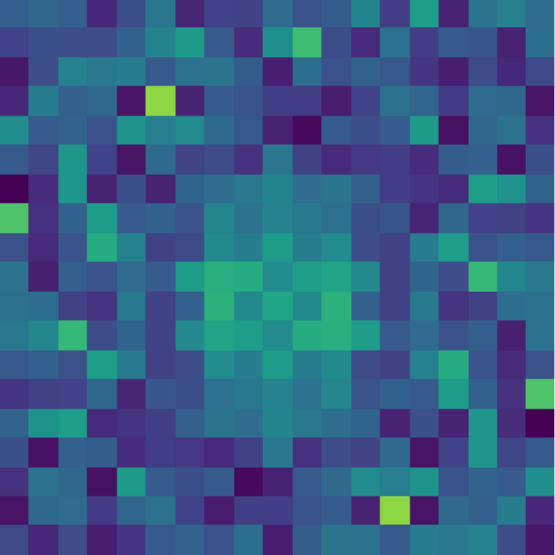} & \includegraphics[width=0.120\linewidth]{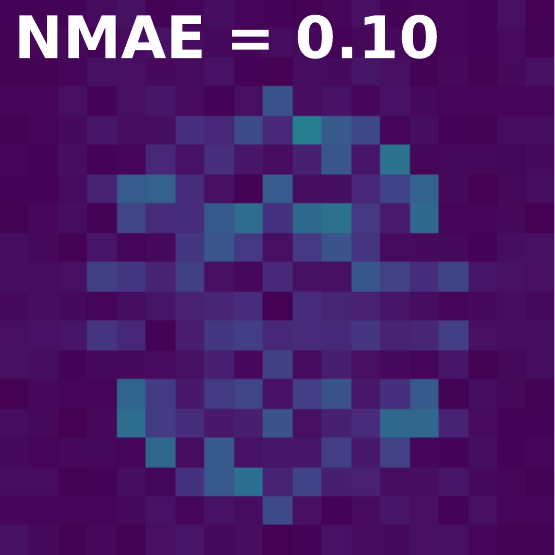} \\
     
     \includegraphics[width=0.120\linewidth]{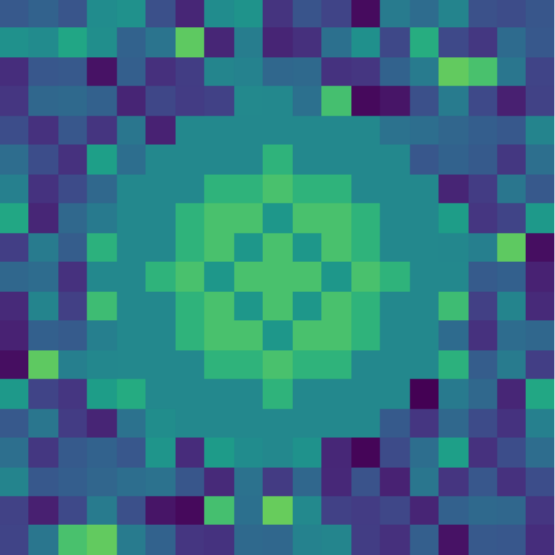}
     & \includegraphics[width=0.120\linewidth]{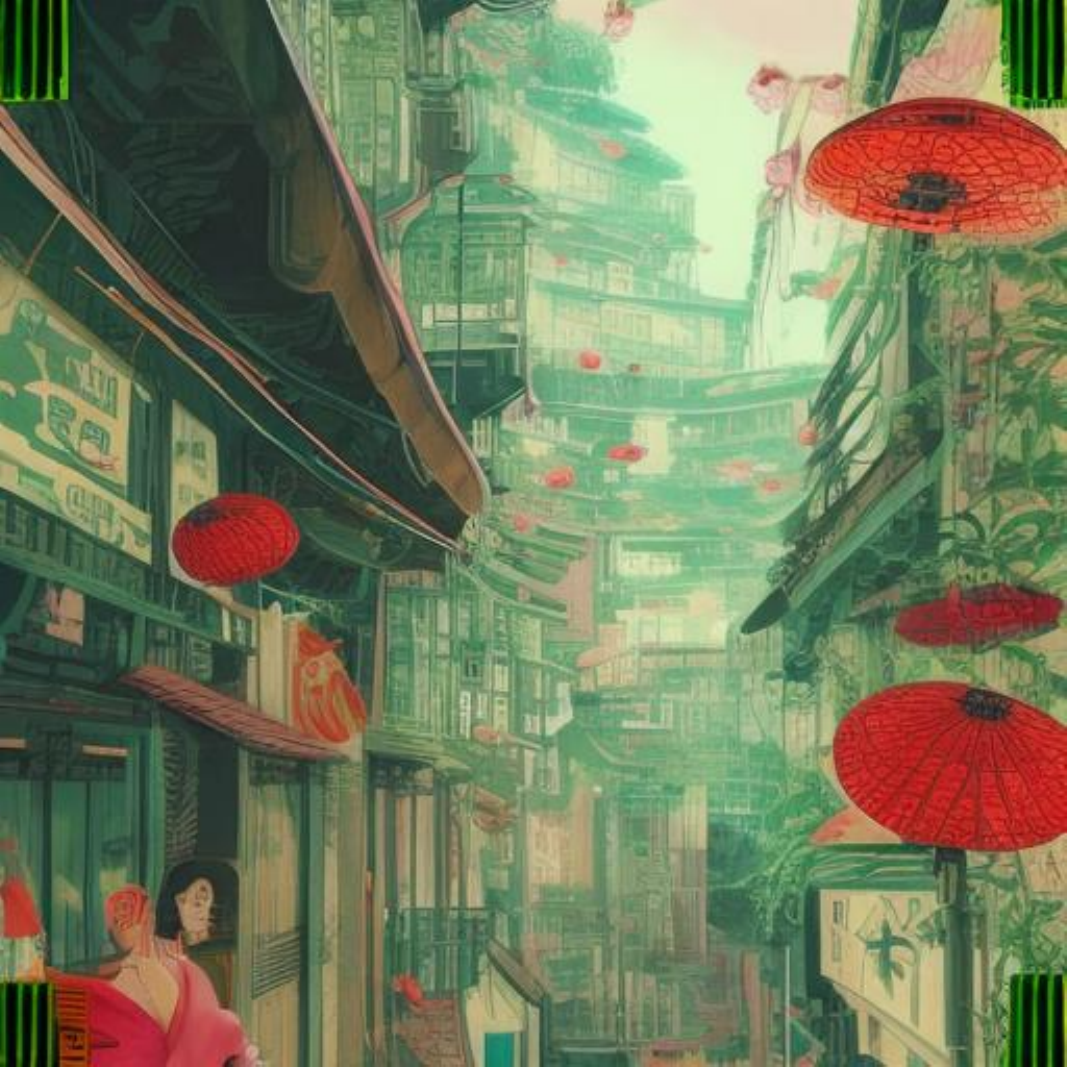} & \includegraphics[width=0.120\linewidth]{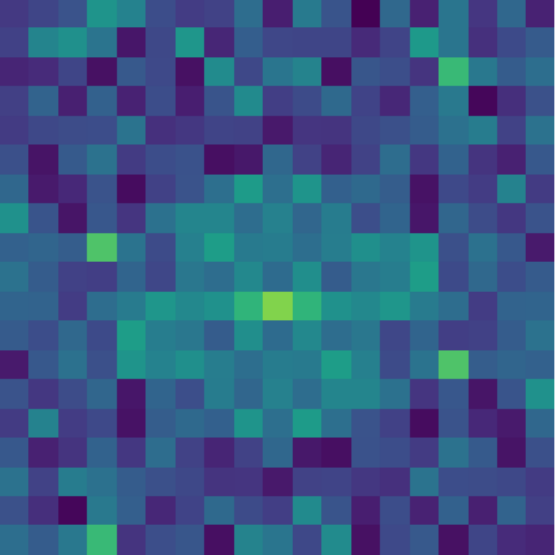} & \includegraphics[width=0.120\linewidth]{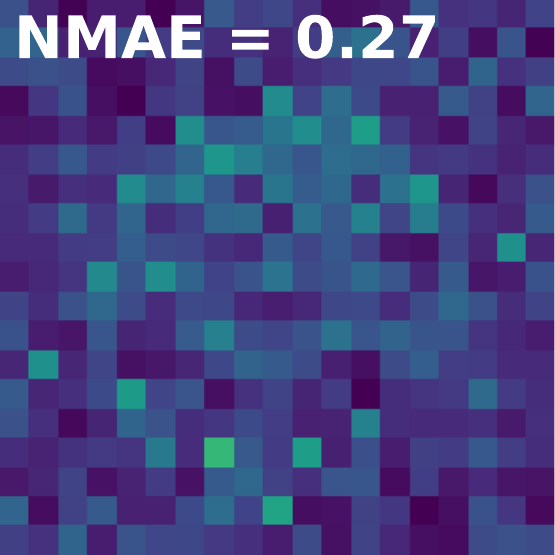} & \includegraphics[width=0.120\linewidth]{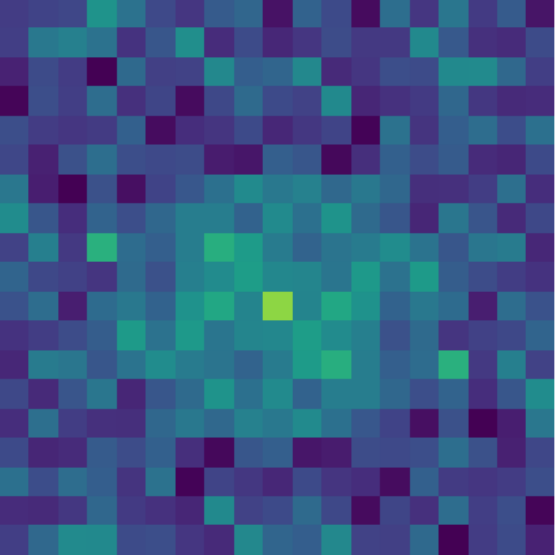} & \includegraphics[width=0.120\linewidth]{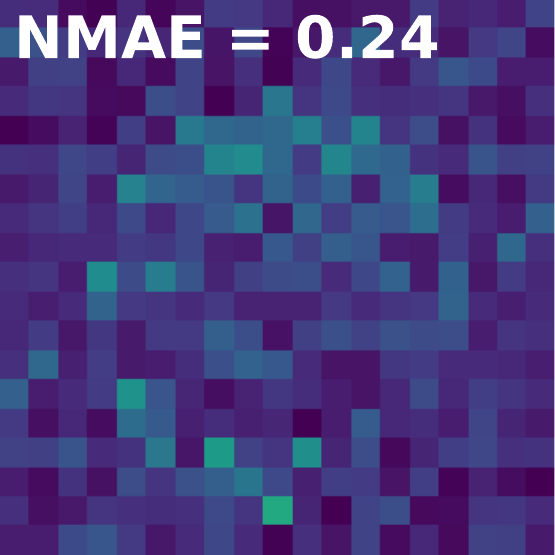} & \includegraphics[width=0.120\linewidth]{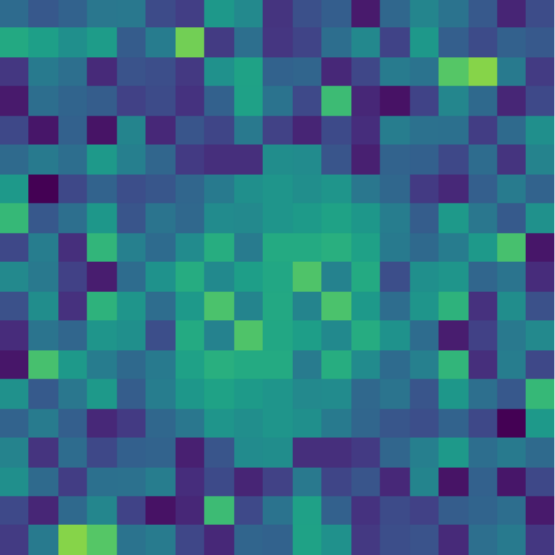} & \includegraphics[width=0.120\linewidth]{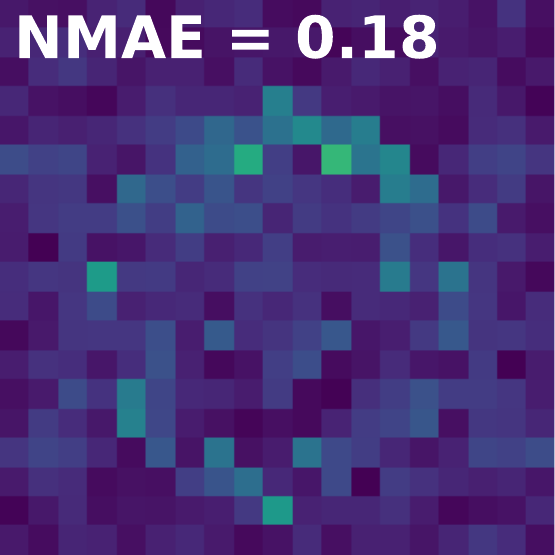} \\
     
     \includegraphics[width=0.120\linewidth]{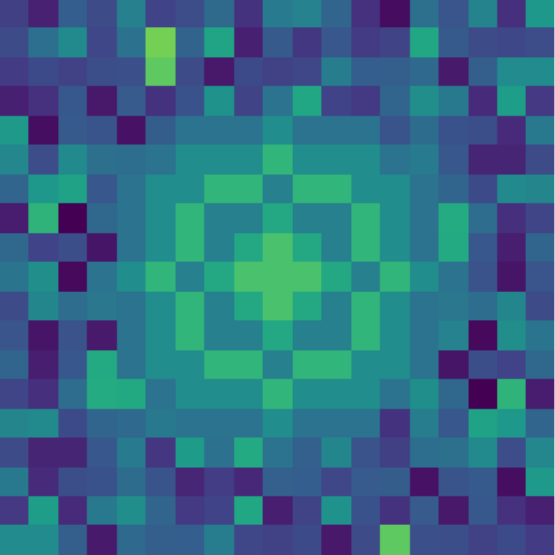}
     & \includegraphics[width=0.120\linewidth]{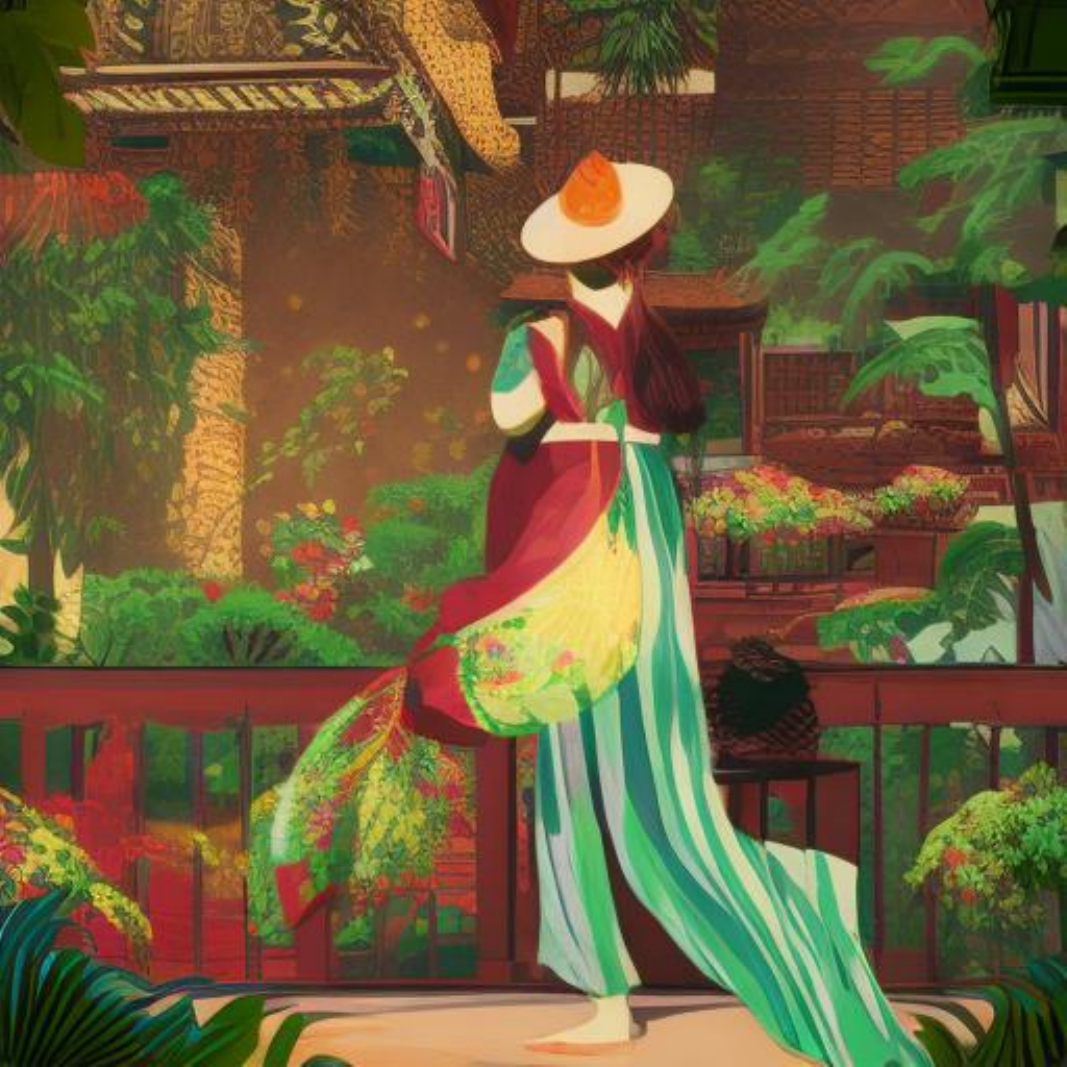} & \includegraphics[width=0.120\linewidth]{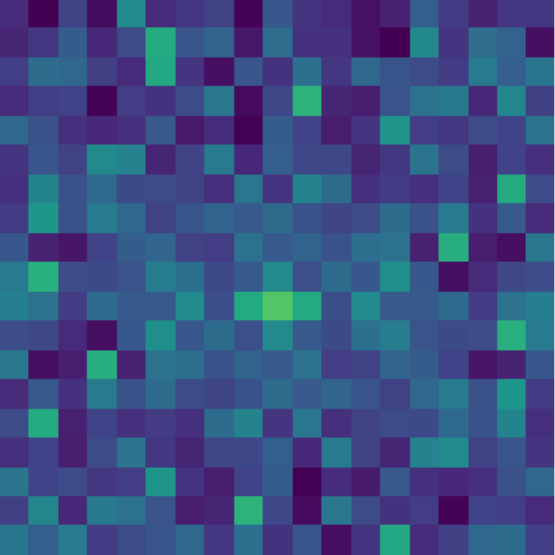} & \includegraphics[width=0.120\linewidth]{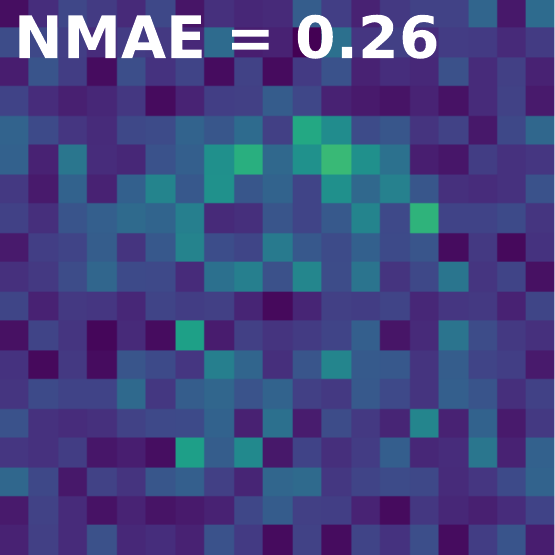} & \includegraphics[width=0.120\linewidth]{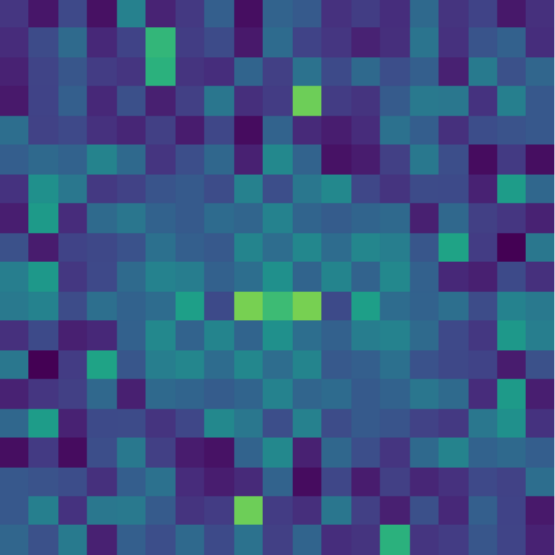} & \includegraphics[width=0.120\linewidth]{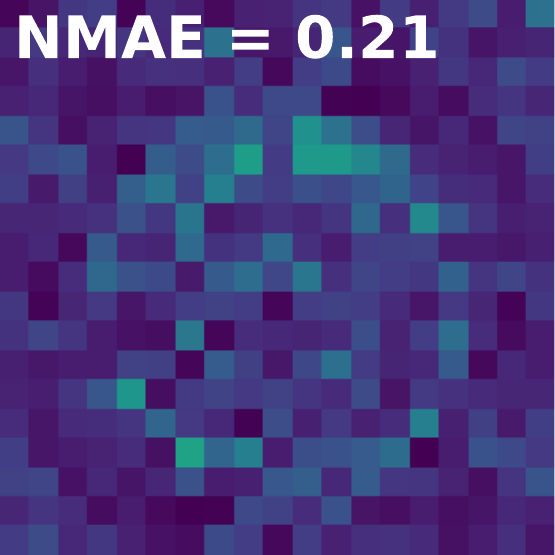} & \includegraphics[width=0.120\linewidth]{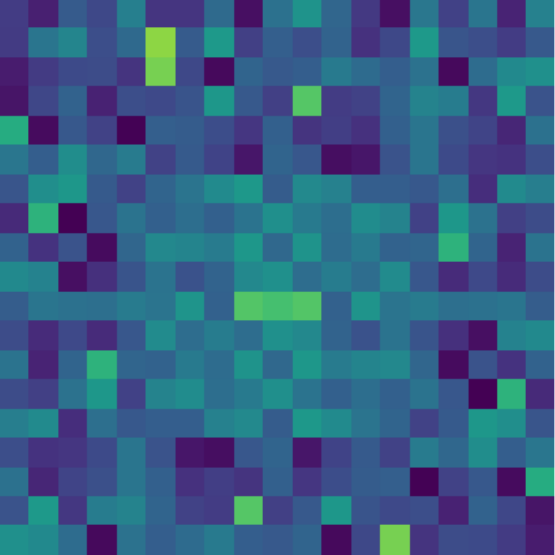} & \includegraphics[width=0.120\linewidth]{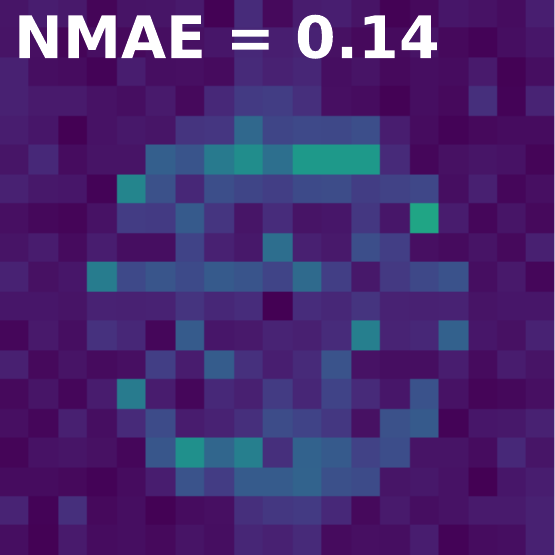} \\
    \end{tabular}
    \caption{
    Our Algorithm 2 enables accurate reconstruction of Tree-ring watermarks~\cite{wen2023tree} in the Fourier space of the initial noise ($\vz_T)$. The Tree-ring watermark is embedded in the Fourier space of the initial noise in the shape of tree-rings and can be utilized for copyright tracing (column 1). Then, the image is generated starting from the watermarked noise. The practical approach is to accelerate image generation using methods like DPM-Solver++(2M)~\cite{lu2022dpm++} (column 2). NMAEs are shown on each error map. Using Algorithm 2 (columns 7-8) for watermark reconstruction results in lower errors compared to employing na\"ive DDIM inversion (columns 3-6), achieving nearly 50\% reduction in NMAE. 
    }
    \label{fig:watermarks}
\end{figure*}

\begin{figure*}[t]
    \small
    \centering   
    \begin{subfigure}[]{0.3\textwidth}
    \centering
        \begin{tikzpicture}
            \begin{axis}[
                    width=0.9\linewidth,  
                    height=0.625\linewidth,  
                    colormap={custom}{
                        color(0cm)=(white);
                        color(1cm)=(yellow);
                        color(2cm)=(green);
                        color(3cm)=(green);
                    },
                    xlabel=Predicted,
                    xlabel style={yshift=0pt},
                    ylabel=Actual,
                    ylabel style={yshift=3pt},
                    xticklabels={WM 1, WM 2, WM 3},
                    xtick={0,...,2},
                    xtick style={draw=none},
                    yticklabels={WM 1, WM 2, WM 3},
                    ytick={0,...,2},
                    ytick style={draw=none},
                    enlargelimits=false,
                    xticklabel style={},
                    nodes near coords={\pgfmathprintnumber\pgfplotspointmeta},
                    nodes near coords style={
                        yshift=-7pt
                    },
                ]
                \addplot[
                    matrix plot,
                    mesh/cols=3,
                    point meta=explicit,draw=gray
                ] table [meta=C] {
                    x y C
                    0 0 99
                    1 0 1
                    2 0 0

                    0 1 44
                    1 1 43
                    2 1 13

                    0 2 20
                    1 2 16
                    2 2 64

                };
            \end{axis}
        \end{tikzpicture}
    \caption{Na\"ive DDIM inversion}
    \label{fig:confusion_naive}
    \end{subfigure}
    \hfill
    \begin{subfigure}[]{0.3\textwidth}
        \centering
        \begin{tikzpicture}
            \begin{axis}[
                    width=0.9\linewidth,  
                    height=0.625\linewidth,  
                    colormap={custom}{
                        color(0cm)=(white);
                        color(1cm)=(yellow);
                        color(2cm)=(green);
                        color(3cm)=(green);
                    },
                    xlabel=Predicted,
                    xlabel style={yshift=0pt},
                    ylabel=Actual,
                    ylabel style={yshift=3pt},
                    xticklabels={WM 1, WM 2, WM 3},
                    xtick={0,...,2},
                    xtick style={draw=none},
                    yticklabels={WM 1, WM 2, WM 3},
                    ytick={0,...,2},
                    ytick style={draw=none},
                    enlargelimits=false,
                    xticklabel style={},
                    nodes near coords={\pgfmathprintnumber\pgfplotspointmeta},
                    nodes near coords style={
                        yshift=-7pt
                    },
                ]
                \addplot[
                    matrix plot,
                    mesh/cols=3,
                    point meta=explicit,draw=gray
                ] table [meta=C] {
                    x y C
                    0 0 100
                    1 0 0
                    2 0 0

                    0 1 44
                    1 1 37
                    2 1 19

                    0 2 18
                    1 2 15
                    2 2 67

                };
            \end{axis}
        \end{tikzpicture}
    \caption{na\"ive DDIM inversion w/ $\mathcal{D}^\dagger$ in Sec. 4.1 }
    \label{fig:confusion_naive+}
    \end{subfigure}
    \hfill
    \begin{subfigure}[]{0.375\textwidth}
    \centering
        \begin{tikzpicture}
            \begin{axis}[
                    width=0.72\linewidth,  
                    height=0.50\linewidth,  
                    colormap={custom}{
                        color(0cm)=(white);
                        color(1cm)=(yellow);
                        color(2cm)=(green);
                        color(3cm)=(green);
                    },
                    xlabel=Predicted,
                    xlabel style={yshift=0pt},
                    ylabel=Actual,
                    ylabel style={yshift=3pt},
                    xticklabels={WM 1, WM 2, WM 3},
                    xtick={0,...,2},
                    xtick style={draw=none},
                    yticklabels={WM 1, WM 2, WM 3},
                    ytick={0,...,2},
                    ytick style={draw=none},
                    enlargelimits=false,
                    xticklabel style={},
                    colorbar,
                    nodes near coords={\pgfmathprintnumber\pgfplotspointmeta},
                    nodes near coords style={
                        yshift=-7pt
                    },
                ]
                \addplot[
                    matrix plot,
                    mesh/cols=3,
                    point meta=explicit,draw=gray
                ] table [meta=C] {
                    x y C
                    0 0 100
                    1 0 0
                    2 0 0

                    0 1 8
                    1 1 49
                    2 1 43

                    0 2 5
                    1 2 14
                    2 2 81

                };
            \end{axis}
        \end{tikzpicture}
    \caption{Algorithm 2 (ours)}
    \label{fig:confusion_alg2}
    \end{subfigure}
    
    \caption{
    Our algorithm's strong reconstruction performance allows for the classification of tree-ring watermarks as well. For copyright tracing, it is possible to generate images by embedding different unique watermarks. Three distinct watermarks (WM 1,2, and 3) are displayed in the first column of \cref{fig:watermarks}. In the confusion matrices, `Predicted' corresponds to the watermark with the smallest $l_1$ difference among the three watermarks. In Figs. \ref{fig:confusion_naive} and \ref{fig:confusion_naive+}, the na\"ive DDIM inversion encounters difficulties in detecting WM 2. In contrast (\cref{fig:confusion_alg2}), our Algorithm 2 performs well in detecting WM 2.}
    \label{fig:confusions}
\end{figure*}

\subsection{Application: Background-preserving editing}\label{sec:add:edit}
\begin{figure*}
    \centerline{\includegraphics[width=0.99\linewidth]{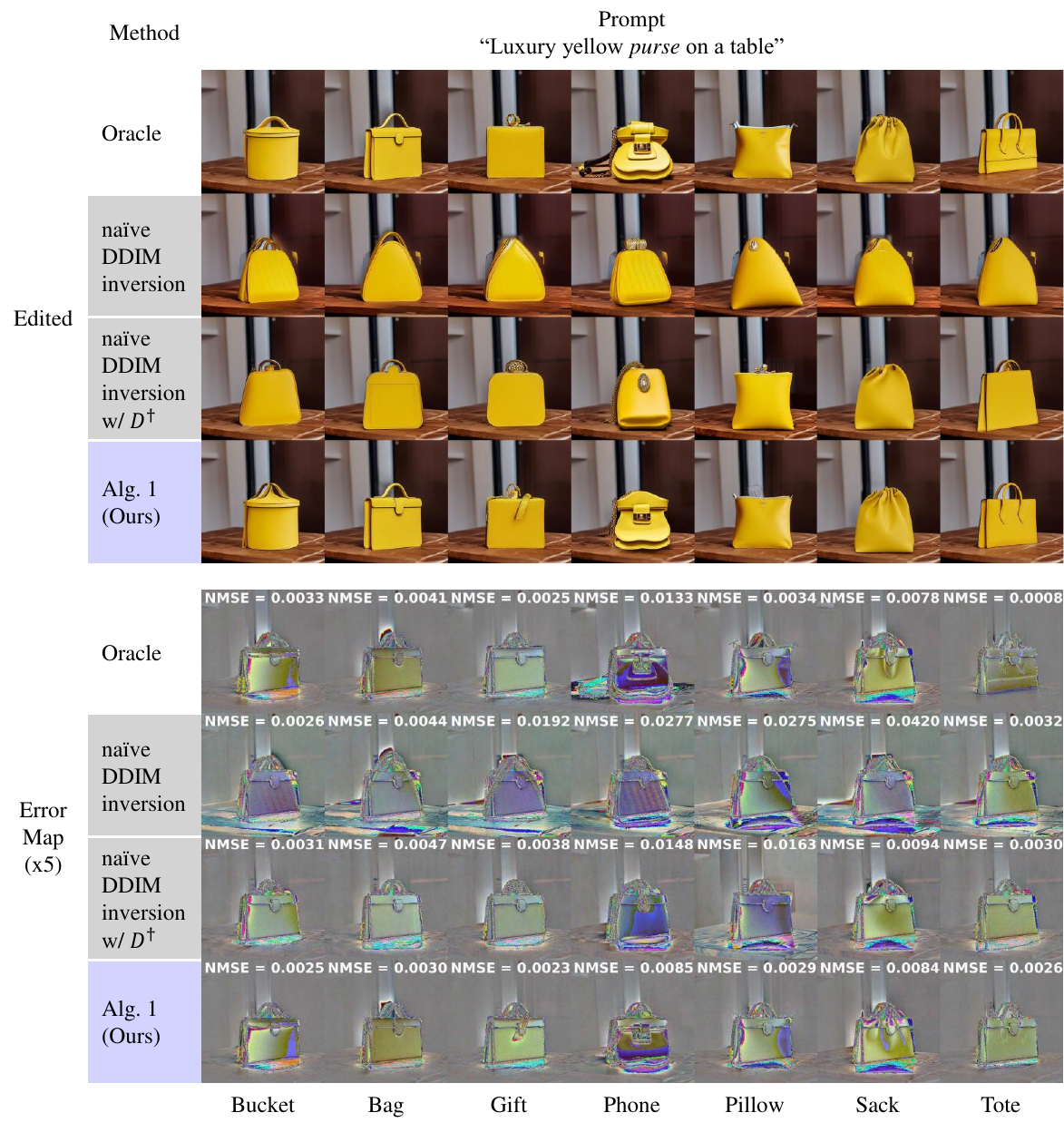}}
    \caption{Additional experiment result on background-preserving editing. The \emph{emphasized} word 
    indicates the object changed in the generated image, replaced with different words at the bottom of the figure (e.g., \emph{purse} becomes `bucket'). Our Alg. 1 preserves the background and allows diverse editing, even when the original image's path is unknown (i.e., $(\vz_{t_i})_{i=0}^M$). The first row (Oracle) shows results with the full trajectory, while subsequent rows use only the generated image (i.e., $\vx_0$). In these cases, we estimate the trajectory through each inversion method and edit based on the inversion results. Estimating the original trajectory using the basic DDIM inversion (rows 2-3) fails to keep the background (background in the error map is not gray) compared to the Oracle and doesn't consistently edit the `purse' as prompted. In contrast, using our Alg. 1 (row 4) preserves the background similarly to the Oracle (with the background in the error map being gray) while consistently editing the `purse' as prompted. Background NMSEs are inset.
    }
    \label{fig:edit_1}
\end{figure*}

\begin{figure*}
    \centerline{\includegraphics[width=0.99\linewidth]{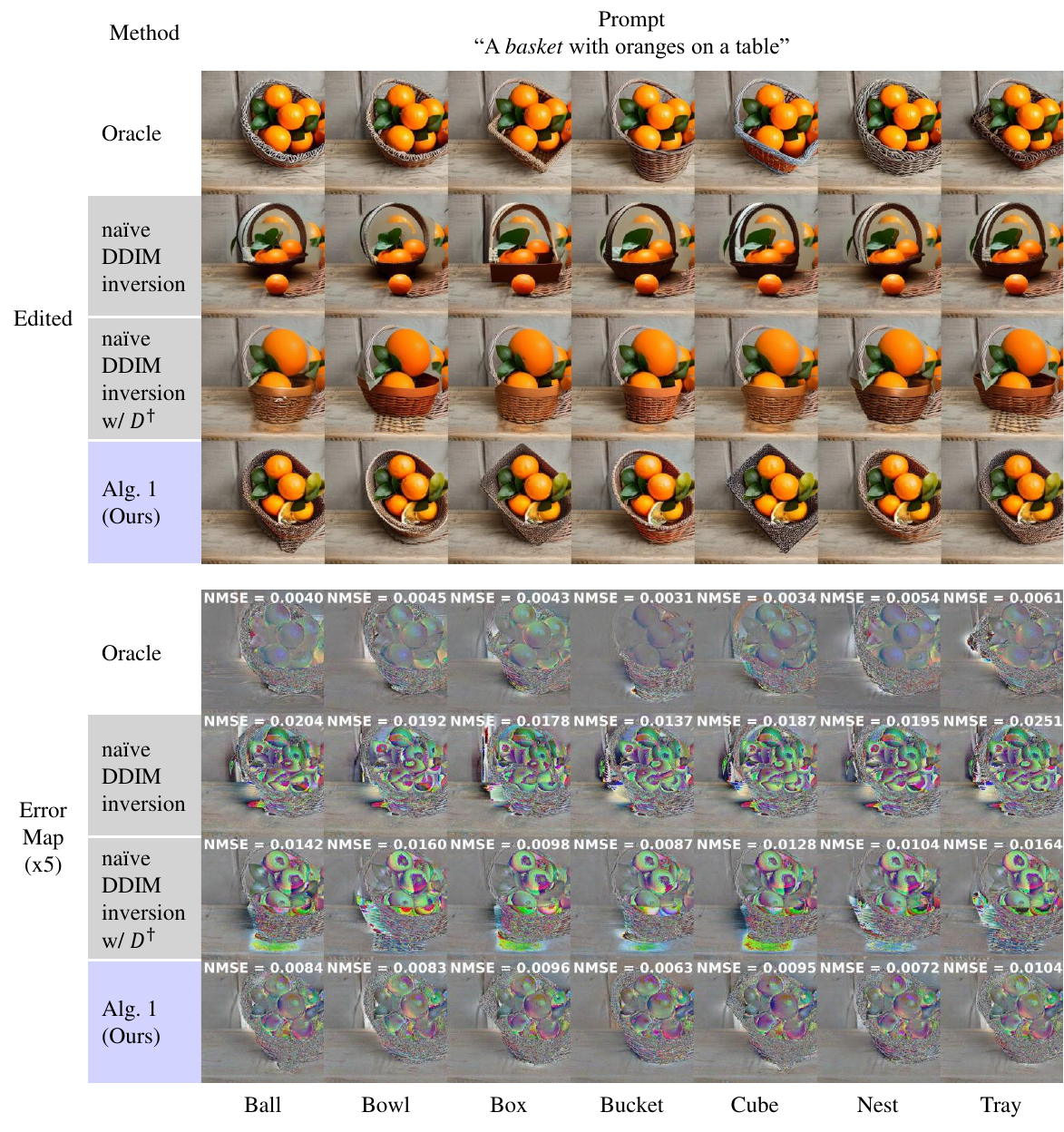}}
    \caption{Additional experiment result on background-preserving editing using prompt "A \emph{basket} with oranges on a table". This experiment has the same setting with \cref{fig:edit_1}. As we attempt to edit the \emph{basket}, the oranges should be treated as part of the background and should not be changed. Our Alg. 1 (row 4) preserves the oranges well, similar to the Oracle (row 1). In contrast, the na\"ive DDIM inversion indicates significant changes to the appearance of oranges (rows 2-3).}
    \label{fig:edit_2}
\end{figure*}

In Sec. 5.3 of the main paper, we experimentally demonstrated our proposed methods enable the background-preserving editing, without the need for the original latents. In Figs. \ref{fig:edit_1} and \ref{fig:edit_2}, we show additional results with different prompts. 

\end{document}

%% file: preamble.tex
\usepackage[dvipsnames]{xcolor}

\usepackage{algpseudocode, algorithm}
\usepackage{amsmath,amsthm,amssymb,mathtools}
\usepackage{mathcommand}
\usepackage{graphicx}
\usepackage{subcaption}
\usepackage{nicefrac}
\usepackage{comment}
\usepackage{booktabs}
\usepackage{multirow}
\usepackage{array}
\usepackage{tikz}
\usetikzlibrary{decorations.text}
\usepackage{pgfplots}
\usepackage{forest}
\usepackage{rotating}
\usepackage{bbding}
\usepackage{pgf}

\newtheoremstyle{funny}
  {}{}
  {\itshape}
  {}
  {\bfseries}
  {.}
  { }
  {%
   \thmname{#1}
   \thmnumber{ #2}
   \thmnote{ {\mdseries\iffunny(\fi#3\iffunny)\fi}}
   \global\funnytrue 
  }
\newif\iffunny
\newcommand{\nobrackets}{\global\funnyfalse}

\theoremstyle{funny}

\newtheorem{proposition}{Proposition}
\theoremstyle{funny}

\newtheorem{lemma}{Lemma}

\algnewcommand\algorithmicswitch{\textbf{switch}}
\algnewcommand\algorithmiccase{\textbf{case}}
\algnewcommand\algorithmicassert{\texttt{assert}}
\algnewcommand\Assert[1]{\State \algorithmicassert(#1)}%
\algdef{SE}[SWITCH]{Switch}{EndSwitch}[1]{\algorithmicswitch\ #1\ \algorithmicdo}{\algorithmicend\ \algorithmicswitch}%
\algdef{SE}[CASE]{Case}{EndCase}[1]{\algorithmiccase\ #1}{\algorithmicend\ \algorithmiccase}%
\algtext*{EndSwitch}%
\algtext*{EndCase}%

\newcommand{\cmark}{\textcolor{green}{\Checkmark}}
\newcommand{\xmark}{\textcolor{red}{\XSolidBrush}}

\usepackage{colortbl}
\usepackage{tabularx}
\definecolor{bGreen}{HTML}{D3FFD3}
\definecolor{bBlue}{HTML}{D3D3FF}
\definecolor{bRed}{HTML}{FFD3D3}
\definecolor{bGray}{HTML}{D3D3D3}

%% file: math_commands.tex
\usepackage{amsmath,amsfonts,bm}

\def\eqref#1{equation~\ref{#1}}

\def\1{\bm{1}}

\def\vtheta{{\bm{\theta}}}

\def\vd{{\bm{d}}}

\def\vx{{\bm{x}}}
\def\vy{{\bm{y}}}
\def\vz{{\bm{z}}}

\def\mI{{\bm{I}}}

\DeclareMathAlphabet{\mathsfit}{\encodingdefault}{\sfdefault}{m}{sl}
\SetMathAlphabet{\mathsfit}{bold}{\encodingdefault}{\sfdefault}{bx}{n}

\def\gD{{\mathcal{D}}}
\def\gE{{\mathcal{E}}}

\def\gO{{\mathcal{O}}}

\def\gX{{\mathcal{X}}}
\def\gY{{\mathcal{Y}}}

\def\sR{{\mathbb{R}}}

\newcommand{\R}{\mathbb{R}}

